\documentclass{article} % For LaTeX2e
\usepackage{iclr2026_conference,times}

\usepackage{amsmath,amsfonts,bm}

\def\eqref#1{equation~\ref{#1}}
\def\1{\bm{1}}

\DeclareMathAlphabet{\mathsfit}{\encodingdefault}{\sfdefault}{m}{sl}
\SetMathAlphabet{\mathsfit}{bold}{\encodingdefault}{\sfdefault}{bx}{n}

\usepackage{hyperref}
\usepackage{url}

\usepackage{CJKutf8}

\usepackage{lineno}
\usepackage{hyperref}       % hyperlinks
\usepackage{url}            % simple URL typesetting
\usepackage{booktabs}       % professional-quality tables
\usepackage{amsfonts}       % blackboard math symbols
\usepackage{nicefrac}       % compact symbols for 1/2, etc.
\usepackage{microtype}      % microtypography
\usepackage{xcolor}         % colors
\usepackage{graphicx}
\usepackage{enumitem}
\usepackage{url}
\usepackage{xcolor}
\usepackage{colortbl}
\usepackage[most]{tcolorbox}
\definecolor{commentcolor}{rgb}{0.4, 0.4, 0.4} 

\usepackage{multirow}
\usepackage{multicol}
\usepackage{makecell}
\usepackage{wrapfig}
\usepackage{booktabs}
\usepackage{algorithm}
\usepackage{algorithmic}
\usepackage{listings}
\usepackage{caption}

\usepackage{subcaption}

\usepackage{tcolorbox}
\usepackage{amsfonts}
\newcommand*{\imgintext}[1]{%
  \raisebox{-.1\baselineskip}{%
    \includegraphics[
      height=.9\baselineskip,
      width=.9\baselineskip,
      keepaspectratio,
    ]{#1}%
  }%
}

\lstdefinestyle{pystyle}{
    language=Python,
    basicstyle=\ttfamily\small,  % 使用等宽字体，大小为 small
    numbers=left,                % 行号在左侧
    numberstyle=\tiny,           % 行号字体大小为 tiny
    stepnumber=0,                % 每行显示行号
    numbersep=5pt,               % 行号与代码之间的距离
    backgroundcolor=\color{white}, % 背景颜色
    showspaces=false,            % 不显示空格
    showstringspaces=false,      % 不显示字符串中的空格
    showtabs=false,              % 不显示制表符
    frame=single,                % 单线框
    rulesepcolor=\color{black},  % 设置边框颜色为黑色
    tabsize=2,                   % 制表符大小为 4
    captionpos=b,                % 标题位置在底部
    breaklines=true,             % 自动换行
    breakatwhitespace=false,     % 不在空白处换行
    keywordstyle=\color{blue},   % 关键字颜色
    commentstyle=\color{brown},  % 注释颜色
    stringstyle=\color{red},     % 字符串颜色
}
\newcommand{\blue}[1]{\textcolor{black}{#1}}

\newcommand{\imgintitle}[2][]{%
  \raisebox{-.27\height}{\includegraphics[height=1.3em,#1]{#2}}%
}

\title{ \imgintitle{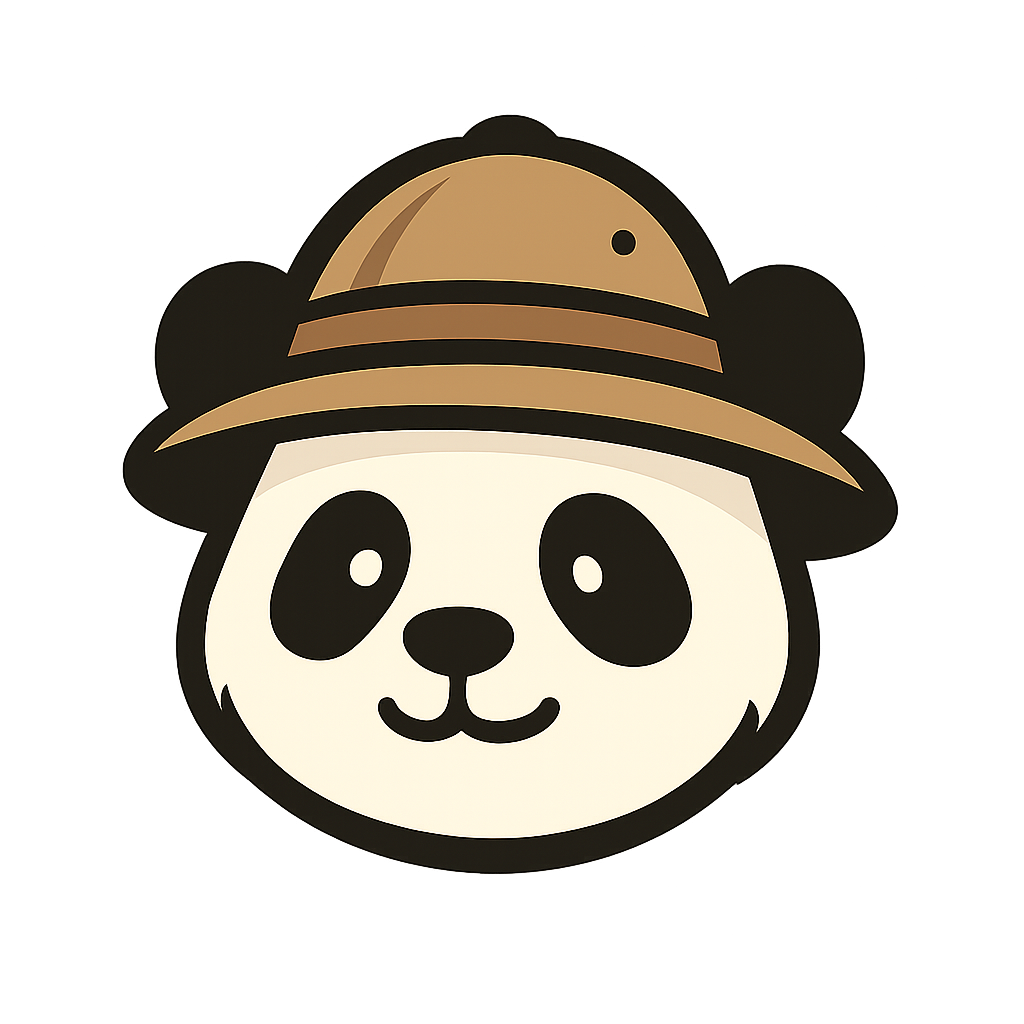} ChinaTravel: An Open-Ended Travel\\Planning Benchmark with Compositional\\Constraint Validation for Language Agents}

\author{Jie-Jing Shao\textsuperscript{1}\thanks{Equal contribution}, Bo-Wen Zhang\textsuperscript{1,2*}, Xiao-Wen Yang\textsuperscript{1,3*}, Bai-Zhi Chen\textsuperscript{1}, \\
\textbf{Si-Yu Han\textsuperscript{1,3}, Jing-Hao Pang\textsuperscript{1,2}, Wen-Da Wei\textsuperscript{1,3}, Guohao Cai\textsuperscript{4}, }\\
\textbf{Zhenhua Dong\textsuperscript{4}, Lan-Zhe Guo\textsuperscript{1,2}\thanks{Corresponding authors}, Yu-Feng Li\textsuperscript{1,3$\dag$}} \\
\textsuperscript{1}State Key Laboratory of Novel Software Technology, Nanjing University, China\\
\textsuperscript{2}School of Intelligence Science and Technology, Nanjing University, China\\
\textsuperscript{3}School of Artificial Intelligence, Nanjing University, China\\
\textsuperscript{4}Noah's Ark Lab, Huawei, China\\
\texttt{\{shaojj,zhangbw,yangxw,guolz,liyf\}@lamda.nju.edu.cn} 
}

\iclrfinalcopy % Uncomment for camera-ready version, but NOT for submission.
\begin{document}

\maketitle

\vspace{-.2in}

\begin{abstract}

   Travel planning stands out among real-world applications of \emph{Language Agents} because it couples significant practical demand with a rigorous constraint-satisfaction challenge. 
   However, existing benchmarks 
   % typically rely on synthetic queries with limited constraints and explicit intent, which diverge from real-world scenarios, 
   \blue{primarily operate on a slot-filling paradigm, restricting agents to synthetic queries with pre-defined constraint menus, which fails to capture the open-ended nature of natural language interaction, 
   where user requirements are compositional, diverse, and often implicitly expressed}. 
   To address this gap, we introduce \emph{ChinaTravel}, with four key contributions: 1) a practical sandbox aligned with the multi-day, multi-POI travel planning, 2) a compositionally generalizable domain-specific language (DSL) for scalable evaluation, covering feasibility, constraint satisfaction, and preference comparison 3) an open-ended dataset that integrates diverse travel requirements and implicit intent from 1154 human participants, and 
   4) fine-grained analysis reveal the potential of neuro-symbolic agents in travel planning, achieving a 37.0\% constraint satisfaction rate on human queries, a 10$\times$ improvement over purely neural models, \blue{yet highlighting significant challenges in compositional generalization}. 
   Overall, ChinaTravel provides a foundation for advancing language agents through compositional constraint validation in complex, real-world planning scenarios. 
   % These findings highlight ChinaTravel as a pivotal milestone for advancing language agents in complex, real-world planning. % scenarios.
   % These results position \emph{ChinaTravel} as a challenging benchmark for advancing language agents in complex real world planning. 
   % These findings position ChinaTravel as a challenging benchmark with compositional constraint validation that advances reliable language agents in complex, real-world planning scenarios. 
   % These findings position ChinaTravel as a challenging benchmark for advancing reliable language agents in complex, real-world planning scenarios. 

\end{abstract}

\vspace{-.2in}
\begin{center}
    \includegraphics[height=1em]{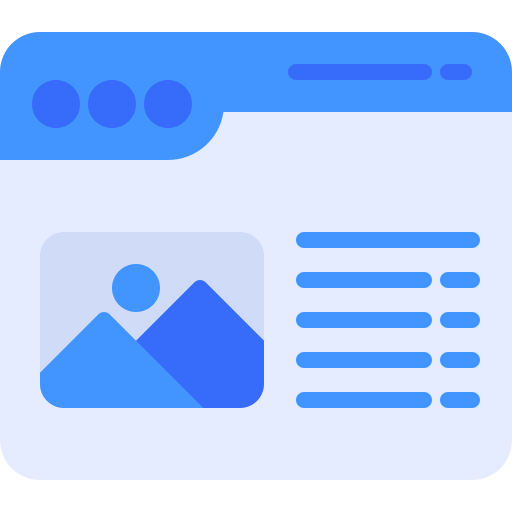} \hspace{1pt}  \href{https://www.lamda.nju.edu.cn/shaojj/ChinaTravel/index.html}{Project Page} \quad
    \includegraphics[height=1em]{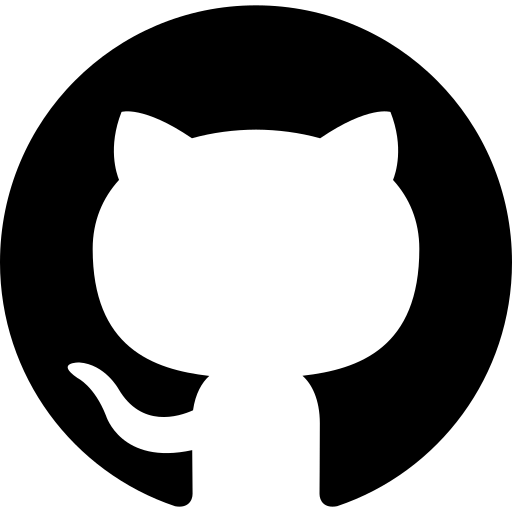} \hspace{1pt} \href{https://github.com/LAMDASZ-ML/ChinaTravel}{Codebase} \quad
    \includegraphics[height=1em]{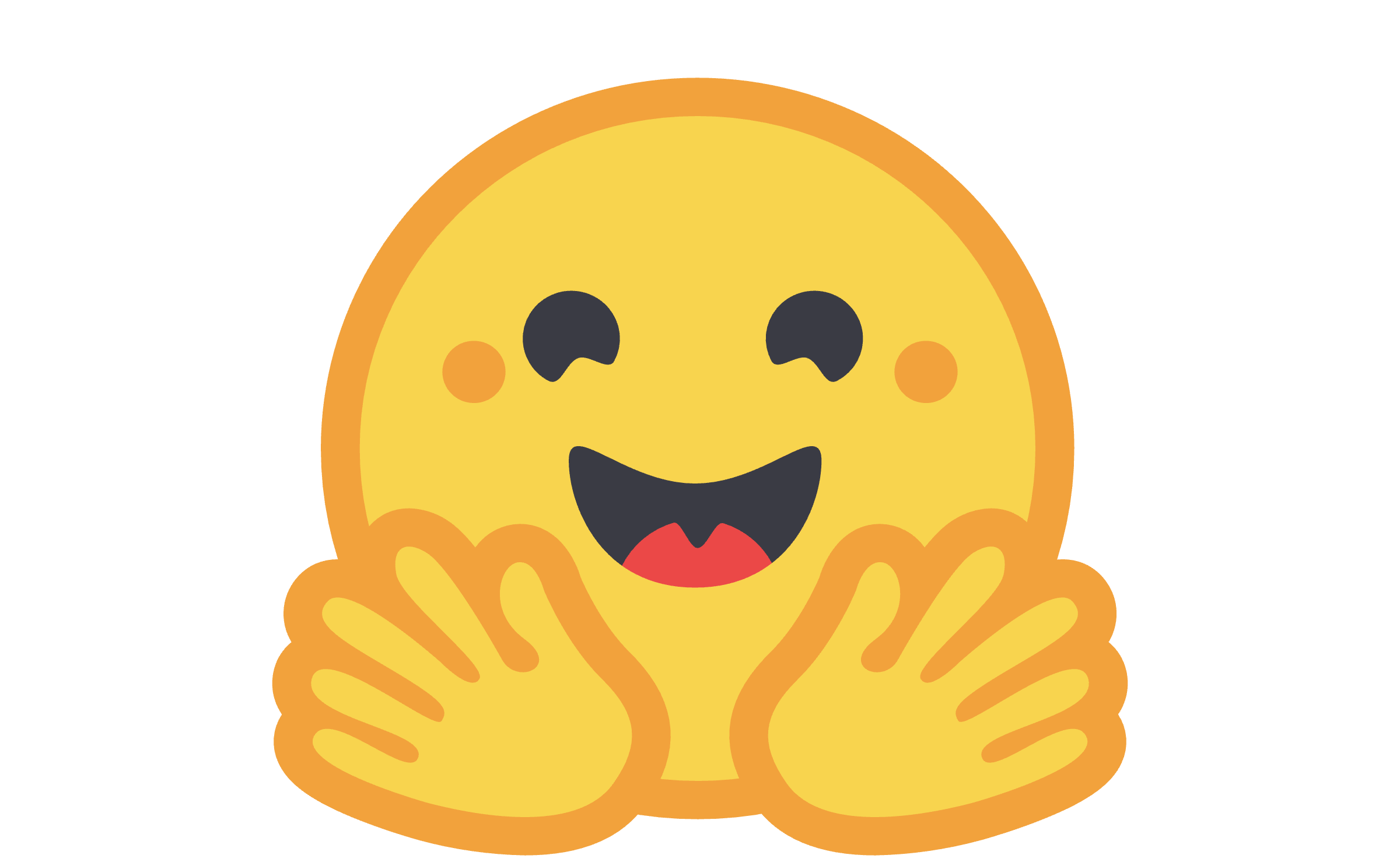} \hspace{.8pt} \href{https://huggingface.co/datasets/LAMDA-NeSy/ChinaTravel}{Dataset}
\end{center}

\vspace{-.2in}

\section{Introduction}

A long-standing goal in AI is to build reliable, general planning agents that assist humans in real-world tasks. Recent advances in LLMs 
have sparked the rapid development of \emph{Language Agents}, which employ LLMs to perceive the surroundings, reason solutions, and take appropriate actions, ultimately building autonomous planning agents~\citep{DBLP:conf/nips/ShinnCGNY23, DBLP:conf/iclr/YaoZYDSN023, DBLP:journals/corr/abs-2309-07864,DBLP:conf/iclr/JimenezYWYPPN24}. 
Among numerous real-world tasks, travel planning stands out as a significant domain, presenting both academic challenges and practical value due to its inherent complexity and real-world relevance. 
\blue{Beyond the travel community itself, such planning scenarios have also become a natural testbed for general constraint-aware planning and reasoning, thereby attracting growing interest from the broader AI community~\citep{DBLP:conf/icml/KambhampatiVGVS24, chen2025sets, choi2025atlas}.} 
Specifically, given a query, agents require information integration from various tools (e.g., searching for flights, restaurants, and hotels) to generate a feasible itinerary. 
This involves making interdependent decisions across multiple aspects such as spatial, temporal, and financial dimensions, all while meeting the user's requirements and preferences (e.g., budget, dining habits, etc). 
% This travel planning task presents both significant practical value and important research challenges. As a pervasive yet complex activity, it demands considerable time investment, creating compelling need for AI assistance. Academically, it constitutes a long-horizon planning objective that involves various hard and soft constraints, posing unique challenges for planning agents. 
% Existing benchmarks fail to capture the complexity of real-world travel planning, relying on synthetic queries and oversimplified constraints. 

To assess whether language agents meet users requirements in travel planning,~\citep{zheng2024natural} present the Trip Planning benchmark for intercity itinerary conditioned on flights information. ~\citet{TravelPlanner} provide a pivotal benchmark, TravelPlanner~\citep{TravelPlanner}, with a real-world travel sandbox and various tools to intergrate multi-dimensional information. 
\blue{However, critical limitations persist in these benchmarks stemming from their reliance on a slot-filling interaction paradigm, where agents merely extract values for fixed attributes (e.g., budget, date), ignoring the compositional logic inherent in human cognition~\citep{fodor1975language, fodor2008lot, piantadosi2016logical}, highlighting a substantial research gap toward LLM Agents capable of genuine natural-language interaction. This effectively restricts agents to a closed set of intents, highlighting a substantial gap toward LLM Agents capable of genuine, open-ended natural language interaction.} 
% However, % the TravelPlanner benchmark 
% there are four critical limitations: 
% This systemic limitation manifests in four key aspects: 
The limitations arise in: 
\romannumeral1) \textbf{Task Bias:} % favors intercity itineraries, omitting scheduling of intracity events, which is ubiquitous in practice yet quite challenging. 
% Favoring intercity itineraries while omitting intracity scheduling. 
% The lack of compositional expressivity in previous benchmarks made it cannot model the complex, interdependent constraints of intracity events. 
favors intercity itinerary while omitting intracity scheduling, where complex, interdependent constraints are desirable. 
\romannumeral2) \blue{\textbf{Inflexible Constraint Verification:}} relies on fixed rule lists, which cannot generalize to diverse, unseen requirements spanning compositional concept space. 
\romannumeral3) \textbf{Synthetic Query Construction:} includes only templated LLM-synthesized queries, underrepresenting open-ended, semantics-rich human requests. 
\romannumeral4) \blue{\textbf{Misleading NeSy Evaluation:}} emphasizes LLM-only shortcomings in constraint adherence, yet largely ignores neuro-symbolic methods that couple neural language understanding with verifiable symbolic reasoning. 
Within months of TravelPlanner's release, \citet{MITSMT} proposed a neuro-symbolic pipeline: an LLM extracts constraints from templated queries and a formal verification tool yields plans, achieving 97\% success rate vs. 4\% for LLM-only baseline. This suggests templated, fixed-constraint setups are near saturation, \blue{failing to expose the true bottleneck of NeSy methods from natural language interaction.}
% underscoring the need for an open-world benchmark grounded in authentic requirements. 

\begin{figure*}[t]
    \centering

    \vspace{-.55in}
    \includegraphics[width=0.9\linewidth]{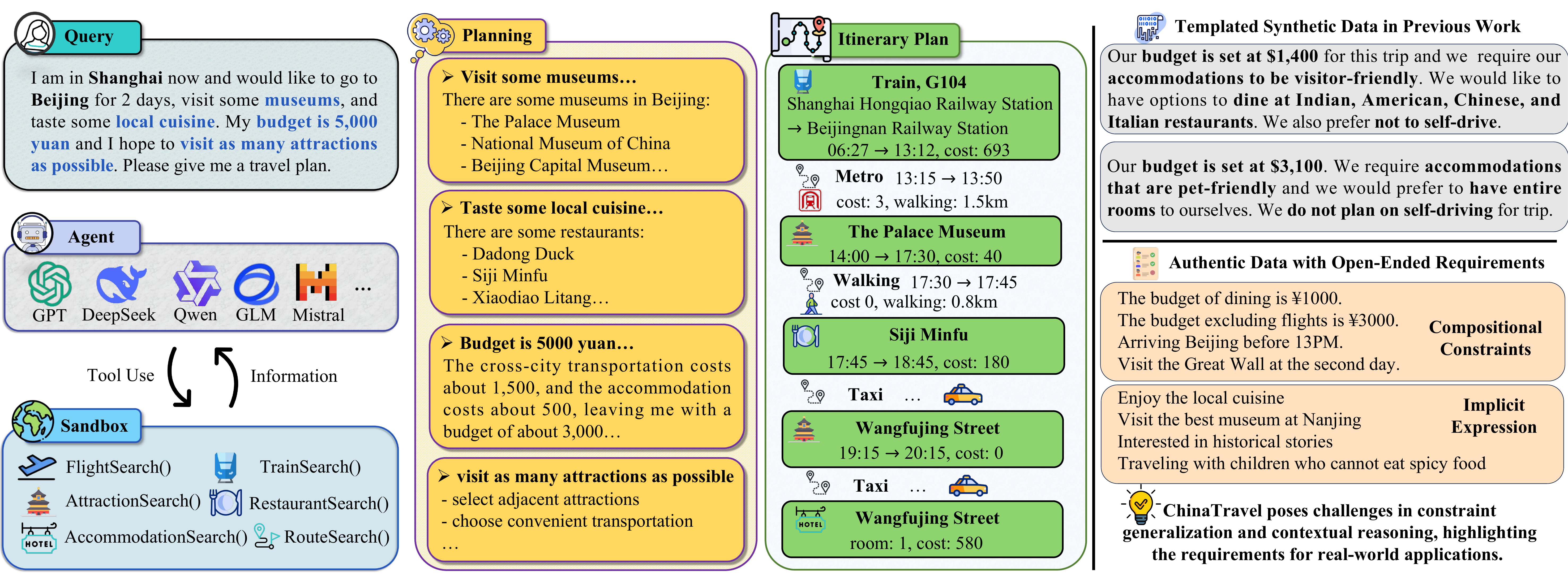}
   
    \vspace{-.1in}
    \caption{\textbf{Overview of ChinaTravel}. Given a query, language agents employ various tools to gather information
    and plan a multi-day multi-POI itinerary. The agents are expected to provide \textbf{a feasible 
    %and reasonable 
    plan}
    while satisfying the \textbf{logical constraints} and \textbf{preference requirements}. \blue{Crucially, ChinaTravel is designed to facilitate the shift from rigid \textbf{slot-filling paradigms} to \textbf{open-ended natural language interaction}, challenging agents to handle the diverse, compositional logic inherent in authentic human intent.} To provide convenience for global researchers, we provide the English version here.} 
    % \vspace{-1cm}
    \label{overview}

    \vspace{-.3in}
\end{figure*}

% In this paper, we introduce ChinaTravel, a novel benchmark for evaluating language agents in travel planning.
To address the gap, we introduce ChinaTravel, an open-ended travel planning benchmark. 
It concentrates on multi-point-of-interest (multi-POI) itineraries (as illustrated in Fig.~\ref{overview}) and supports the compositional constraints evaluation with authentic Chinese travel queries. 
It is more realistic and challenging, providing a desirable testbed for real-world travel planning.  
The main contributions are: 

% ChinaTravel is built in a modular framework with (1) a rich sandbox environment with Chinese travel information, (2) diverse evaluation metrics covering feasibility, constraint satisfaction, and  preference comparison, and (3) realistic travel requirements contain both LLM-synthetic and human questionnaire queries. 
% We constructed ChinaTravel in five stages, including manual schema and API design, LLM-assisted generation of data entries, manual quality control, data collection from human users with open requirements, and preference data construction. 
% Our evaluation pipeline automatically verifies the provided plans with the requirements annotations. 
% An additional subset with rich travel preferences is constructed to provide an evaluation for future language agents. 
% We introduce ChinaTravel, a benchmark tailored to the complex requirements of travel planning in a real-world Chinese context. 
% \begin{itemize}[left=0pt] 
   % \item 

\noindent\textbf{ChinaTravel Sandbox:} 
We introduce a real-world sandbox with a suite of tools aligned with the ubiquitous multi-day multi-POI itinerary planning. 
It provides the detailed travel information and supports the integration and planning of spatiotemporal schedules.

\noindent\textbf{Compositional Constraints Evaluation:} 
We present a domain specific language that programmatically composes atomic concepts of travel attributes across spatial, temporal, cost, and type dimensions to express compositional constraints. It supports scalable requirement specification and automated constraint verification, with metrics for feasibility, constraint satisfaction, and preference ranking.

   % \item 
\noindent\textbf{Open-Ended Travel Dataset:} % ChinaTravel includes both LLM-generated and human-derived queries, offering a realistic and open testbed for evaluating agents in addressing authentic and multifaceted travel needs, which requires contextual reasoning.  
Beyond the data synthesis pipeline as previous benchmarks, ChinaTravel integrates human-authored queries to create realistic travel planning scenarios. The validation set contains 154 human queries with combinatorial constraint requirements absent from synthetic data, while the test subset provides 1000 open-scenario queries. This structure specifically assesses agents' generalization capabilities for unseen constraint composition. 

    % \item 
\noindent\textbf{Empirical Analysis of Neuro-Symbolic Agents:} Extensive experiments are conducted and the results reveal that neuro-symbolic agents significantly outperform pure LLM-based solutions on constraints satisfaction, achieving a success rate of 37.0\% compared to 2.60\% by purely neural methods, thus highlighting their promise for travel planning tasks. 
We also identify the key challenges of open-world requirements: open contextual grounding, and unseen concept composition, providing a foundation for advancing reliable agents toward real-world applicability. 
% \end{itemize}

Overall, ChinaTravel provides a challenging benchmark that rigorously assesses constraint satisfaction for travel planning, 
serving as a % critical 
bridge between academic research and practical applications.

\section{ChinaTravel Benchmark}

% Motivated by the significant travel demand in China, ChinaTravel offers a sandbox environment for generating multi-day, multi-POI itineraries for specified cities. 
Motivated by China's substantial travel demand, ChinaTravel provides a sandbox environment for generating multi-day itineraries with multiple POIs within specified cities. 
% It includes arrangements for attractions, restaurants, accommodations, and transportation between events, aiming to advance the practice of language agent solutions for real-world travel planning. 
It is meticulously designed to provide a comprehensive and scalable evaluation framework % for language agents 
in travel planning, encompassing three critical dimensions: environmental feasibility, constraint satisfaction, and preference comparison. 
\subsection{Environment Information}
ChinaTravel provides a sandbox with real-world travel information. We collect information from 10 of the most popular cities in China. %, including Beijing, Chengdu, Chongqing, Guangzhou, Hangzhou, Nanjing, Shanghai, Shenzhen, Suzhou, and Wuhan. 
It includes 720 airplanes and 5,770 trains connecting these cities, with records detailing departure and arrival times, origins, destinations, and ticket prices. 
Additionally, the dataset contains 3,413 attractions, 4,655 restaurants, and 4,124 hotels, each annotated with name, location, opening hours, and per-person prices. Type annotations for these POIs are included to meet user needs. 
% Fig.~\ref{environment} has provided an illustration of the collected information from Beijing and Nanjing, two of the most popular cities in China. 
For a realistic interaction, we simulate the API interface of real market applications to query real-time information. 
We present 25 environmental constraints grouped into six categories: dietary, accommodation, transportation, temporal, spatial, and attraction-related. 
% The detailed designs of the sandbox are available in App.~\ref{app_sandbox}. 
It acts as a feasibility metric, ensuring that the generated plans are both valid and effective. For example, POIs in the plan must exist in the designated city, transportation options must be viable, and time information must remain accurate. 
See App.~\ref{app_sandbox} for design details of sandbox and environmental constraints. 
% Tab.~\ref{Metrics} summarizes the 25 environmental constraints across six categories: Dietary, Accommodation, Transportation, Temporal, Spatial, and Attractions. 
% The environmental constraints are designed to ensure the reliability of the results. That is, the POIs visited in the plan must exist in the corresponding city, the transportation methods provided in the plan must be feasible, and the corresponding time information should also be reliable. 
% For example, there should indeed be a subway line that can depart from Beijing Capital International Airport and arrive at the Palace Museum in 80 minutes. 

\subsection{Logical Constraint}
\label{sec2_2}
\begin{table*}
    \vspace{-.37in}
  \caption{ChinaTravel's Domain-Specific Language (DSL) for logical constraints.}
    \vspace{-.07in}
  \small
\setlength{\tabcolsep}{.2pt}
\begin{tabular}{p{2.0cm} p{3.0cm} p{8.5cm}}
\toprule
Name & Syntax & Description \\
\midrule
variables & $x,y,z,\cdots$ & Variables that refer to activities in the travel planning domain.\\
not & $not\ expr$ & The negation of an Boolean-valued expression.\\
and,or & $expr_1 \textit{ and } expr_2$ & The conjunction/disjunction of an Boolean-valued expression.\\
% $\forall, \exists$ & $\textit{forall}(\{var\}, expr)$ & $expr$ is an expression that contains a variable $var$; return true if all variables in set $\{var\}$ that satisfy $expr$.\\
$<,>,==$ & $expr_1 < expr_2$ & Return an expression with built-in number comparison functions.\\
$+, -, *, /$ & $expr_1 + expr_2$ & Return an expression with built-in number calculation functions.\\
attributes & $cost(var)$ & A function that takes activities as inputs and returns the attributes, such as cost, type or time. \\ 
relation & $dist(expr_1, expr_2)$ & A function that takes locations as inputs and returns the distance.\\ 
effect & $var=expr$ & An assignment affects a variable $var$ with the expression $expr$.\\
\multirow{2}{*}{\makecell[l]{union, inter,\\ diff}} & $uni(\{var\}_1, \{var\}_2)$ & Return a set with the built-in union/intersection/difference operations of given two sets. \\
enumerate & $\textit{for } var \textit{ in } \{var\}$ &  Enumerate all variables in the collection $\{var\}$.\\
when & $\textit{if } expr: \textit{effect}$ & The conditional effect takes a Boolean-valued condition of the expression $expr$, and the effect \textit{effect}. \\
\bottomrule
\end{tabular}
\label{tab_dsl}
    \vspace{-.05in}
\end{table*}

\begin{figure}[t]
   \centering
   % 第一张图片
   \begin{subfigure}{0.32\textwidth}
       \centering
       \includegraphics[width=\textwidth]{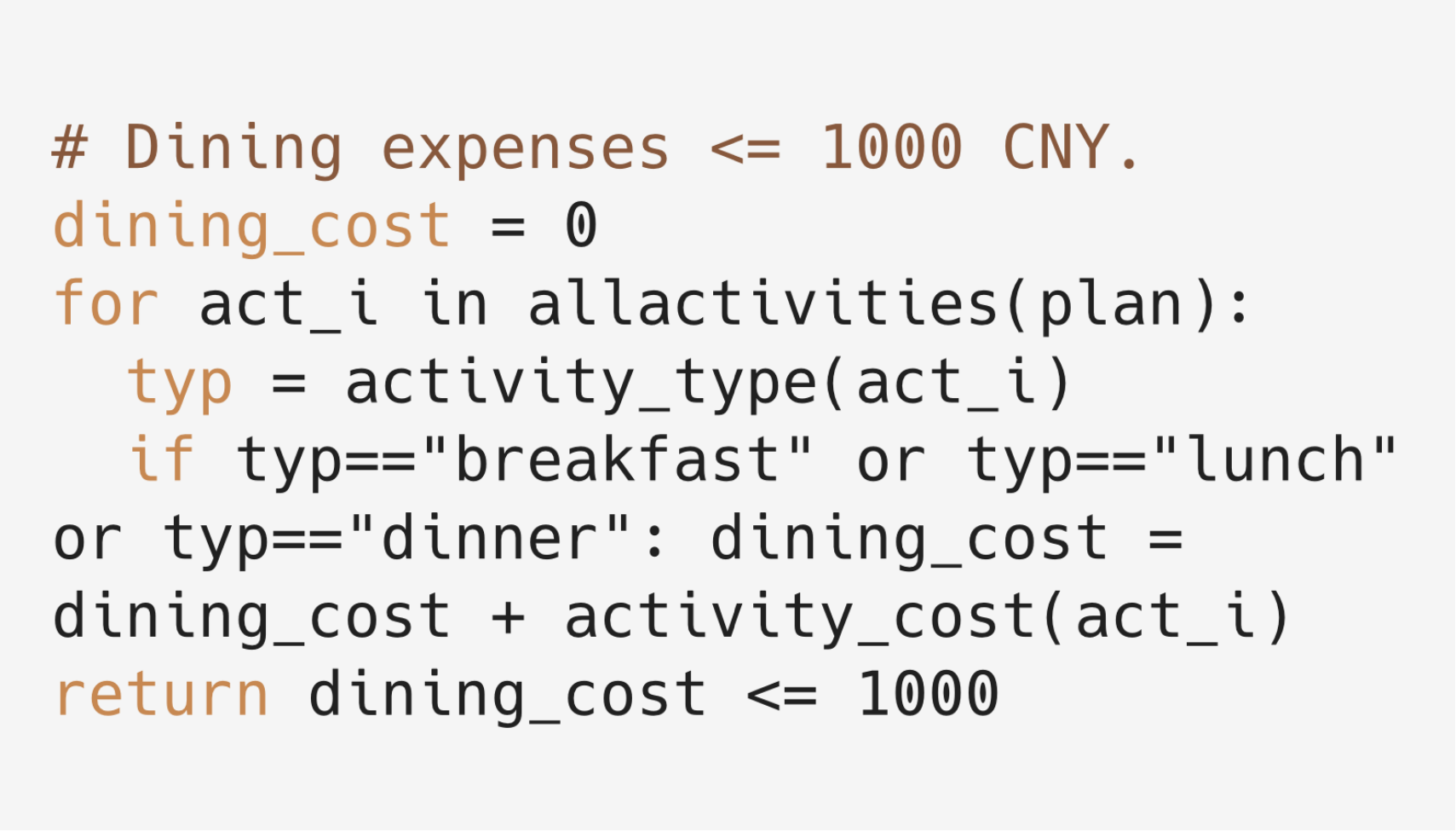}
       \caption{Dining expenses.}
       \label{fig:dsl1}
   \end{subfigure}
   % 第二张图片
   \begin{subfigure}{0.32\textwidth}
       \centering
       \includegraphics[width=\textwidth]{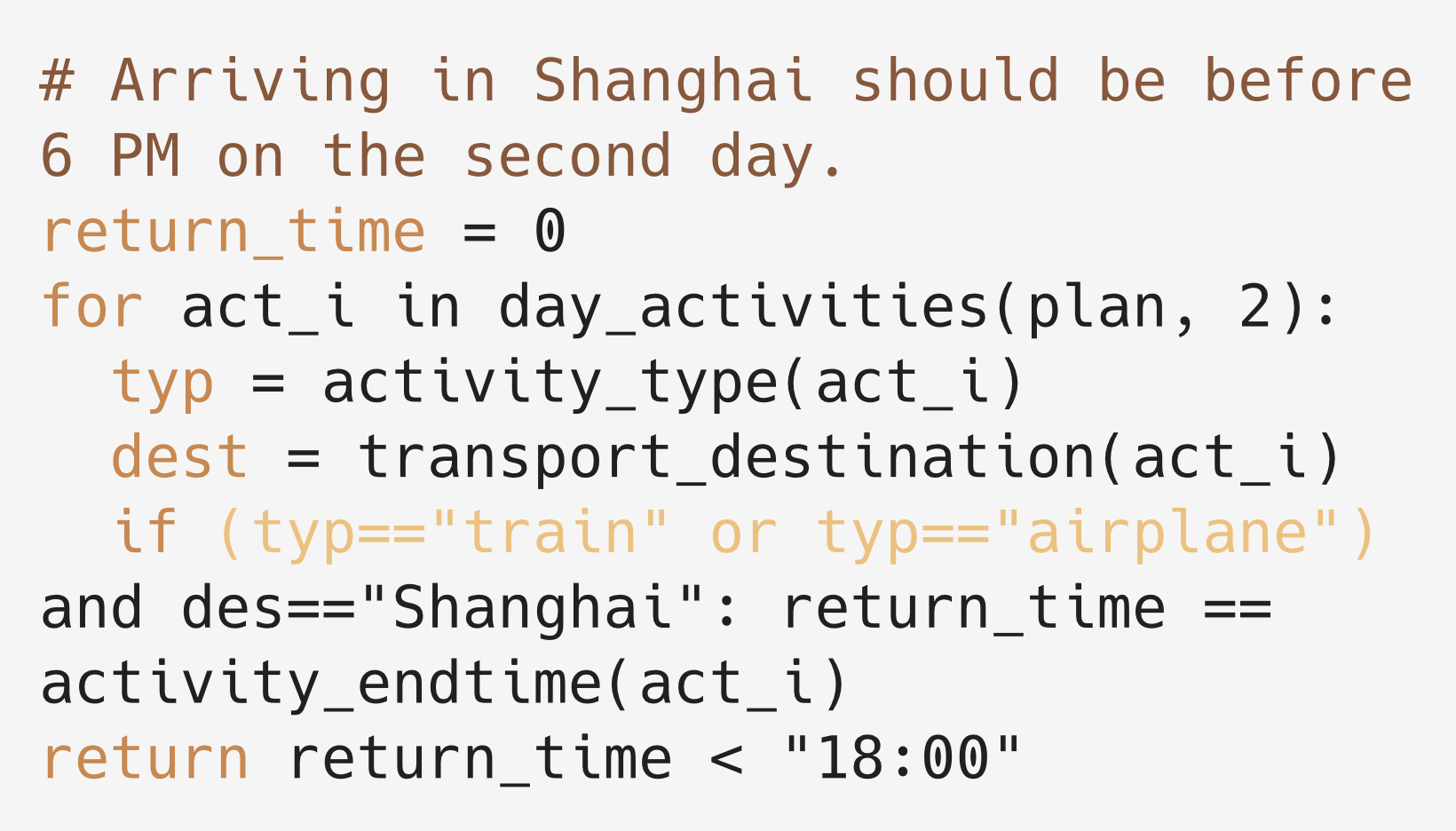}
       \caption{Arrived Time.}
       \label{fig:dsl2}
   \end{subfigure}
   \begin{subfigure}{0.32\textwidth}
       \centering
       \includegraphics[width=\textwidth]{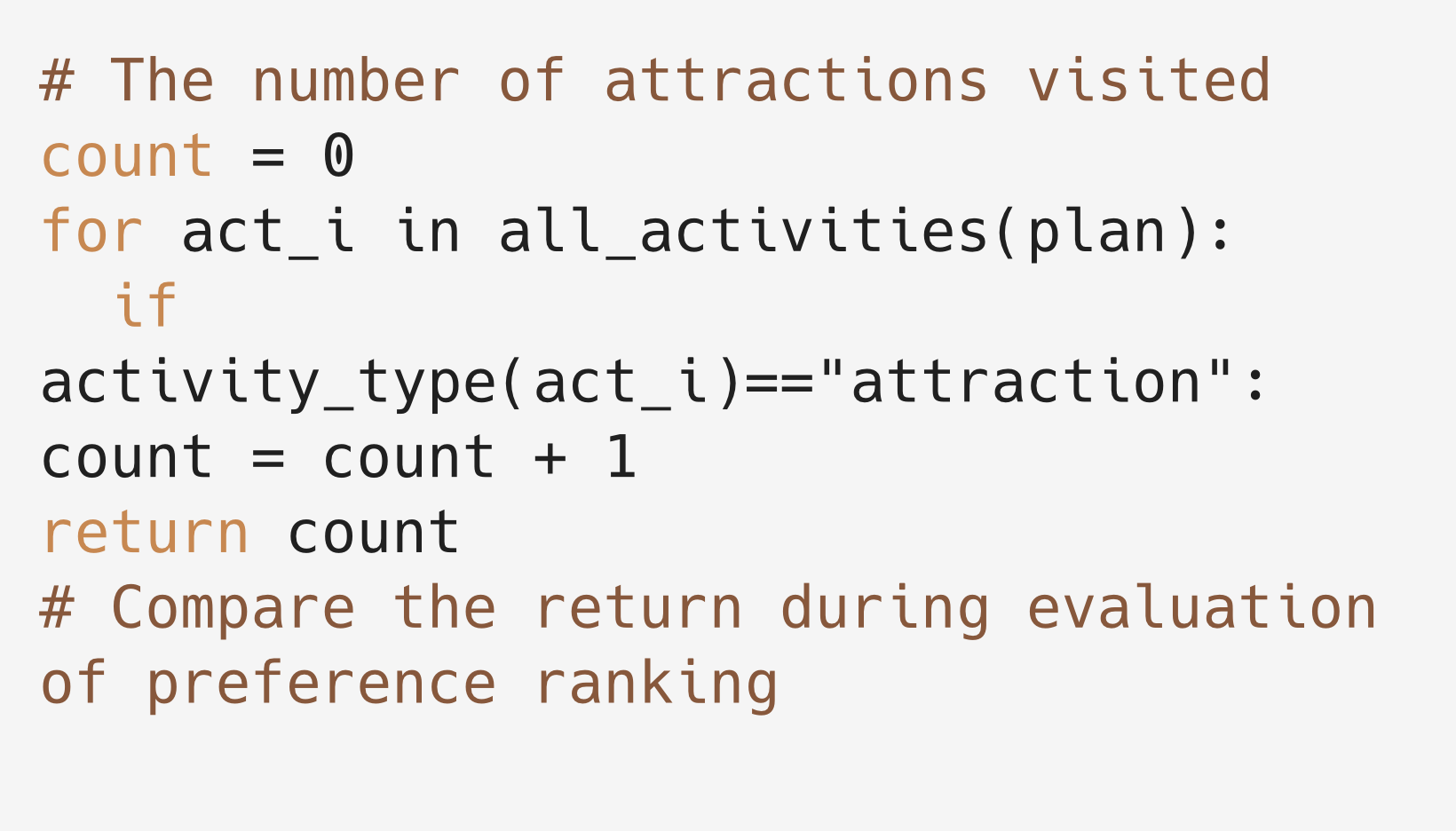}
       \caption{Count of attraction visited.}
       \label{fig:dsl3}
   \end{subfigure}
   \caption{Examples of DSL expressions for logical constraints and preference ranking.}

   \label{fig:all_images}
\end{figure}

A crucial ability for travel planning is to effectively satisfy personalized user needs. 
Prior benchmark~\citep{TravelPlanner} evaluates logic with five fixed concepts (total budget, room rules, room types, cuisines, transportation types), where each concept is mapped to a specific requirement. 
% This fixed list validates only a narrow slice of open-world needs. 
\blue{Although it has gained much attention, it effectively confines constraint satisfaction to \emph{propositional logic}, where extracting constraints from template-synthesized queries is relatively straightforward. In this setting, the system reasons about truth relations between atomic propositions without examining the complex internal structure or relationships of the travel events.} 
% We extend the form of logical constraints from TravelPlanner~\citep{TravelPlanner} and present a Domain-Specific Language (DSL) to support general compositional reasoning in logical constraints. 
% ChinaTravel's DSL is a general set of pre-defined concept functions with built-in implementations, as listed in Tab.~\ref{tab_dsl}. 
% TravelPlanner relies on 5 pre-defined concepts \{total budget, room rules, room types, cuisines, and transportation types\}, to evaluate the logical constraints, where each concept is equivalent to a specific logical requirement. We find this design limits the ability to validate diverse logical needs in an open-world context. 
% For example, it cannot express “total dining expenses $\leq$ 1000 CNY” or “arrive in Shanghai before 18:00 on day 2,” even when the generated plan includes costs and timestamps. Each new requirement therefore requires manual rule additions. 
For example, it cannot express that ``dining budget is 1000 CNY'' or that ``arriving in Shanghai should be before 6 PM on day 2'', despite the generated itinerary already including the expenses and time information of each activity. 
Each new logical requirement necessitates human intervention for incremental definition and validation. 
\blue{It is desirable to extend the constraint design and validation into a combinatorial language space, which can combines and validates predicates to enable expressive requirements over travel events.} 
% To address this issue, our approach is grounded in a DSL-based solution that leverages basic concept functions and syntax to express and fulfill various logical requirements. 
We address this gap with a DSL-based solution that enables compositional specification and validation of logical constraints. 
The proposed DSL provides a small set of basic concept functions and a Python-like syntax, so diverse requirements can be written as compositions of primitives and automatically perform validation of plans using a Python compiler. % (see Table X for primitives and Appendix Y for a short tutorial). 
Fig.~\ref{fig:dsl1} and~\ref{fig:dsl2} illustrate how the DSL express the user requirements (see Tab.~\ref{tab_concept_func_0} for basic concepts and App.~\ref{app_concept_func} for a hand-on tutorial with more examples). 
% The DSL can represent varying requirements through concept composition in a Python format, and perform automated validation of plans using a Python compiler. 
% This strategy maximizes the evaluation capability of the ChinaTravel. 
% In App.~\ref{app_concept_func}, we provide a detailed tutorial on DSL expression with more examples. x x x
This approach removes the need for per-requirement rule engineering and yields scalable evaluation of compositional logical constraints from open-world travel planning.

\subsection{Preference Requirement}
Travel requirements encompass not only hard logical constraints but also soft preferences. 
The term ``soft" implies that these preferences cannot be addressed as binary constraint satisfaction problems, instead, they involve quantitative comparisons based on continuous values. 
This distinction highlights the unique nature of preference-based requirements compared to logical constraints. Common preferences from our surveys include maximizing the number of attractions visited, minimizing transport time between POIs, and visiting positions near the specific POI. 
In ChinaTravel, we formalize such preferences as minimization or maximization objectives via our DSL, thereby providing an automated evaluation. 
% Fig.~\ref{fig:dsl3} provides a detailed case for maximizing the attractions visited. More examples are provided in the App.~\ref{preference_example}. 
Fig.~\ref{fig:dsl3} illustrates maximizing attractions visited, more examples appear in App.~\ref{preference_example}. 

\subsection{Benchmark Construction}

% ChinaTravel provides user queries reflecting diverse requirements through a four-stage process: % that integrates LLM-based generation with questionnaires.

\noindent\textbf{Stage \MakeUppercase{\romannumeral1}: Manual design of database %schema 
and APIs.} 
We collect travel information for multi-day, multi-POI itineraries across attractions, accommodations, and transportation. We define essential POI attributes (e.g., cuisine types, hotel amenities) and build a structured database from public information. APIs are designed to support agent queries via regular expressions and modeled after commercial APIs to ensure realism. See App.~\ref{app_sandbox} for the details of databse. 

\noindent\textbf{Stage \MakeUppercase{\romannumeral2}: Automatic data generation with LLMs.} 
We model travel tasks with core parameters (origin, duration, etc.) and logical constraints. 
% We define common travel information (e.g., origin, destination, days, number of people) and logical constraints to model travel tasks.
% To enable scalable queries, 
% query skeletons are randomly constructed from this information and transformed into natural language queries using an advanced LLM, DeepSeek-V2.5, which is selected for its strong Chinese proficiency, robust instruction-following capabilities, and cost efficiency. 
For scalable generation, we randomly construct query skeletons converted to natural language via DeepSeek-V2.5. 
% The generated queries are categorized into two difficulty levels: \textit{Easy}, with 1 logical requirement beyond basic constraints like people number and trip duration, and \textit{Medium}, with 3-5 additional logical requirements. 
% We encourage the LLM to generate diverse, human-like expressions, such as turning ``Taste Beijing cuisine" into ``Try local food in Beijing." 
% See App.~\ref{app_querygen} for more details about the synthesis. 
Queries are stratified by complexity: \textit{Easy} (1 extra constraint), vs. \textit{Medium} (3-5 constraints), 
with LLM-generated varying expressions (% e.g., 
encouraging ``Taste Beijing cuisine"$\rightarrow$``Try local food"). See App.~\ref{app_querygen} for synthesis details.

% The automatically generated data is categorized into two difficulty levels: In the \textit{Easy} level, user inputs encompass a single logical requirement, sourced from categories such as transportation, restaurants, attractions, and accommodations. In the \textit{Medium} level, user inputs involve 2 to 5 logical requirements, introducing more complex constraints. During the generation, we encourage the LLMs to provide varied and human-like expressions, necessitating a deeper understanding and processing to accurately interpret and fulfill the user's needs. 
% For instance, the logical requirement "taste Beijing cuisine" could correspond to the natural language query: "Try local food in Beijing." We utilize prompt engineering to guide LLMs in refining natural language expressions to facilitate automated generation, as provided in App.~\ref{app_querygen}. 

\noindent\textbf{Stage \MakeUppercase{\romannumeral3}: % Manual 
Quality control and auto-validation.} To ensure data quality, we manually check if the generated query conform to symbolic skeletons, and re-calibrate natural language description that contain ambiguities. 
% Additionally, we calibrate the natural language concept descriptions to closely align with human questioning habits.
Based on the symbolic skeletons, %of queries, 
we verify if the plan can pass the required logical constraints by executing the DSL code via Python compiler. Building on this, we ensure that each query has at least one solution that satisfies the logical constraints %by implementing 
via heuristic search. 

\noindent\textbf{Stage \MakeUppercase{\romannumeral4}: Open requirements from humans.} After the first round of closed-loop development, including LLM-based data generation and annotation, baseline development, and evaluation, 
% we further collected travel requirements from more than 250 humans through questionnaires. Based on a new round of quality control on these data, a more challenging set with 154 queries is constructed. 
we gathered over 250 human requirements via questionnaires. Rigorous quality control yielded 154 queries with novel constraints (e.g., departure time/dining cost), constructing the \textit{Human-154} validation set with DSL-annotated automated evaluation. 
% These queries even include unseen logical constraints in the deployment process, such as `departure time' and `dining cost', reflecting the real challenges of the travel planning system. 
% neural-symbolic systems in travel planning. 
% We carefully annotate the required logical constraints for each query based on the DSL, enabling the automated evaluation of these challenging samples and forming the \textit{Human-154} validation dataset. 
% Finally, we collect data through \href{https://www.wjx.cn}{WJX}, a professional survey platform with nationwide coverage. After another round of quality control and DSL annotation, \textit{Human-1000} test subset is built. 
Subsequent scaling through \href{https://www.wjx.cn}{WJX} (survey platform) yielded the \textit{Human-1000} test set after analogous quality control and DSL annotation.

\section{Benchmark Characteristics}
This section analyzes the challenges instantiated by ChinaTravel, rooted in authentic human requests and central to real-world applications yet overlooked by prior travel planning benchmarks. 

% \begin{wrapfigure}[10]{r}{0.6\columnwidth}
\begin{figure}[t]
% \centering
\centering
\begin{subfigure}[t]{0.49\linewidth}
    \centering
    \includegraphics[width=.49\linewidth]{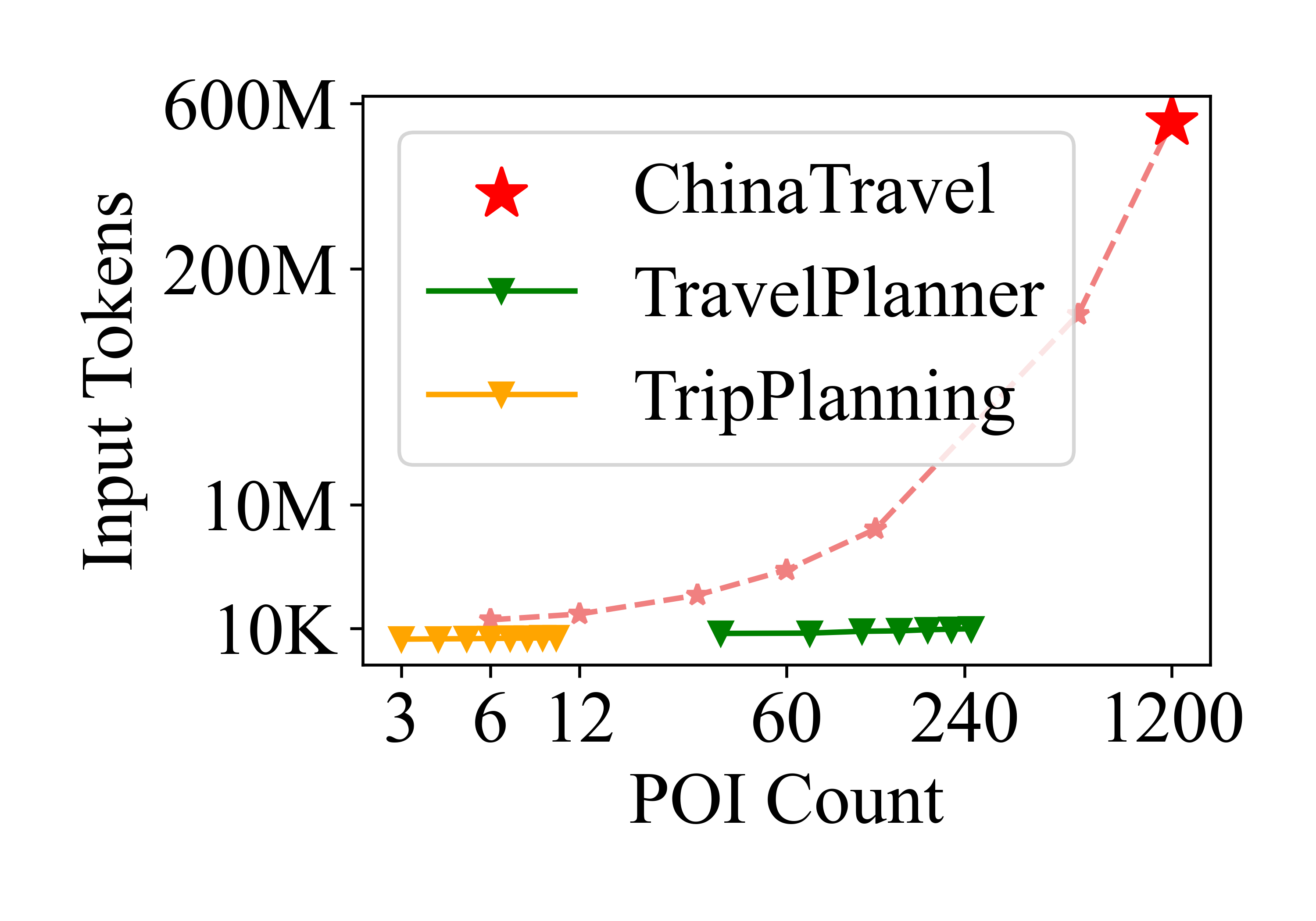}\vspace{-.1in}
    \includegraphics[width=.49\linewidth]{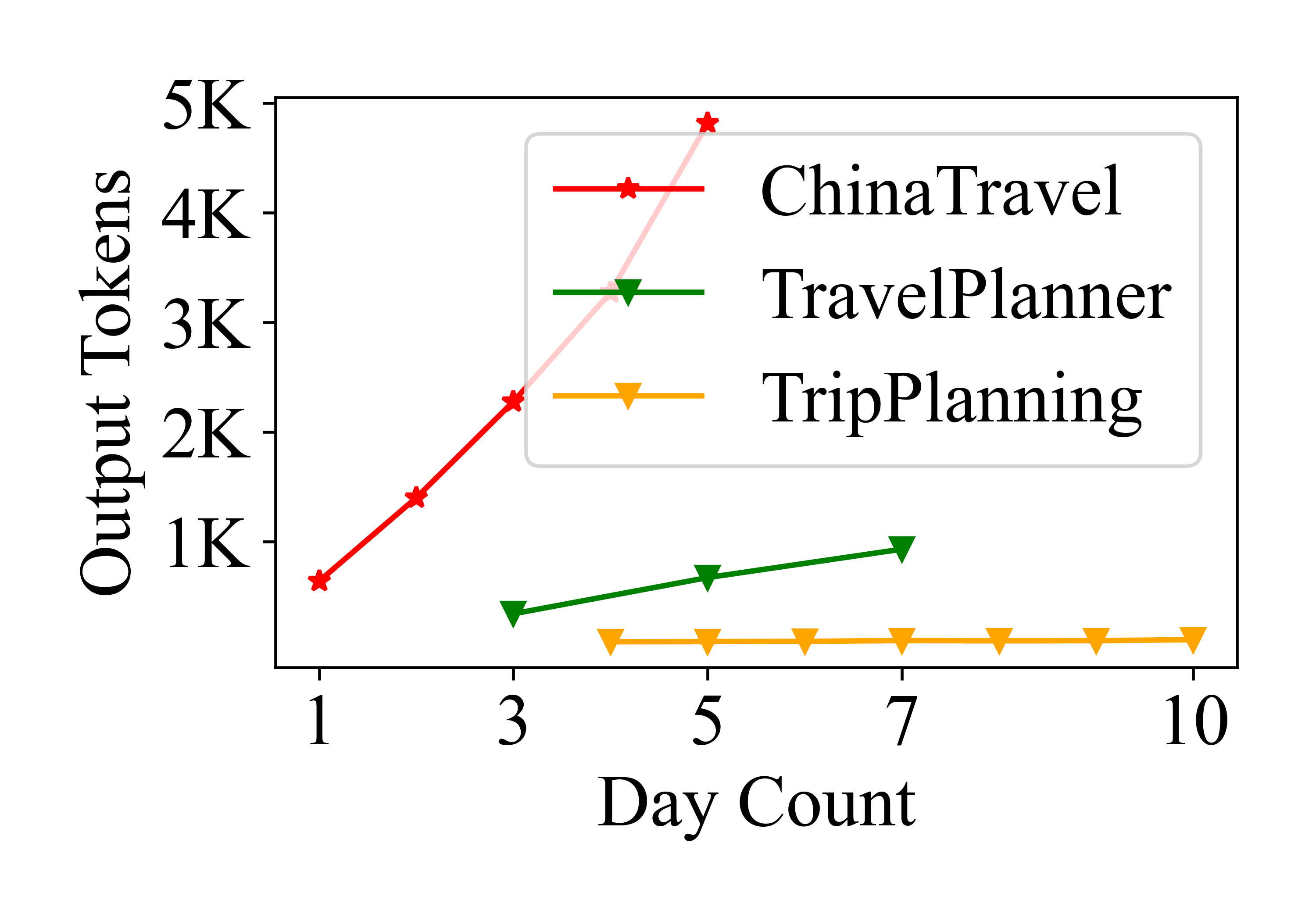}
    \caption{Token count across different benchmarks.}
    \label{fig_token}
\end{subfigure}
\begin{subfigure}[t]{0.49\linewidth}
    \centering
    \includegraphics[width=.49\linewidth]{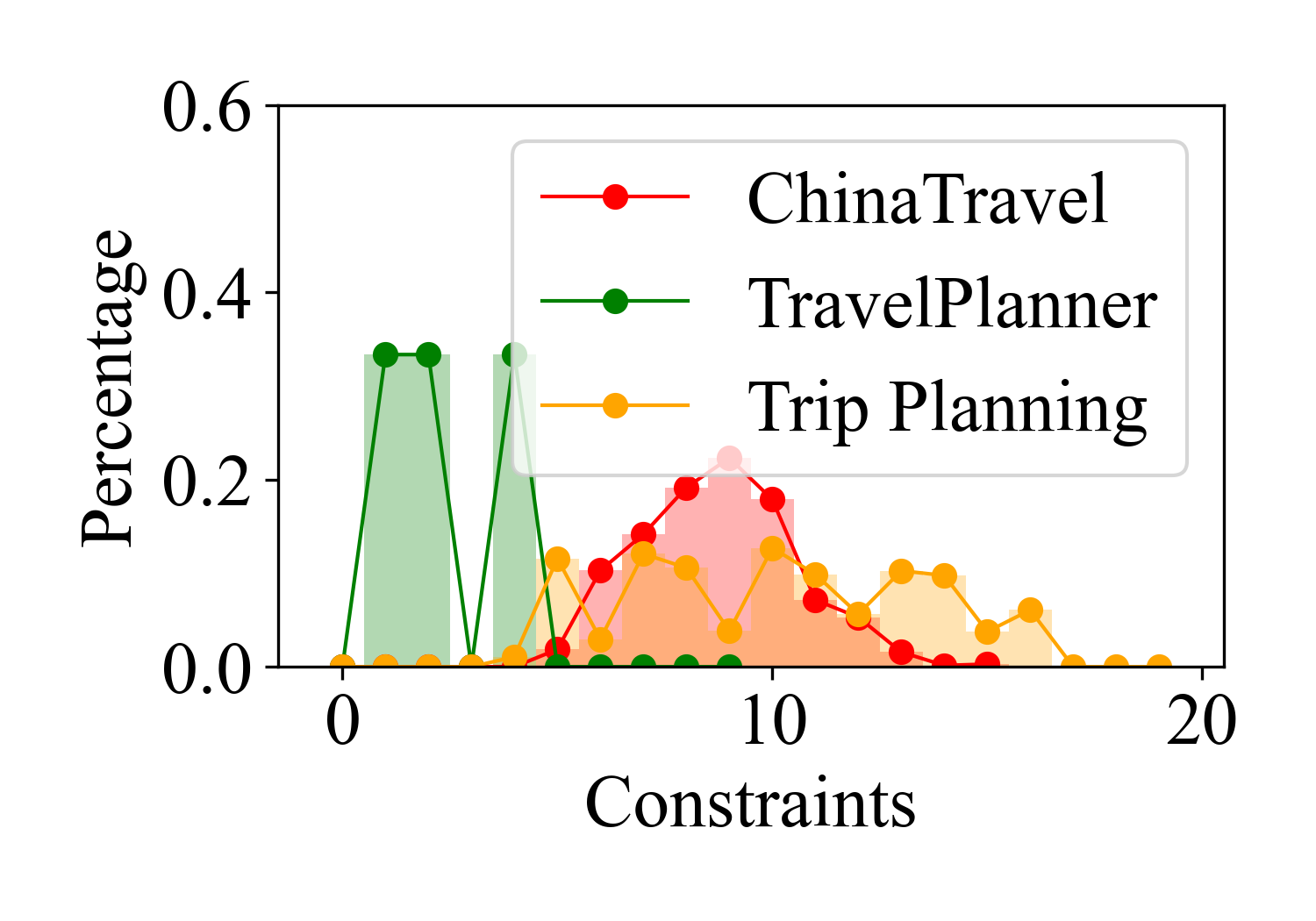}\vspace{-.1in}
    \includegraphics[width=.49\linewidth]{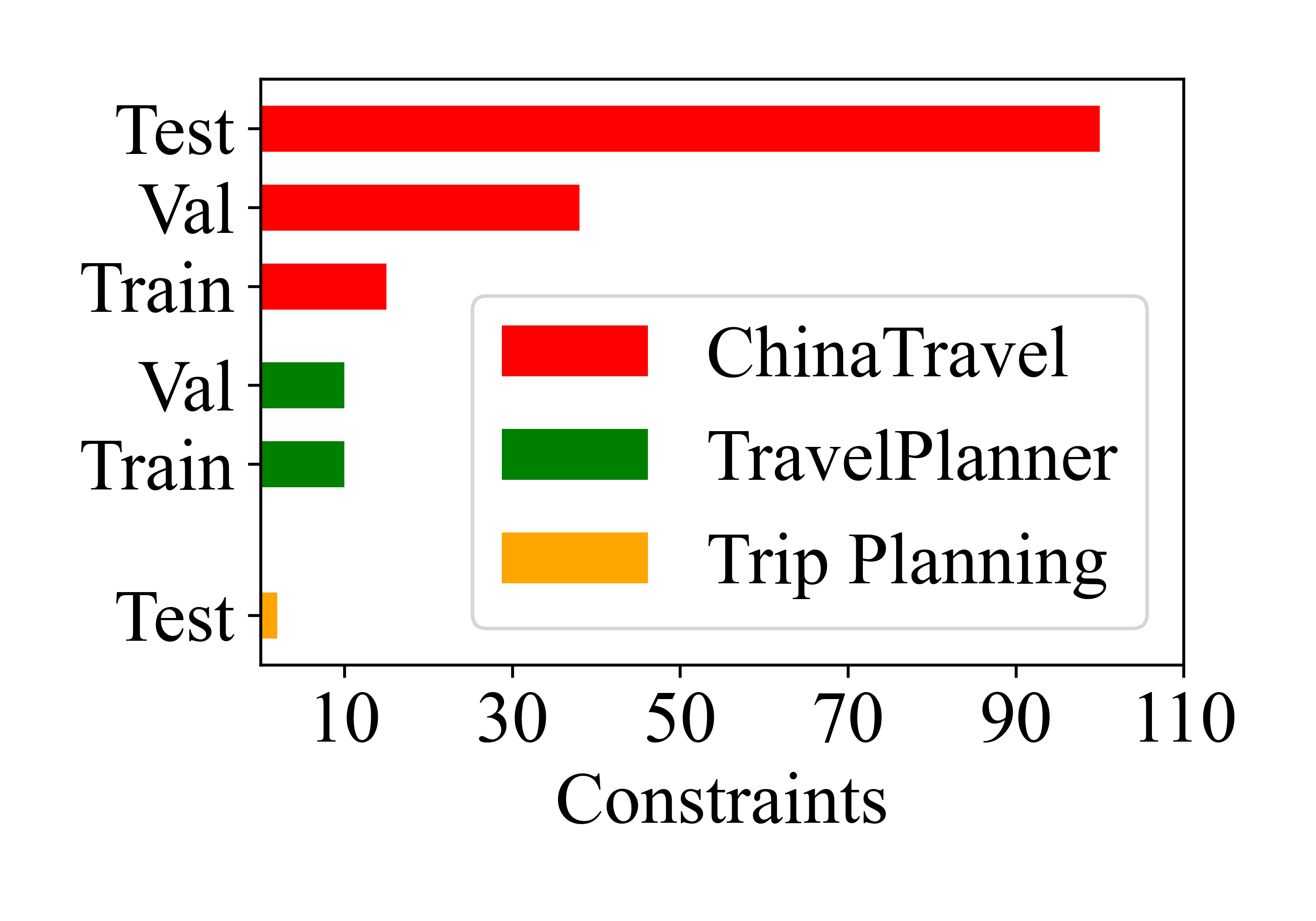}
    \caption{Constraints across different benchmarks.}
    \label{fig_constriants}
\end{subfigure}
\caption{(a) ChinaTravel's fine-grained spatiotemporal planning demands extremely larger input/output text volumes than existing benchmarks, posing fundamental challenges to text-wise planning. \\(b) ChinaTravel's authentic requirements, with combinatorial scalable constraints formulation, systematically surpasses conventional closed-form benchmarks in diversity and openness.}
\vspace{-.1in}
\label{img_tokencount}
\end{figure}

\paragraph{Context-Rich Long-Horizon Planning.} 
ChinaTravel poses unprecedented contextual complexity % in travel planning 
compared to existing benchmarks, TripPlanning~\citep{zheng2024natural} and TravelPlanner~\citep{TravelPlanner}. As quantified in Fig.~\ref{fig_token}, (1) Processing over 1,200 candidate POIs per query (4$\times$ TravelPlanner's 244 max, 120$\times$ Trip Planning's 10) with detailed square-order transportation. 
(2) Generating 540M contextual tokens from dense POI networks, surpassing both DeepSeek-V3 (64K) and GPT-4o (128K) capacities, even aggressive 6-POI downsampling retains 40K tokens (Fig.~\ref{fig_token}). 
(3) Producing 4.8K output tokens for 5-day plans, versus 0.9K (TravelPlanner's 7-day) and 0.5K (Trip Planning's 30-day). 
These findings necessitate a paradigm shift: the traditional single-pass text generation approach proves inadequate for such ultra-long-horizon planning tasks~\citep{ye2025longproc}. Effective solutions may require agents to adopt human-like hierarchical decomposition or symbolic planning techniques, % iteratively 
executing subtasks to achieve final %planning 
objectives through % multi-step 
sequential 
decisioning.

\begin{figure}[t]
\centering

\vspace{-.3in}
\hspace{-.1in}
\begin{subfigure}[t]{0.2\linewidth}
    \centering
    \includegraphics[width=\linewidth]{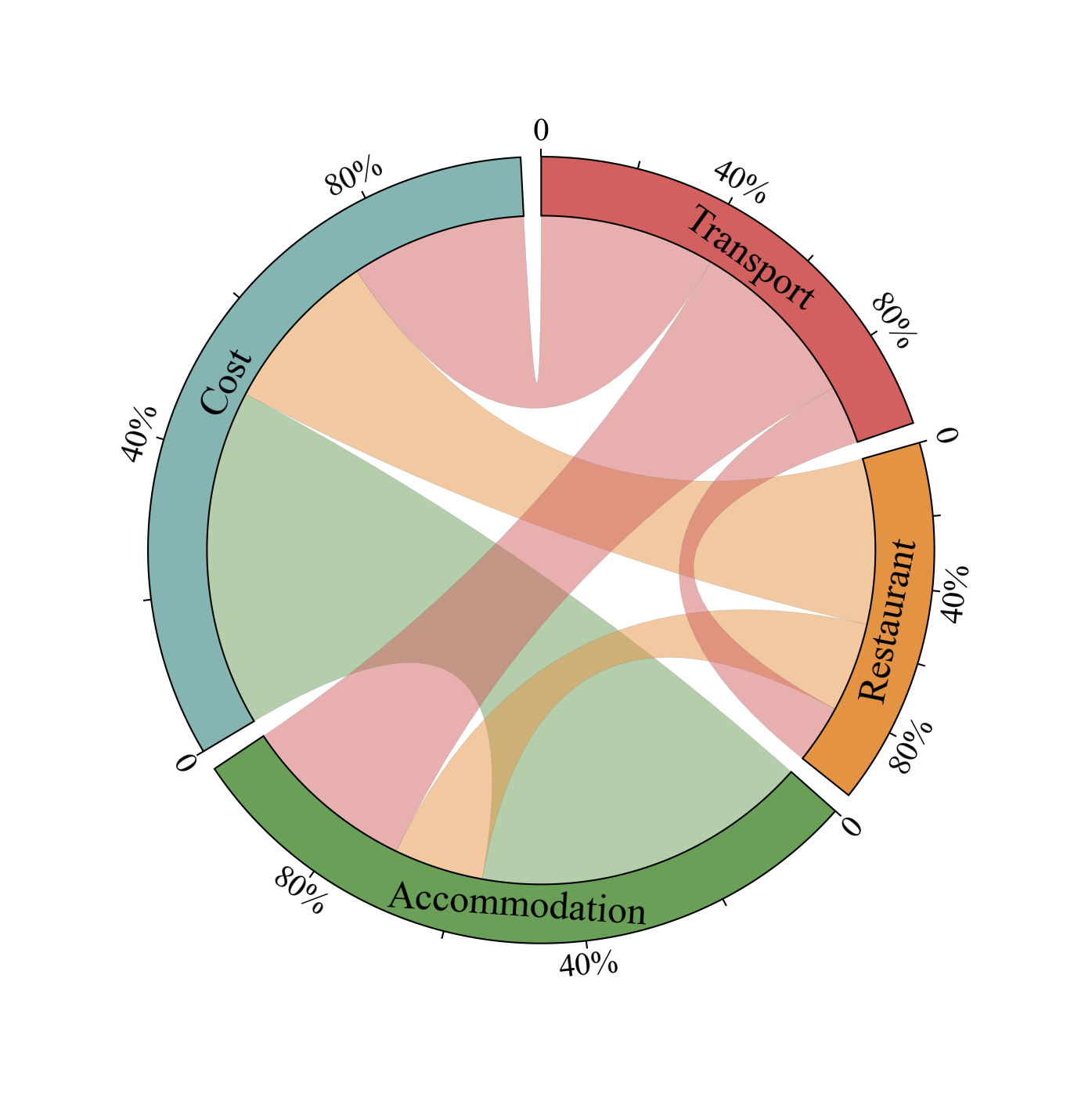}
    % \vspace{-.2in}
    \caption{TravelPlanner}
    \label{tpco}
\end{subfigure}
\hspace{-.1in}
\begin{subfigure}[t]{0.2\linewidth}
    \centering
    \includegraphics[width=\linewidth]{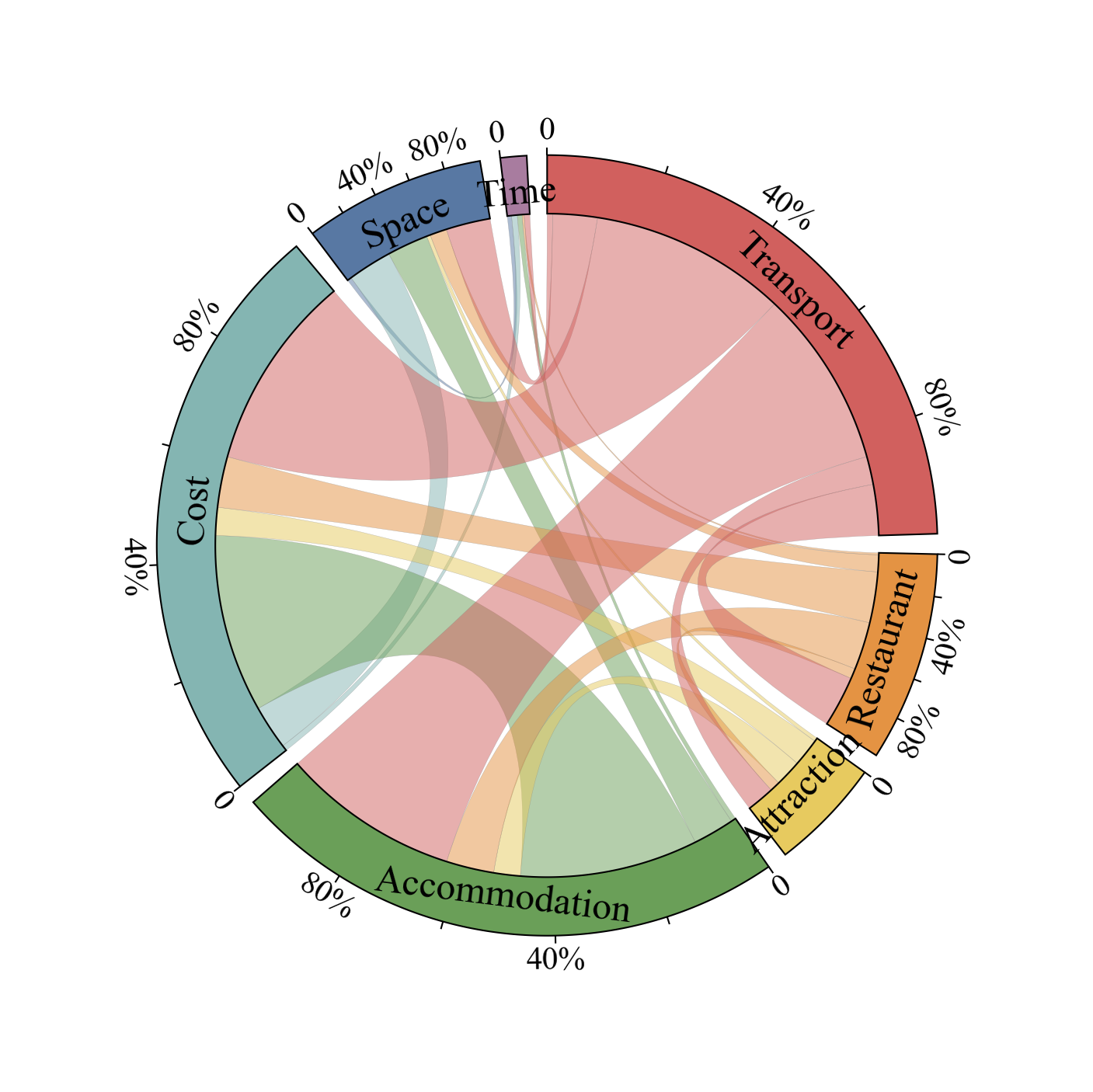}
    % \vspace{-.2in}
    \caption{ChinaTravel}
    \label{ctco}
\end{subfigure}
\begin{subfigure}[t]{0.49\linewidth}
\includegraphics[width=.48\linewidth]{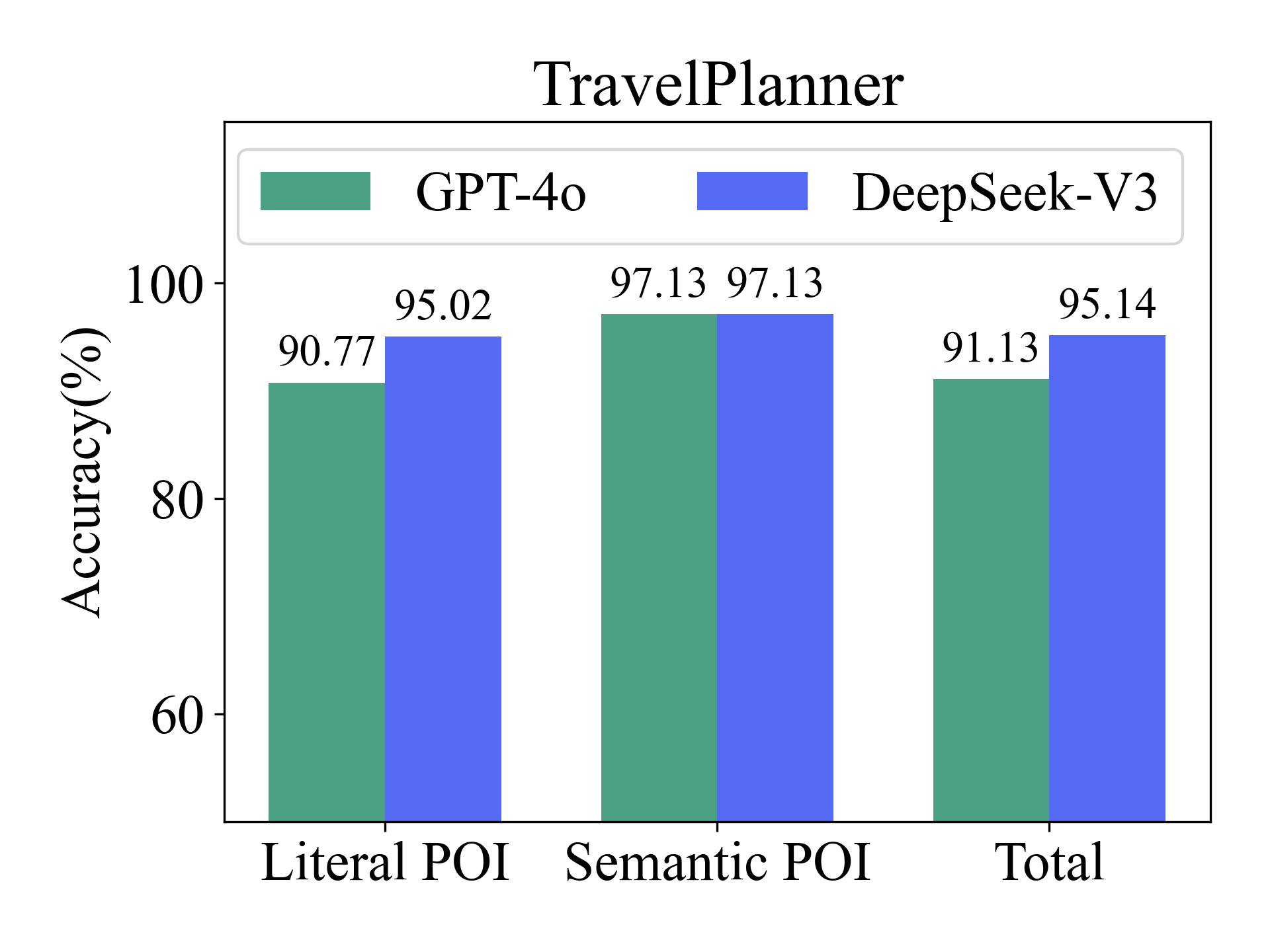}
\includegraphics[width=.48\linewidth]{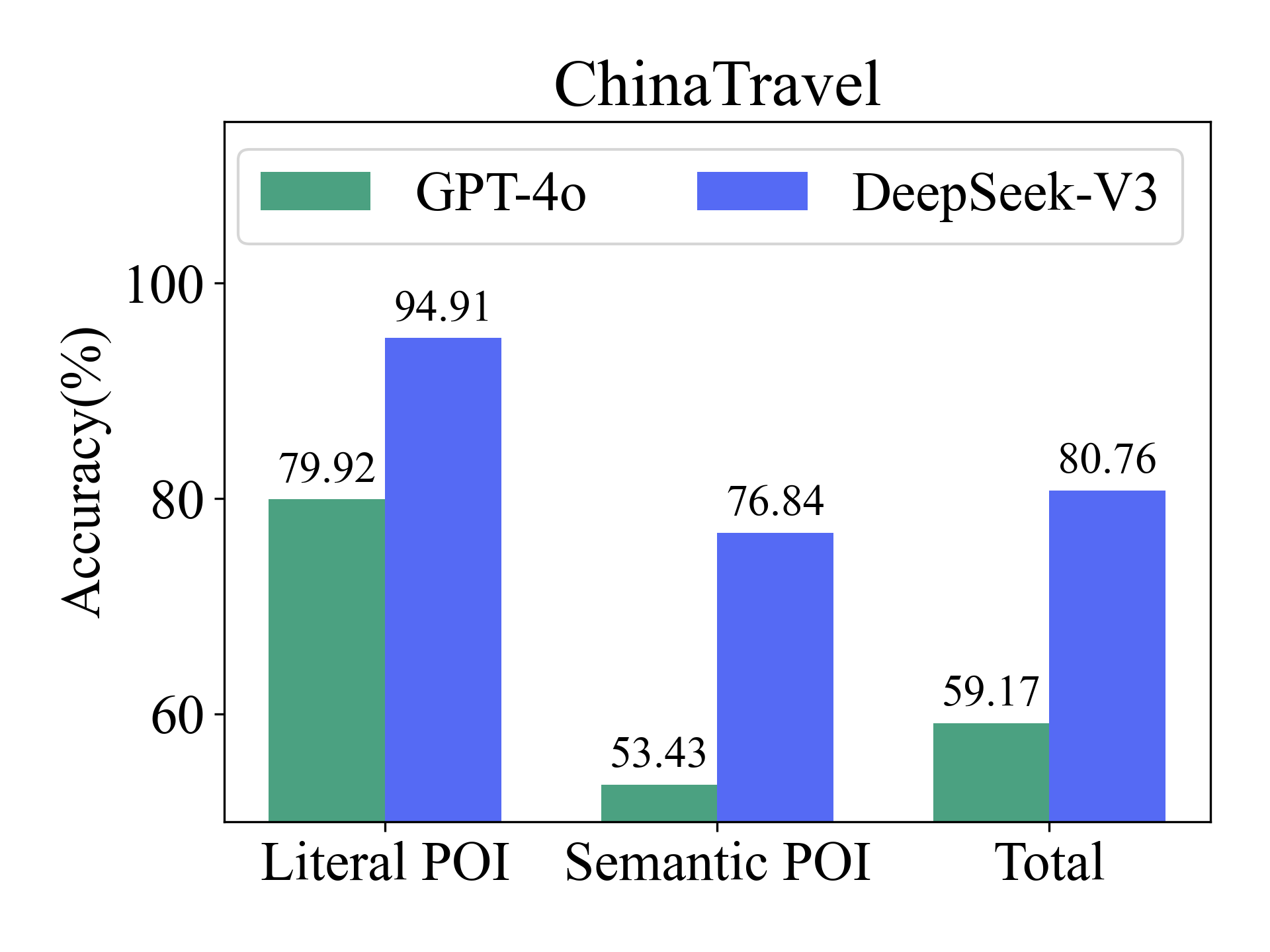}
    % \vspace{-.05in}
\caption{Intent Grounding}
\label{poi_reasoning}
\end{subfigure}
\caption{Co-occurrence distribution of differnt constraints on TravelPlaner (a) and ChinaTravel's Human1000 (b). (c) The unsatisfactory performance of advanced LLMs on the auxiliary task, intent grounding, reveals the challenges of open contextual grounding in ChinaTravel's dataset.}
\label{img_reasoning}
\end{figure}

\paragraph{Diversity and Openness of Travel Requirements.} 
ChinaTravel surpasses TravelPlanner and TripPlanning in diverse requirement modeling. 
As Fig.~\ref{fig_constriants} shown: (1) Constraint volume: ChinaTravel exhibits approximately Gaussian distribution (6-12 constraints per query) versus TravelPlanner's simplicity bias ($\leq$5 constraints) and TripPlanning's limited diversity (allowing up to 16 constraints but spanning only two types). (2) Combinatorial capacity: TravelPlanner's atomic constraints yield only 10 combinations, while ChinaTravel scales exponentially from 15 (synthetic) to 100 types (human1000 test), including 85 novel constraints formulated through Tab.~\ref{tab_dsl}'s compositional system. 
% Analysis of constraint counts per query reveals distinct patterns: ChinaTravel demonstrates a Gaussian-style distribution shape, concentrating within 6-12 constraints (Fig. 3c). 
% In contrast, TravelPlanner's synthetic queries show heavy bias towards simpler tasks with less than 5 constraints. While the TripPlanning benchmark accommodates up to 16 concurrent requirements, its constraint diversity remains severely limited, only 2 types identified (Fig. 3d). 
% While TravelPlanner accounts for multi-dimensional factors like cuisine preferences and budget ranges, its constraint definition methodology remains fundamentally restricted by predefined atomic concepts - a methodological limitation that ultimately yields merely 10 combinatorial types. This starkly contrasts with ChinaTravel's compositionally generalizable system, enables exponential combinatorial possibilities, evidenced by constraint type growth from 15 (synthetic) to 38 (human150 validation) and ultimately 100 types (human1000 test). 
% Crucially, ChinaTravel's human1000 test set contains 88 novel constraints not observed in either easy or human150 subsets, which are all expressible through basic concepts as established in Section 2.2 and could be validated against the standardized output format. 
% This constraint distribution pattern is quantitatively illustrated in Figure 3 through comparative histogram analysis. 
We further investigate co-occurrence of constraint types within individual queries, we categorize basic concepts in our DSL into seven clusters as visualized in Fig.~\ref{ctco}. 
In ChinaTravel, the co-occurrence distribution follows Zipf's law~\citep{adamic2002zipf} with a characteristic long-tail pattern, contrasting sharply with TravelPlanner (Fig.~\ref{tpco}), whose synthetic data demonstrates relatively uniform frequencies.  % \begin{figure}[t]
% \begin{wrapfigure}[15]{r}{0.6\columnwidth}
% \centering
% \vspace{-.15in}
% \begin{subfigure}[t]{0.49\linewidth}
%     \centering
%     \includegraphics[width=\linewidth]{imgs/tp_chord_diagram.png}
%     \caption{TravelPlanner}
%     \label{tpco}
% \end{subfigure}
% \begin{subfigure}[t]{0.49\linewidth}
%     \centering
%     \includegraphics[width=\linewidth]{imgs/chord_diagram.png}
%     %\vspace{-.1in}
%     \caption{ChinaTravel's Human1000}
%     \label{ctco}
% \end{subfigure}
% \vspace{-.05in}
% \caption{Co-occurrence patterns of constraints.}
% % \vspace{-.5in}
% \label{long_tail}
% % \end{figure}
% \end{wrapfigure}
We could also find a strong correlation between cost-related constraints and transportation/accommodation requirements, which aligns with common sense %, because transportation and accommodation typically constitute primary cost components.
given that transport and accommodation are primary cost components. 
% \end{wrapfigure}
% All of these natures originates from systematic user studies that methodologically integrate the open-ended and inexhaustible nature of real-world travel requirements into our benchmark. Users continually present novel composite constraint combinations, elevating constraint diversity to unprecedented levels while ensuring full verifiability via the compositional verification logic framework established in Section 3.2. This characteristic poses significant challenges for travel agents in achieving human-like composite logical reasoning capabilities. 
These characteristics stem from systematic user studies that integrate the evolving, open-ended nature of travel requirements into our benchmark. 
Users continually introduce novel composite constraints, making it impossible to exhaustively enumerate all possibilities during development. 
% ChinaTravel remains verifiable over this ever-expanding requirement space under our DSL system, thereby posing fundamental challenges for agents aiming at human-like composite reasoning. 
By preserving scalable verifiability through our compositional DSL design, ChinaTravel can embrace an evolving requirement space, thereby systematically revealing and rigorously evaluating open-world challenges of travel planning.
% to human-like composite reasoning. 

\paragraph{Contextual Grounding for Implicit Intent.} 
\label{open_reasoning}
% From human queries, we could find that travel requirements often exhibit contextual ambiguity not directly aligned with predefined database attributes. 
From human queries, we observe that travel intent is often expressed implicitly, leading to contextual ambiguity that is not directly aligned with predefined database attributes. 
For example, when users express intent for ``local cuisine", which contextually maps to \emph{$<$Benbang cuisine$>$} in Shanghai versus \emph{$<$Beijing cuisine$>$} in Beijing. Another representative case involves users specifying ``traveling with children who cannot eat spicy food", requiring agents to logically exclude Sichuan and Chongqing cuisines from restaurant selections, beacuse both of them are well-known as their spaicy style. 
These observations arise the necessity for % real-world 
travel agents to conduct contextual grounding that bridges arbitrary user expressions with verifiable symbolic semantics in databases, a % evaluation capability 
critical challenge 
inadequately supported by existing benchmarks% with synthetic queries
~\citep{zheng2024natural, TravelPlanner}. % like TravelPlanner. 
To systematically investigate this challenge, we designed a auxiliary task, Intent Grounding. % , within ChinaTravel. 
It involves replacing all explicit POIs in DSL-defined constraints with a \emph{$<$placeholder$>$} tag, requiring LLMs to complete masked-DSL sentences through contextual grounding. This simplified formulation isolates POI inference from full DSL generation. We further categorize POIs as Literal (explicitly mentioned in user queries) or Semantic (requiring cultural/contextual inference). Quantitative analysis shows 78.4\% of DSL statements from Human1000 contain Semantic POIs needing contextual grounding, contrasting sharply with TravelPlanner's 5.4\% rate that predominantly requires literal mapping. 
Experimental results from DeepSeek-V3 and GPT-4o are shown in Fig.~\ref{poi_reasoning}. 
Both models achieve the accuracy over 90\% on TravelPlanner, where semantic POIs follow simplistic synthetic patterns. However, on ChinaTravel's authentic Semantic POIs, performance significantly declines (DS: $94\%\rightarrow76\%$, GPT: $79\%\rightarrow53\%$). % DeepSeek-V3 maintains 80.76\% overall accuracy, potentially benefiting from Chinese corpus advantages, yet still demonstrates substantial room for improvement in contextual reasoning. 
This performance gap underscores the ChinaTravel's contextual grounding challenges for real-world travel planning. 

\section{Empirical Study}
\label{sec_exp}

\textbf{LLMs.} We evaluate the state-of-the-art LLMs, \imgintext{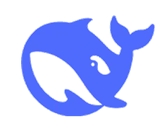}\href{https://www.deepseek.com/}{DeepSeek-V3}, \imgintext{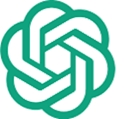}\href{https://openai.com/api/}{OpenAI GPT-4o}, recognized for their world-leading performance. We also examine the open-source LLMs, \imgintext{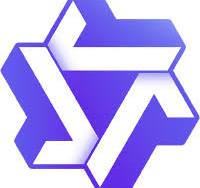}\href{https://qwenlm.github.io/blog/qwen3/}{Qwen3-8B}, \imgintext{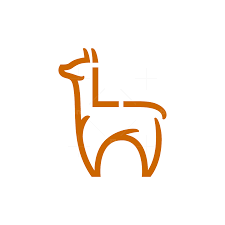}\href{https://ai.meta.com/research/publications/the-llama-3-herd-of-models/}{Llama3.1-8B}, and \imgintext{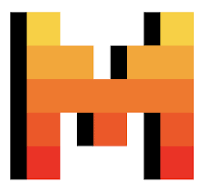}\href{https://mistral.ai/news/announcing-mistral-7b}{Mistral-7B}, selected based on their computationally efficient 7B/8B scales, which enables practical deployment in resource-constrained academic computing environments. \blue{We also report the additional results with large reasoning models, like DeepSeek-R1, at App.~\ref{results_lrm}}. 

\noindent\textbf{Metrics.} We examine the Delivery Rate (DR), Environmental Pass Rate (EPR), Logical Pass Rate (LPR), and Final Pass Rate (FPR) from~\citep{TravelPlanner}. 
\blue{For EPR and LPR, we report both Micro (the average proportion of satisfied constraints per plan, allowing partial credit) and Macro scores (the percentage of plans that strictly satisfy all constraints). The computation details of all metrics are provided in the App.~\ref{app:metric}. } 
To address potential evaluation biases caused by unrealistic constraint prioritization, e.g., misreporting costs to satisfy budget requirement, we design a novel metric, \textbf{Conditional Logical Pass Rate (C-LPR)}. It assesses the success rate of travel plans that \textbf{\textit{first satisfy environmental constraints}} before meeting logical requirements, thereby ensuring logical validity within realistic contextual boundaries. The introduction of C-LPR provides a more rigorous viewpoint for quantifying meaningful constraint satisfaction. % in travel planning. 
\begin{equation*}
    \textit{C-LPR}\!=\!\frac{\sum_{p\in P} \1_{\textit{passed}(Env, p)}\!\cdot\! \sum_{c\in C_p} \1_{\textit{passed}(c, p)}}{\sum_{p\in P}  |C_p|}
\end{equation*}
$P$ is the plan set, $C_p$ is the constraints set for plan $p$, and passed($c$, $p$) indicates whether $p$ satisfies $c$.

\noindent\textbf{Methods.} 
In this work, we mainly focus on the training-free methods with both pure-LLM-based and neuro-symbolic solutions. % on the ChinaTravel benchmark. 
For the former category, we implement ReAct~\citep{DBLP:conf/iclr/YaoZYDSN023}, a widely-adopted reasoning-and-acting framework, along with its Act-only ablation variant. We exclude Reflexion~\citep{DBLP:conf/nips/ShinnCGNY23} due to its performance being similar to ReAct on the TravelPlanner~\citep{TravelPlanner} and the high economic overhead associated with the larger input token size. 
For neuro-symbolic methods, we assess three frameworks: (1) TTG~\citep{ju2024globe}, which converts natural language requirements into mixed-integer linear programming formulations for solver-based optimization. We adapt their formulation into ChinaTravel. The rapied growth of transformed constraints in TTG becomes computationally prohibitive. To address this, we employ LLMs to extract the most relevant POIs for constraint reduction, with detailed linear constraint formulations and experimental configurations provided in App.~\ref{app_ttg}. 
(2) LLM-modulo~\citep{DBLP:conf/icml/KambhampatiVGVS24, gundawar2024robust}, employing ground-truth symbolic verification to guide iterative LLM self-refinement, which could be regrad as an enhanced variant of Reflexction. To ensure compatibility with mainstream LLMs, we perform POI subsampling within a 64K context window. 
(3) NeSy Planning, extending prior NeSy pipelines~\citep{MITSMT, DBLP:conf/emnlp/PanAWW23, DBLP:conf/nips/YaoYZS00N23, DBLP:journals/corr/abs-2410-03136, xiong2026enhancing} through our DSL enhancements to address complex multi-day, multi-POI itineraries. 

\label{nesy_section}

\subsection{Neuro-Symbolic Planning} \begin{wrapfigure}[15]{r}{0.5\columnwidth} % {r} 表示图片在右侧，0.5\columnwidth 表示半列宽度
    \vspace{-.5in}
   \centering
   \includegraphics[width=\linewidth]{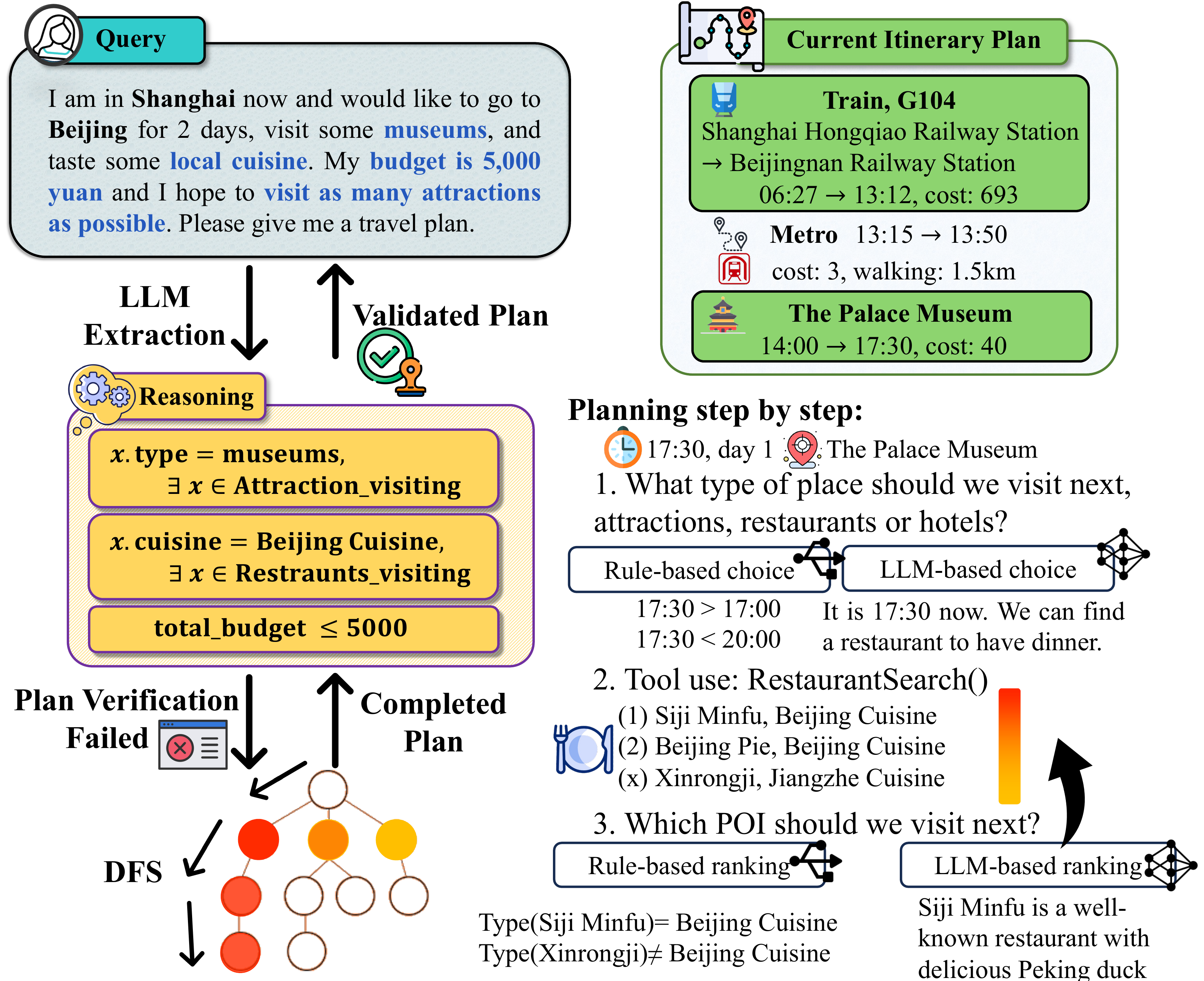} % 图片宽度设置为 wrapfigure 环境宽度
    % \vspace{-.2in}
   \caption{NeSy Planning with search-based solver.}
   \label{fig_dfs}
\end{wrapfigure}
This subsection presents a NeSy solution as a preliminary baseline for ChinaTravel. 
This solution consists of two stages. \textbf{(\MakeUppercase{\romannumeral 1}) NL2DSL translation} transforms natural language queries into logic and preference DSL requirements. 
We use Reflexion~\citep{DBLP:conf/nips/ShinnCGNY23} and a DSL syntax checker to iteratively assist the LLMs (5 rounds in experiments). 
\textbf{(\MakeUppercase{\romannumeral 2}) Interactive search} uses a neuro-symbolic solver to sequentially arrange activities, guided by a symbolic sketch and LLM-driven POI recommendations, generating a multi-day itinerary with DSL validation. % Fig.~\ref{fig_dfs} provide an illustration. 
% Specifically, in the first stage, we leverage reflexion~\citep{DBLP:conf/nips/ShinnCGNY23} and a DSL syntax checker to assist the LLM % in generating logical constraint codes that comply with the DSL 
% through multiple iterations (5 rounds in our experiments). 
% % These codes 
% DSL enables constraint validation of the generated plan in subsequent stages, ensuring it aligns with user requirements. 
% In the second stage, 
% We employ a step-by-step plan generation process using depth-first search. 
% For each step, a pre-defined symbolic program prompts the LLM to recommend the next activity type and then calls the corresponding tool to gather POI information. The LLM assigns priority scores to POIs based on user requirements, such as ranking restaurants by relevance to specific natural language expressions. 
% The symbolic program iteratively fills in the activity details until the itinerary is completed, ending with the return event.  Finally, the completed plan is validated using the DSL code generated in the first stage. 
% The symbolic program iteratively completes the itinerary, ending with validation using Stage 1 DSL code.
% If the plan fails the validation, the process backtracks and continues the search until a constraint-satisfying solution is found. 
If constraints are violated, the process backtracks until a feasible solution is found. 
% Given that some queries are particularly challenging due to the limited feasible plans, we set the maximum runtime for the symbolic sketch from interactive search to 5 minutes per query, excluding the LLM inference time, to ensure a fair comparison across different models. 
To ensure fairness, the symbolic sketch search is limited to 5 minutes per query, excluding LLM inference time. 
To observe the performance across the two stages, we also evaluated the planning results based on the Oracle DSL. In App.~\ref{app_nesy}, we provide the search algorithm's pseudo-code and LLM prompts to enhance reproducibility and support future research. 
% App.~\ref{app_nesy} includes pseudo-code and LLM prompts.% for reproducibility.
\begin{wraptable}[7]{r}{0.5\columnwidth}
    \vspace{-.5in}
   \centering
  \caption{Cost per query across different methods.}
   \label{cost_table}
   \setlength{\tabcolsep}{1.2pt}
   \begin{tabular}{c|cccc}
    \toprule
    Method & \#Input & \#Output  & \makecell{\includegraphics[width=.035\linewidth]{imgs/deepseek.jpg}$(\$)$} & \makecell{\includegraphics[width=.034\linewidth]{imgs/gpt.jpg}$(\$)$} \\ 
    \midrule
    Act & 88K  & 2K & 007 & .219\\
    ReAct (0-shot) & 206K& 3K & .021 & .638\\
    ReAct (1-shot)& 1058K & 4K & .081 & 2.43\\
    LLM-modulo & 362K & 11K & .025 & 1.12\\ 
    NeSy Planning & 467K & 3K & .003 & .306 \\
    \bottomrule
   \end{tabular}
\end{wraptable}
\begin{table*}[t]
    %\small
    \footnotesize
    \centering
   \vspace{.2in}
   % \resizebox{.9\textwidth}{!}{%
   \setlength{\tabcolsep}{1.pt}
   \caption{Main results of different LLMs and planning strategies on the ChinaTravel benchmark.}
   %\\ LLMs: \imgintext{imgs/deepseek.jpg}: DeepSeek-V3, \imgintext{imgs/gpt.jpg}: GPT-4o,\imgintext{imgs/qwen_logo.jpeg}:Qwen3-8B.\imgintext{imgs/llama.png}:Llama3.1-8B,\imgintext{imgs/mistral_logo.png}:Mistral-7B.} %\imgintext{imgs/glm.jpg}: GLM-4PLUS.}
   \begin{tabular}{ll|ccccccc|ccccccc}
      \toprule      
       %&\multirow{2}{*}{\makecell{LLMs}} 
       & &\multirow{2}{*}{DR} & \multicolumn{2}{c}{EPR} & \multicolumn{2}{c}{\makecell{LPR}} & \multirow{2}{*}{\makecell{C-LPR}} & \multirow{2}{*}{\makecell{FPR}} & \multirow{2}{*}{DR} & \multicolumn{2}{c}{EPR} & \multicolumn{2}{c}{\makecell{LPR}} & \multirow{2}{*}{\makecell{C-LPR}} & \multirow{2}{*}{\makecell{FPR}} \\
     \cmidrule(lr){4-7}
       &  & & Mic. & Mac. & Mic. & Mac.  & & & & Mic. & Mac. & Mic. & Mac. \\
      \midrule
      % \multicolumn{8}{c}{\textit{Direct-Prompt}}  \\
      % \hline
      % DeepSeeK-V2.5 \\
      % GPT-4o \\
      % \rowcolor{green!40} 
      % \rowcolor{blue!20}
      & & \multicolumn{7}{>{\columncolor{purple!20}}c}{\textbf{Easy (\#300)}} & \multicolumn{7}{>{\columncolor{orange!20}}c}{\textbf{Human-Val (\#154)}} \\
      \midrule
      % \multicolumn{8}{c}{\textit{Act}}  \\
      % \hline
      \multirow{2}{*}{\makecell{Act}} & \makecell{\includegraphics[width=.03\linewidth]{imgs/deepseek.jpg}} & 70.4 & 49.9 & 0 & 64.6 & 30.6 &  0 & 0 & \multicolumn{7}{c}{-}\\
      & \makecell{\includegraphics[width=.023\linewidth]{imgs/gpt.jpg}} & \textbf{97.5} & 70.8& 0 & 86.8 & 68.6 & 0 & 0 & \multicolumn{7}{c}{-}\\
      \midrule
      % \multicolumn{8}{c}{\textit{ReAct}} \\
      % \hline
      % LLAMA-3.1-13B \\
      \multirow{2}{*}{\makecell{ReAct \scriptsize{(zero-shot)}}} % &\makecell{\includegraphics[width=.023\linewidth]{imgs/glm.jpg}} & 86.5 &32.2 &0 &58.4 &18.5 &0  \\
      & \makecell{\includegraphics[width=.023\linewidth]{imgs/deepseek.jpg}} & 43.3 & 40.8 & 0 & 41.9 & 19.6 & 0 & 0 & 36.4 & 29.5 & 0.65 & 35.2 & 16.2 & 0.38 & 0\\
      & \makecell{\includegraphics[width=.023\linewidth]{imgs/gpt.jpg}} & 95.4 & 48.2 & 0 & 71.3 & 33.0 & 0 & 0 & \textbf{96.1} & 50.5 & 0 & \textbf{72.4} & 32.5 & 0 & 0\\
      % GPT-4o-mini \\
      \multirow{2}{*}{\makecell{ReAct \scriptsize{(one-shot)}}}   & \makecell{\includegraphics[width=.023\linewidth]{imgs/deepseek.jpg}}& 77.5 & 68.3 & 6.00 & 74.1 & 52.3 & 5.77 & 5.33 & 55.2 & \textbf{57.3} & 2.59 & 64.6 & 44.2 & 1.71 & 2.59\\
      & \makecell{\includegraphics[width=.023\linewidth]{imgs/gpt.jpg}}& 94.2 & 68.1 & 0 & \textbf{89.4} & \textbf{70.6} & 0  & 0 & 69.5 & 46.3 & 0 & 63.6 & 46.8 & 0 & 0\\
      \midrule
      \multirow{2}{*}{\makecell{NeSy Planning}} 
      % & \makecell{\includegraphics[width=.023\linewidth]{imgs/glm.jpg}} &90.4 &90.4 &90.4 &88.3 &89.8  &89.8\\
      & \makecell{\includegraphics[width=.023\linewidth]{imgs/deepseek.jpg}} & 75.3 & \textbf{75.3} & 75.3 & 70.4 & 52.6 & 70.4 & 52.6 & 51.9 & 53.2 & 52.5 & 47.0 & \textbf{37.6} & 46.5 & \textbf{37.0}\\
      & \makecell{\includegraphics[width=.023\linewidth]{imgs/gpt.jpg}} & 75.0 &73.6 & \textbf{64.0} & 73.5 & 63.3 & \textbf{61.7} & \textbf{60.6}& 45.4 & 50.1 & 45.4 & 40.9 & 29.8 & \textbf{38.5} & 27.9\\
      & \makecell{\includegraphics[width=.023\linewidth]{imgs/qwen_logo.jpeg}} &  72.3 & 67.0 & 34.0 & 70.4 & 49.6 & 32.6& 28.3 & 42.8 & 47.4 & 42.2 & 36.2 & 27.2 & 34.4 & 25.3 \\
      & \makecell{\includegraphics[width=.023\linewidth]{imgs/llama.png}}& 32.0& 31.9 & 31.3& 29.1& 21.0& 28.3& 21.0& 25.9 & 25.8 & 24.0 & 22.3 & 12.3 & 20.5 & 11.0\\
      & \makecell{\includegraphics[width=.023\linewidth]{imgs/mistral_logo.png}} & 30.3 & 30.3 & 30.3 & 27.6 & 19.6 & 27.6 & 19.6 & 37.6 & 38.2 & 37.6 & 32.7 & 18.8 & 32.2 & 18.8\\
      \midrule
      TTG \scriptsize{(oracle)} & \makecell{\includegraphics[width=.023\linewidth]{imgs/deepseek.jpg}}& \cellcolor{black!10}18.3 & \cellcolor{black!10}21.5 & \cellcolor{black!10}8.66 & \cellcolor{black!10}17.2 & \cellcolor{black!10}15.0 & \cellcolor{black!10}8.23 & \cellcolor{black!10}8.66 & \cellcolor{black!10}9.09 & \cellcolor{black!10}12.8 & \cellcolor{black!10}2.59 & \cellcolor{black!10}7.65 & \cellcolor{black!10}5.19 & \cellcolor{black!10}2.39 & \cellcolor{black!10}1.29\\
      \midrule
      \multirow{3}{*}{\makecell[l]{LLM-Modulo*\\ \scriptsize{(Oracle Verifier)}}} & \makecell{\includegraphics[width=.023\linewidth]{imgs/deepseek.jpg}} &\cellcolor{black!10}48.3 & \cellcolor{black!10}94.5 & \cellcolor{black!10}4.33 & \cellcolor{black!10}58.4 & \cellcolor{black!10}43.6& \cellcolor{black!10}4.11 & \cellcolor{black!10}4.33 & \cellcolor{black!10}61.6 & \cellcolor{black!10}90.2 & \cellcolor{black!10}2.59 & \cellcolor{black!10}75.9 & \cellcolor{black!10}51.2 & \cellcolor{black!10}2.75 & \cellcolor{black!10}2.59 \\
      & \makecell{\includegraphics[width=.023\linewidth]{imgs/gpt.jpg}} & \cellcolor{black!10}91.6 & \cellcolor{black!10}88.2 & \cellcolor{black!10}7.66 & \cellcolor{black!10}\textbf{95.5} & \cellcolor{black!10}\textbf{84.6} & \cellcolor{black!10}7.66 & \cellcolor{black!10}7.00 & \cellcolor{black!10}91.5 & \cellcolor{black!10}87.2 & \cellcolor{black!10}3.24 & \cellcolor{black!10}\textbf{92.9} & \cellcolor{black!10}\textbf{66.2} & \cellcolor{black!10}2.87 & \cellcolor{black!10}3.24 \\
      & \makecell{\includegraphics[width=.023\linewidth]{imgs/qwen_logo.jpeg}} & \cellcolor{black!10}30.0 & \cellcolor{black!10}80.5 & \cellcolor{black!10}0.0 & \cellcolor{black!10}62.7 & \cellcolor{black!10}25.0 & \cellcolor{black!10}0.0 & \cellcolor{black!10}0.0 & \cellcolor{black!10}35.0 & \cellcolor{black!10}75.3 & \cellcolor{black!10}0.0 & \cellcolor{black!10}61.6 & \cellcolor{black!10}19.4 & \cellcolor{black!10}0.0 & \cellcolor{black!10}0.0\\
      & \makecell{\includegraphics[width=.023\linewidth]{imgs/llama.png}} & \cellcolor{black!10}28.6 & \cellcolor{black!10}69.4 & \cellcolor{black!10}0.0 & \cellcolor{black!10}55.2 & \cellcolor{black!10}8.33 & \cellcolor{black!10}0.0 & \cellcolor{black!10}0.0 & \cellcolor{black!10}19.4 & \cellcolor{black!10}74.1 & \cellcolor{black!10}0.0 & \cellcolor{black!10}43.4 & \cellcolor{black!10}5.19 & \cellcolor{black!10}0.0 & \cellcolor{black!10}0.0\\
      & \makecell{\includegraphics[width=.023\linewidth]{imgs/mistral_logo.png}} & \cellcolor{black!10}10.3 & \cellcolor{black!10}90.5 & \cellcolor{black!10}0.0 & \cellcolor{black!10}39.1 & \cellcolor{black!10}9.0 & \cellcolor{black!10}0.0 & \cellcolor{black!10}0.0 & \cellcolor{black!10}3.24 & \cellcolor{black!10}\textbf{92.2} & \cellcolor{black!10}0.0 & \cellcolor{black!10}31.4 & \cellcolor{black!10}4.54 & \cellcolor{black!10}0.0 & \cellcolor{black!10}0.0\\
      \midrule
      \multirow{3}{*}{\makecell[l]{NeSy Planning*\\ \scriptsize{(Oracle Translation)}}} 
      % & \makecell{\includegraphics[width=.023\linewidth]{imgs/glm.jpg}} &90.4 &90.4 &90.4 &88.3 &89.8  &89.8 \\
      & \makecell{\includegraphics[width=.023\linewidth]{imgs/deepseek.jpg}} & \cellcolor{black!10}\textbf{82.6} & \cellcolor{black!10}\textbf{81.7} & \cellcolor{black!10}\textbf{75.0} & \cellcolor{black!10}\textbf{82.2} & \cellcolor{black!10}75.3 & \cellcolor{black!10}\textbf{75.0}& \cellcolor{black!10}\textbf{74.0} & \cellcolor{black!10}\textbf{58.4} & \cellcolor{black!10}59.6 & \cellcolor{black!10}\textbf{57.7} & \cellcolor{black!10}53.8 & \cellcolor{black!10}46.1 & \cellcolor{black!10}\textbf{52.0} &\cellcolor{black!10}\textbf{45.4}\\
      & \makecell{\includegraphics[width=.023\linewidth]{imgs/gpt.jpg}} & \cellcolor{black!10}66.6 & \cellcolor{black!10}66.7 & \cellcolor{black!10}66.0 & \cellcolor{black!10}64.6 & \cellcolor{black!10}63.6 & \cellcolor{black!10}64.6 & \cellcolor{black!10}62.6 & \cellcolor{black!10}52.6 & \cellcolor{black!10}46.9 & \cellcolor{black!10}42.9 & \cellcolor{black!10}47.6 & \cellcolor{black!10}40.9 & \cellcolor{black!10}43.9 &\cellcolor{black!10}40.9  \\
      & \makecell{\includegraphics[width=.023\linewidth]{imgs/qwen_logo.jpeg}} & \cellcolor{black!10}69.3 & \cellcolor{black!10}69.3 & \cellcolor{black!10}59.3 & \cellcolor{black!10}70.2 & \cellcolor{black!10}59.6 & \cellcolor{black!10}59.3& \cellcolor{black!10}57.9  & \cellcolor{black!10}53.2 & \cellcolor{black!10}55.1 & \cellcolor{black!10}54.5 & \cellcolor{black!10}48.0 & \cellcolor{black!10}42.8  & \cellcolor{black!10}47.6 & \cellcolor{black!10}40.9 \\
      & \makecell{\includegraphics[width=.023\linewidth]{imgs/mistral_logo.png}} & \cellcolor{black!10}52.6 & \cellcolor{black!10}52.6 & \cellcolor{black!10}52.6 & \cellcolor{black!10}50.4 & \cellcolor{black!10}45.3 & \cellcolor{black!10}50.4 & \cellcolor{black!10}45.6 & \cellcolor{black!10}40.9 & \cellcolor{black!10}42.8 & \cellcolor{black!10}42.8 & \cellcolor{black!10}37.7 & \cellcolor{black!10}28.5 & \cellcolor{black!10}37.7 & \cellcolor{black!10}27.9 \\
      & \makecell{\includegraphics[width=.023\linewidth]{imgs/llama.png}} & \cellcolor{black!10}33.3&\cellcolor{black!10}33.2&\cellcolor{black!10}32.6&\cellcolor{black!10}32.1&\cellcolor{black!10}32.0& \cellcolor{black!10}31.4& \cellcolor{black!10}32.3&  \cellcolor{black!10}29.2 & \cellcolor{black!10}29.1 & \cellcolor{black!10}26.6 & \cellcolor{black!10}25.4 & \cellcolor{black!10}20.1 & \cellcolor{black!10}23.4 & \cellcolor{black!10}19.4 \\
      \midrule

      & & \multicolumn{7}{>{\columncolor{green!20}}c|}{\textbf{Human-Test (\#1000)}} & \multicolumn{7}{>{\columncolor{green!20}}c}{NeSy Planning* \scriptsize{(Oracle Translation)}}\\
      % \midrule
      % & & & & & & & & &\multicolumn{3}{c}{Input Tokens}&\multicolumn{3}{c}{Output Tokens} & Price\\
    %   \midrule
    %   \multirow{2}{*}{\makecell{NeSy Planning}} 
    %   % & \makecell{\includegraphics[width=.023\linewidth]{imgs/glm.jpg}} &62.3 &62.2 &61.0 &49.6 &42.2  &41.6\\
    %   & \makecell{\includegraphics[width=.023\linewidth]{imgs/deepseek.jpg}} & \\
    %   & \makecell{\includegraphics[width=.023\linewidth]{imgs/gpt.jpg}} & \\
    %   % & \makecell{\includegraphics[width=.023\linewidth]{imgs/qwen_logo.jpeg}} & \\
      \midrule
       \multirow{3}{*}{\makecell{NeSy Planning}} 
    % \multirow{3}{*}{\makecell[l]{NeSy Planning*\\\scriptsize{(Oracle Translation)}}} 
        
      % & \makecell{\includegraphics[width=.023\linewidth]{imgs/glm.jpg}} &90.4 &90.4 &90.4 &88.3 &89.8  &89.8 \\
      & \makecell{\includegraphics[width=.023\linewidth]{imgs/deepseek.jpg}} & \textbf{44.6} & \textbf{44.5} & \textbf{42.6} & \textbf{38.7} & \textbf{23.3}  & \textbf{37.6}& \textbf{23.3}& \cellcolor{black!10}\textbf{60.6} & \cellcolor{black!10}\textbf{60.3} & \cellcolor{black!10}\textbf{59.0} & \cellcolor{black!10}\textbf{53.6} & \cellcolor{black!10}\textbf{32.0} & \cellcolor{black!10}\textbf{52.5} &\cellcolor{black!10}\textbf{31.6} \\
      & \makecell{\includegraphics[width=.023\linewidth]{imgs/gpt.jpg}} &37.3& 37.2 & 35.0 & 30.7 & 11.3& 29.2 & 11.3& \cellcolor{black!10}27.8 & \cellcolor{black!10}27.8 & \cellcolor{black!10}27.1 & \cellcolor{black!10}24.8 & \cellcolor{black!10}12.8 & \cellcolor{black!10}24.4 &\cellcolor{black!10}12.8 \\
      & \makecell{\includegraphics[width=.023\linewidth]{imgs/qwen_logo.jpeg}} & 36.6 & 36.5& 34.6 & 29.6& 6.43& 28.5&6.43 &   \cellcolor{black!10}41.1 & \cellcolor{black!10}41.1 & \cellcolor{black!10}40.6 & \cellcolor{black!10}34.6 & \cellcolor{black!10}13.8  & \cellcolor{black!10}34.2 & \cellcolor{black!10}13.8 \\
     \bottomrule
   \end{tabular}% }
   % \vspace{-.5in}
   \centering\label{tab:main_tab}
\end{table*}

\subsection{Main Results}
Based on the results presented in Tab.~\ref{tab:main_tab} and~\ref{cost_table}, we have the following finding and analyses: 

\noindent\textbf{Pure LLMs struggle in ChinaTravel.} 
The DR evaluates the capability to generate valid JSON travel plans (see Fig.~\ref{overview}). While high DR values indicate that advanced LLMs can produce structurally correct outputs, the near-zero EPR reveals their fundamental limitations in acquiring and strictly adhering to required constraints. 
The sole exception is the DeepSeek, which achieves the $6\%$ EPR and $5\%$ FPR at easy level. 
A plausible explanation is broader training-data coverage for Chinese queries. 
ReAct (one-shot, GPT-4o) excels in Macro LPR but achieves no FPR, suggesting it circumvents constraints via shortcuts. Our C-LPR metric offers a more reliable measure of logical constraints, serving as a supplement to FPR. 
As shown in Tab.~\ref{cost_table}, 
purely neural baselines require large input/output tokens and correspondingly high cost. 
With GPT-4o, the average cost is \$2.43 per query, yet they produce on constraint-satisfying plans. 
Given the substantial cost and their persistently low FPR, further pure-neural variants offer diminishing returns under our budget. We therefore concentrate on NeSy solutions. 

\begin{figure*}[t]
    \centering
    \begin{subfigure}{0.295\textwidth}
       \centering
    \includegraphics[width=\linewidth]{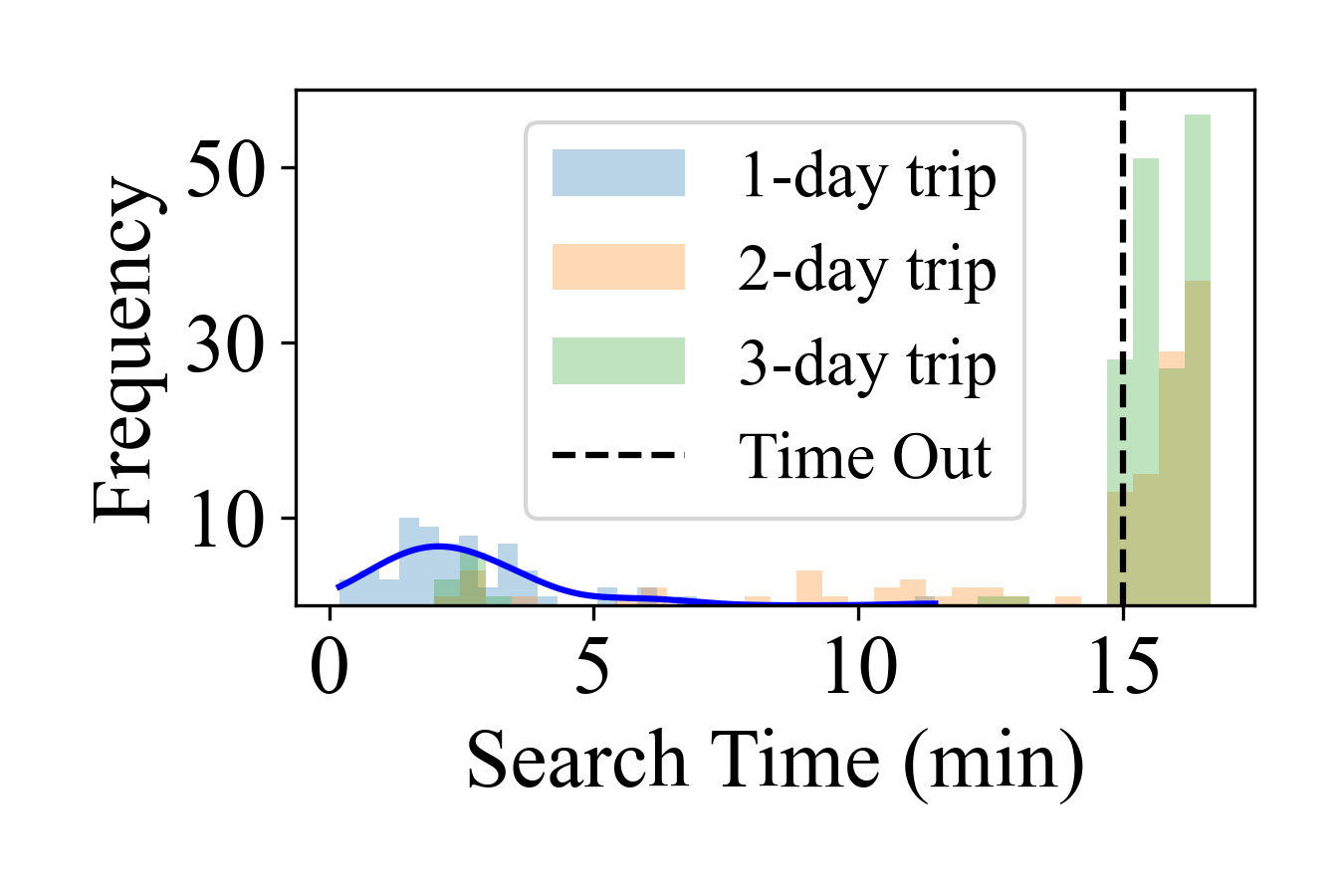}
       \caption{Solve Time of TTG.}
       \label{ttg_time}
   \end{subfigure}
    \begin{subfigure}{0.66\textwidth}
       \centering
    \includegraphics[width=\linewidth]{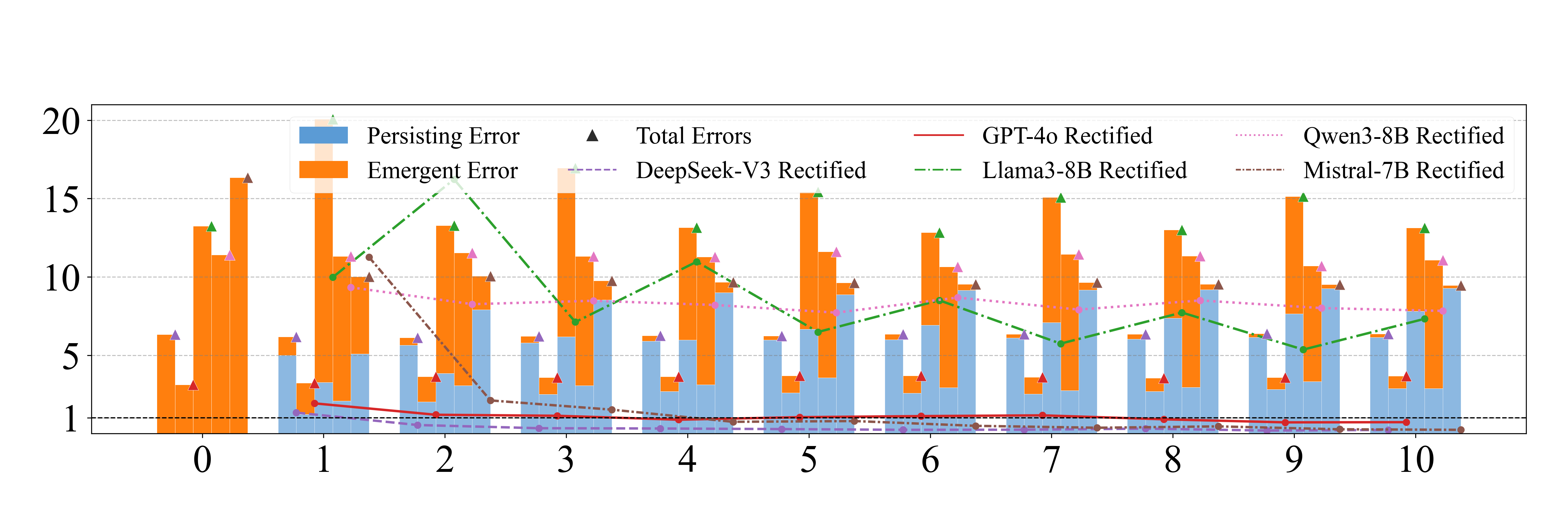}
       \caption{Refinement of LLM-modulo.}
       \label{modulo_error}
   \end{subfigure}
    \caption{(a) The high computational complexity of TTG renders it infeasible for real-world multi-day itineraries. (b) LLM-modulo's error correction declines during iteration, causing emergent errors.}
    \vspace{-.1in}
    \label{challenge_nesy}
\end{figure*}

\noindent\textbf{The Inadequacies of Existing NeSy Approaches}. 
% The fundamental limitation of TTG arises from its computational complexity, where the constraint count scales as \( O(N^3T) \) with \( N \) POIs and \( T \) time windows. Even when subsampling to 22 POIs and discretizing time into 1-hour slots (\( T=24 \)), TTG generates approximately 600,000 constraints for 2-day itinerary. In our main experiments using the SCIP solver from the PuLP package, TTG was allocated a relaxed 15-minute search limitation. However, this configuration yielded only 18\% valid solutions on easy-subet instances, with the FPR further reduced to 8\% due to the solver's pruning heuristics. Fig.~\ref{ttg_time} illustrates the solve time of TTG on 1-3 day itinerary. Within the time limit, solutions were found for merely 23\% for 2-day and 6\% for 3-day itineraries. 
TTG's complexity grows rapidly with the size of the POI candidates and the temporal discretization: the number of constraints scales on the order of \( O(N^3T) \) with \( N \) POIs and \( T \) time slots. Even after subsampling to (\( N=22 \)) and using 1-hour slots (\( T=24 \)), a 2-day instance contains on the order of 600,000 constraints. We run TTG with SCIP solver, allocated a relaxed 15-minute search limitation per query. This configuration yielded only 18\% valid solutions on easy-subet instances, with the FPR further reduced to 8\% due to the solver's pruning heuristics. Fig.~\ref{ttg_time} illustrates the solve time of TTG on 1-3 day itinerary. Within the time limit, solutions were found for only 23\% for 2-day and 6\% for 3-day itineraries. 
LLM-modulo introduces an oracle symbolic verifier to detect constraint violations and feeds back error messages to the LLM for iterative plan revision. Fig.~\ref{modulo_error} depicts the error dynamics over 10 refinement rounds. 
GPT-4o attains the lowest cumulative error ($\mu=3.2 \pm 0.8$), followed by DeepSeek ($\mu=5.1 \pm 1.2$). However, their rectification capacity, quantified by successfully rectified errors per iteration, rapidly decays to $\leq 1$ after 3-5 rounds, indicating diminishing returns from further refinements. 
Smaller models (Qwen3-8B and Llama3-8B) show higher per-step rectification, but also introduce more emergent errors, yielding no significant refinement across iterations. 
% This pattern suggests that while LLM-modulo enables basic constraint feedback from previous travel benchmarks~\citep{zheng2024natural,TravelPlanner}, its effectiveness diminishes for complex multi-day itineraries. 
Taken together, the verifier-feedback loop, effective on earlier travel benchmarks, does not scale well to complex multi-day itineraries: after a few rounds, refinement stalls while additional iterations incur extra cost and latency. 

\noindent\textbf{Nesy Planning provides a promising solution}. 
Our NeSy Planning method orchestrates tool use and planning via symbolic programs while utilizing LLMs to parse natural-language requirements and prioritize POIs. By decoupling understanding (flexible natural language handling), planning (DSL-guided backtracking/verification) and actioning (precise tool execution), it improves adaptability and adherence to constraints in context-rich long-horizon settings. Across the evaluated subsets, it outperforms TTG and LLM-modulo, even without the help of oracle translation. 
% Among the evaluated models, DeepSeek-V3 achieves the best performance. 
With the DeepSeek as backend, it achieves FPRs of 52.6\%, 37.0\% and 23.3\% on three subsets, highlighting the effectiveness of NeSy solutions for travel planning with complex constraints. 
% These gains on human subset, also suggests the improved robustness to constraint generalization in compositionally novel situations. 
On the \emph{human-val} and \emph{human-test} subsets, these gains persist, suggesting robustness to unseen constraint compositions. 

% Another potential explanation is that the model is developed by a Chinese company. As a result, it has been trained on a vast amount of Chinese-language data. This extensive exposure to Chinese text has enabled it to perform exceptionally well in our Chinese travel planning scenarios, giving it advantages over others. 

% \begin{figure}[t]
%   \centering
%   \includegraphics[width=\linewidth]{imgs/err_flex.png}
%   \caption{Syntax errors across reflexion rounds $\tau$.}
%   \label{Syntaxerrors}
% \end{figure}

\begin{figure*}[t]
    \centering
    \begin{subfigure}{0.48\textwidth}
       \centering
    \includegraphics[width=\linewidth]{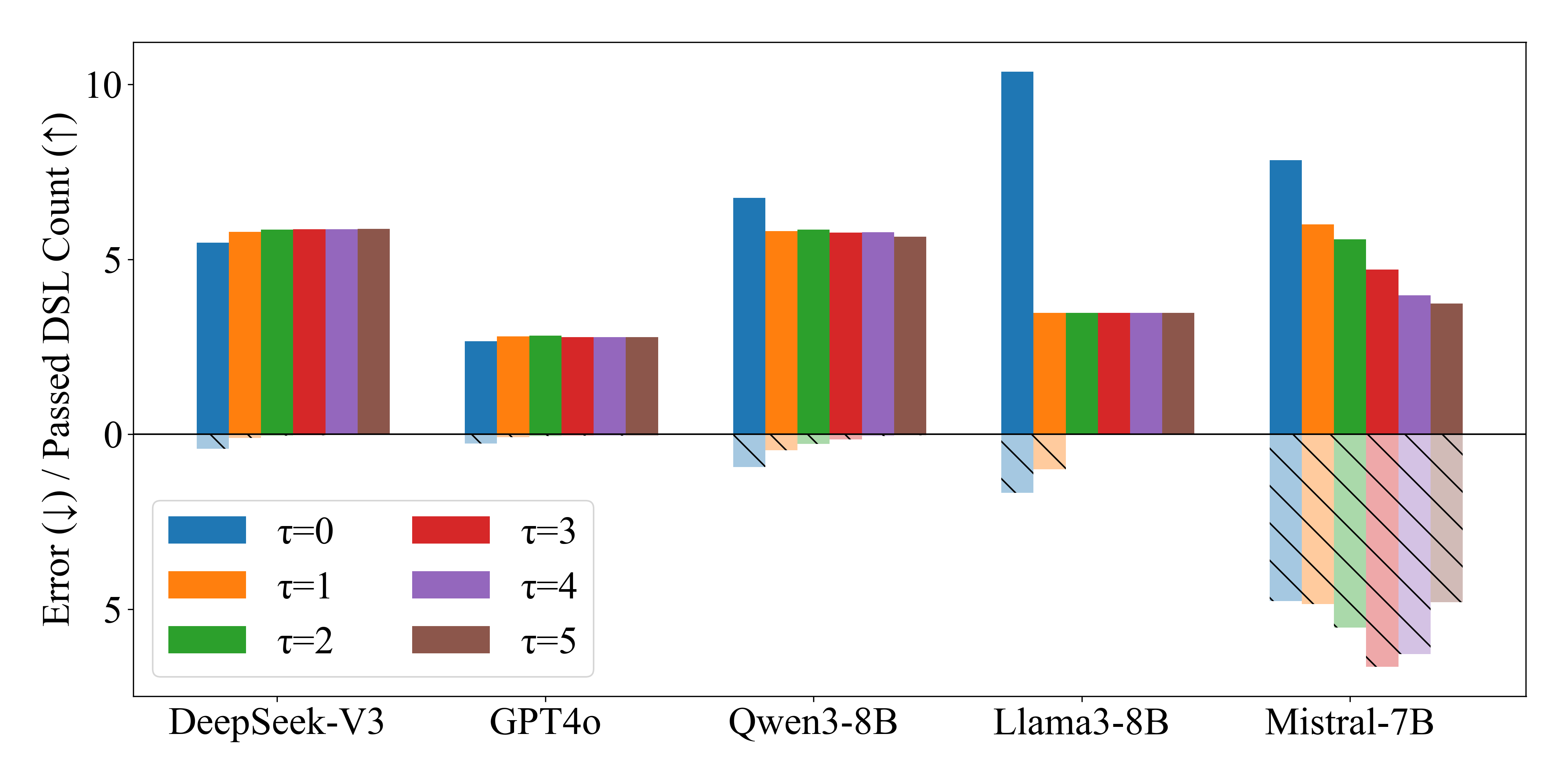}
       \caption{Syntax error of DSL translation.}
       \label{syntax_error}
   \end{subfigure}
    \begin{subfigure}{0.48\textwidth}
       \centering
    \includegraphics[width=\linewidth]{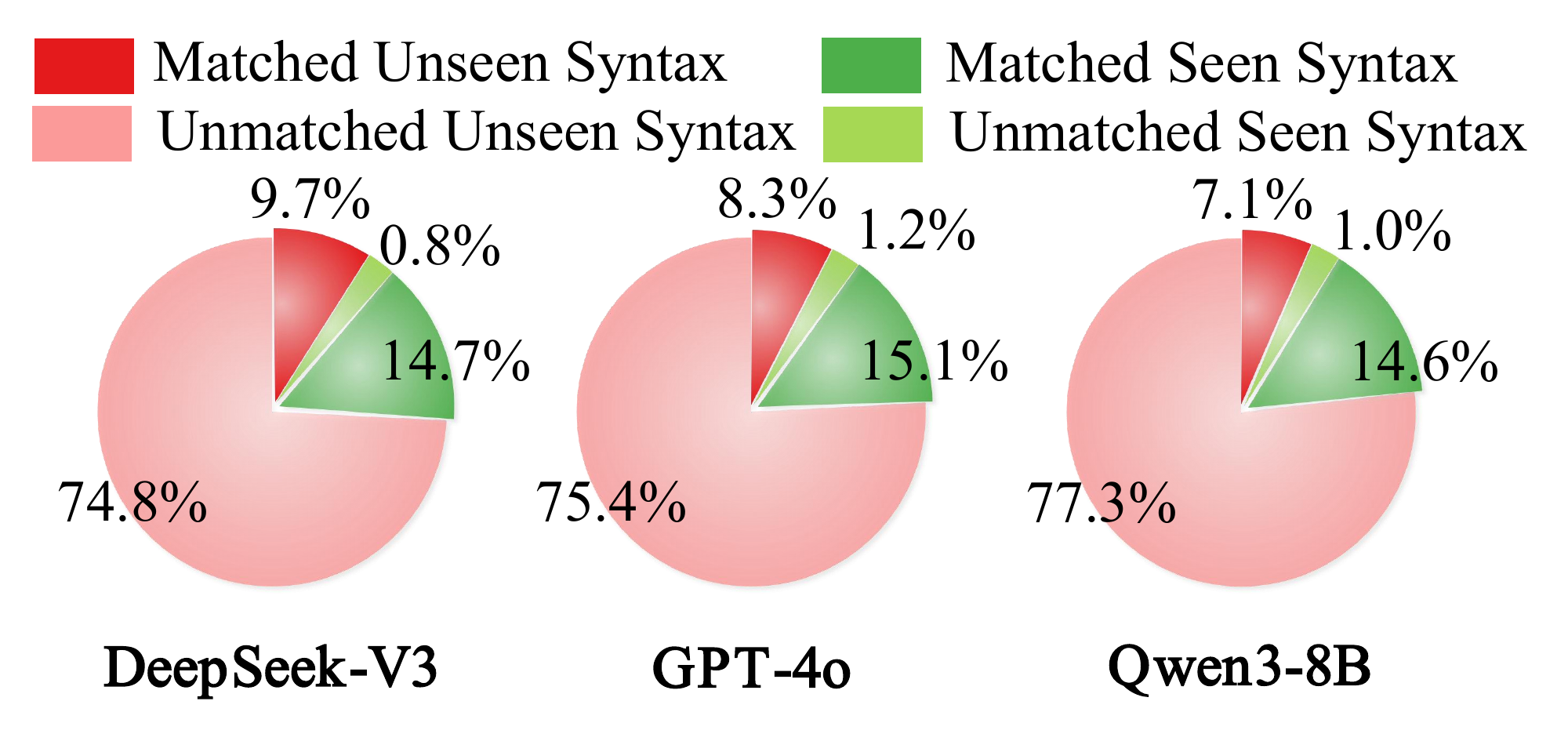}
       \caption{Syntax generalization of DSL translation.}
       \label{generalization_error}
   \end{subfigure}
    \caption{Challenges in NL2DSL translation.}
    \vspace{-.2in}
    \label{challenge_NL2DSL}
\end{figure*}

\noindent\textbf{Challenges Persist for Nesy Planning}. 
% The performance gap between the two modes (standard and oracle) highlights that DSL translation is a crucial step in NeSy planning.  
The performance gap between standard and oracle modes underscores the importance of DSL translation in NeSy planning. 
Inadequate translations may result in plan searches failing to meet user requirements, while incorrect translations can misguide the search, making feasible solutions unattainable. 
We conclude with three challenges and provide the corresponding cases in the Fig.~\ref{challenge_NL2DSL_2}. \textbf{(1) DSL Syntax Compliance}: 
As evidenced in Fig.~\ref{syntax_error}, while the reflexion process with syntactic checking effectively reduces parser-level errors, it inadvertently triggers constraint dropping across multiple LLMs. % Specifically, Qwen3-8B, Llama3-8B, and Mistral-7B exhibit progressive reduction in extracted DSL constraints during iterative refinement. 
For Qwen3-8B, Llama3-8B, and Mistral-7B, the number of DSL constraint clauses decreases across iterations. 
Notably, GPT-4o generates approximately two fewer constraints than DeepSeek-V3 on average under the same loop. 
Although this conservative strategy enables rapid error convergence (achieving zero detected errors within limited iterations), it risks oversimplifying constraint specifications, critical dependencies may be prematurely discarded, ultimately yielding solutions that fail to satisfy complex requirements. 
This conservative pattern often drives fast convergence to zero parser errors within a few rounds, but may prune required constraints and under-specify the plan, leaving the outcome cannot satisfy complex queries. 
This observed conservatism on constraint extraction likely contributes to GPT-4o's underperformance on Human-154 and Human-1000 compared with DeepSeek-V3. 
\textbf{(2) Contextual Grounding}: % Although DeepSeek-V3 exhibits relatively fewer syntax errors in translation, it still struggles with  diverse queries and context-dependent meanings. 
% understand diverse human queries. The variability in language, including context-dependent meanings and open-ended expressions, complicates mapping these descriptions to predefined concepts. 
In the Sec.~\ref{open_reasoning} we have provided a quantitative analysis for this challenge. 
Overcoming this might require domain-adaptive training, enabling LLMs to better interpret implicit user intent.
% In App.~\ref{app_concept_func}, more examples are provied for this challenges. 
% For instance, when a user requests 
% % \begin{CJK}{UTF8}{gbsn}本地菜\end{CJK} 
% (local cuisine), GPT-4o maps it to 
% % \begin{CJK}{UTF8}{gbsn}本帮菜\end{CJK}
% , ignoring the logical connection that in Beijing, it should align with 
% % \begin{CJK}{UTF8}{gbsn}北京菜\end{CJK} 
% (Beijing cuisine).
% As shown in the second case in Fig.~\ref{challenge_NL2DSL}, a user requests to try local cuisine. The LLM extracts \begin{CJK}{UTF8}{gbsn}本帮菜\end{CJK} as a similar expression for \begin{CJK}{UTF8}{gbsn}本地特色\end{CJK}, overlooking the intermediate logical connection that, given the destination is Beijing, it should explicitly map to \begin{CJK}{UTF8}{gbsn}北京菜\end{CJK} available in the database.
\textbf{(3) Unseen Concept Composition}: 
Real-world requirements are diverse and open-ended, so it is unrealistic to expect models to encounter all possible needs during development. 
A more realistic way is to emulate human reasoning by generalizing existing knowledge to novel problems. 
Fig.~\ref{generalization_error} compares three LLMs on seen vs unseen DSL structures under POI-anonymized evaluation with syntax-level pattern matching. 
% Results reveal critical gaps: 84\% novel DSLs show only 12\% alignment (9\% overall), vs 93\% accuracy on 16\% known patterns. GPT-4o and Qwen3 also demonstrate this limitation, excelling on same concepts but failing on unseen concept compositions. 
Unseen compositions constitute 84\% of cases but achieve only 12\% structure alignment (9\% overall when weighted by frequency), whereas seen patterns (16\% of cases) reach 93\% accuracy. This gap holds across the evaluated LLMs, which perform well on seen patterns but drop sharply on unseen concept compositions, suggesting limited compositional generalization. 

In summary, NeSy methods outperform LLM-only baselines on constraint satisfaction, yet open-world challenges remain. With authentic queries and DSL-based compositional validation, ChinaTravel surfaces these limitations and delineates actionable directions for further research. 

\blue{
\textbf{Path Forward:} 
The substantial performance gap between standard NeSy planning and its oracle-DSL variant indicates that the primary bottleneck lies in constraint grounding, i.e., the ability to faithfully translate open-ended natural language into compositional constraints. General-purpose LLMs that rely solely on in-context prompting still struggle to generalize to unseen constraint syntax (as detailed in Fig.~\ref{challenge_NL2DSL}) and leave considerable room for improvement in open-ended semantic grounding (as detailed in Fig.~\ref{poi_reasoning}). Although ChinaTravel provides (query, DSL) pairs for supervised fine-tuning and a sandbox with verifiable signals for reinforcement learning strategies, achieving robust compositional generalization remains an open challenge, likely requiring sophisticated data sampling and augmentation strategies~\citep{wu2025compact, akyurek2020learning} as well as advanced learning paradigms~\citep{yang2024exploring, liu2020compositional, lake2019compositional}. Since this work primarily serves as a benchmark to identify the gap between current research and real-world scenarios, rather than to deliver a complete solution, we leave the development of such methods to future work.}

\subsection{Ablation Study with Preference} 
% \begin{wrapfigure}[8]{r}{0.5\columnwidth}
%     \vspace{-.5in}
% % \begin{figure}[t]
%    \centering
%     \includegraphics[width=.5\columnwidth]{imgs/preference.pdf}
%     \vspace{-.2in}
%     \caption{Ablation on preference ranking.}
%     \label{fig_preference}
% % \end{figure}
% \end{wrapfigure}

\begin{figure*}[t]
    \centering
    \begin{subfigure}{0.48\textwidth}
       \centering
    \includegraphics[width=\columnwidth]{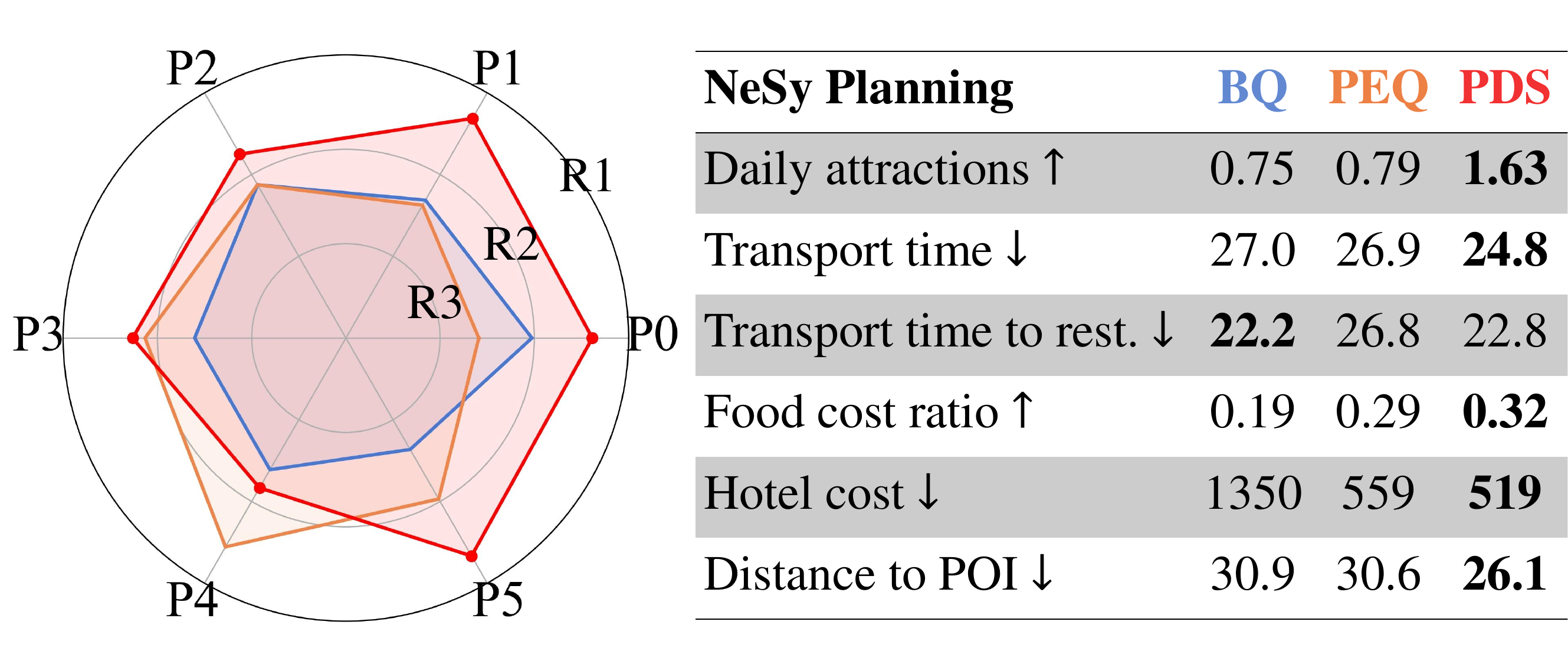}
       \caption{Ablation on preference ranking.}
        \label{fig_preference}
   \end{subfigure}
    \begin{subfigure}{0.48\textwidth}
       \centering
    \includegraphics[width=0.48\columnwidth]{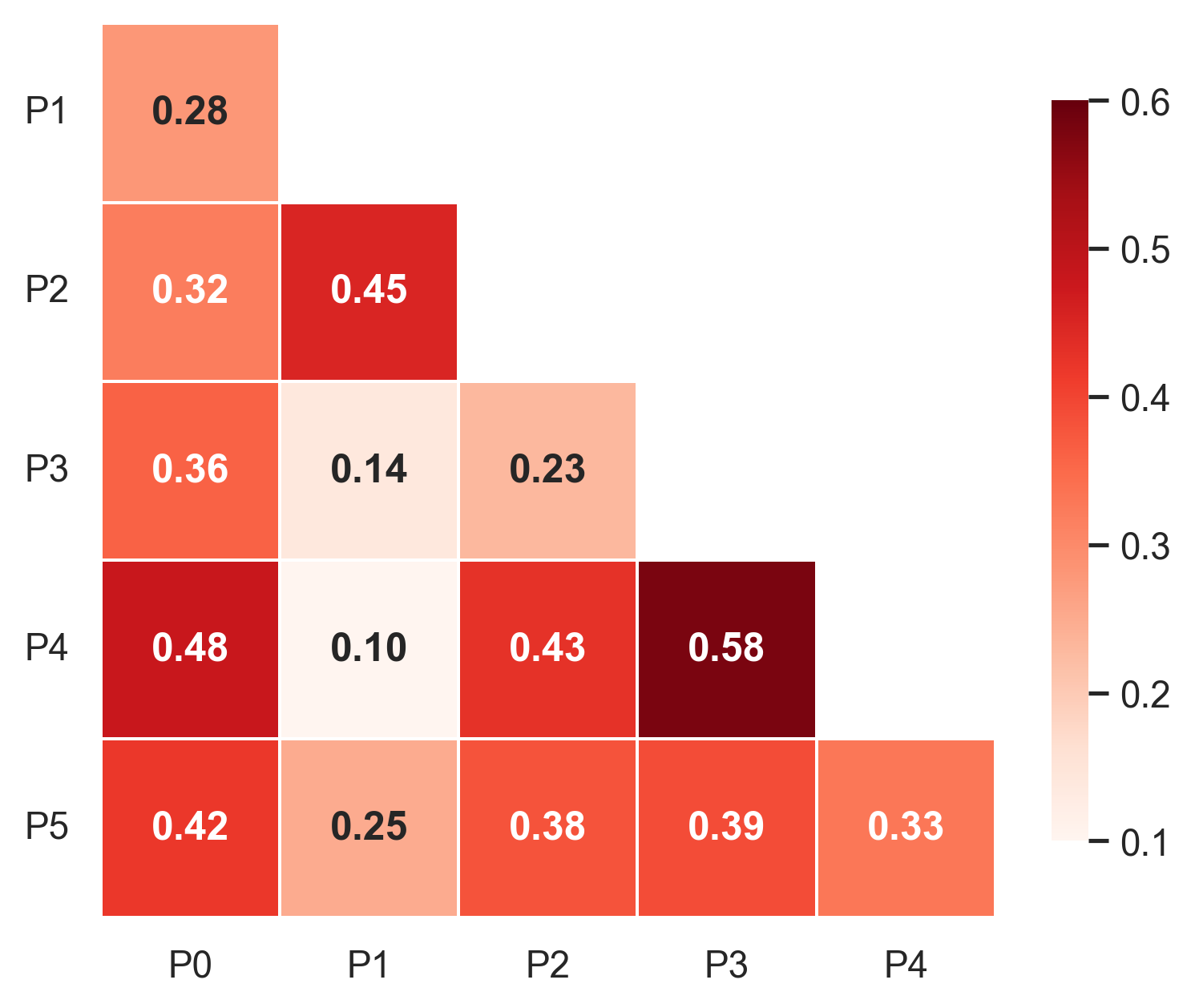}
    \includegraphics[width=0.48\columnwidth]{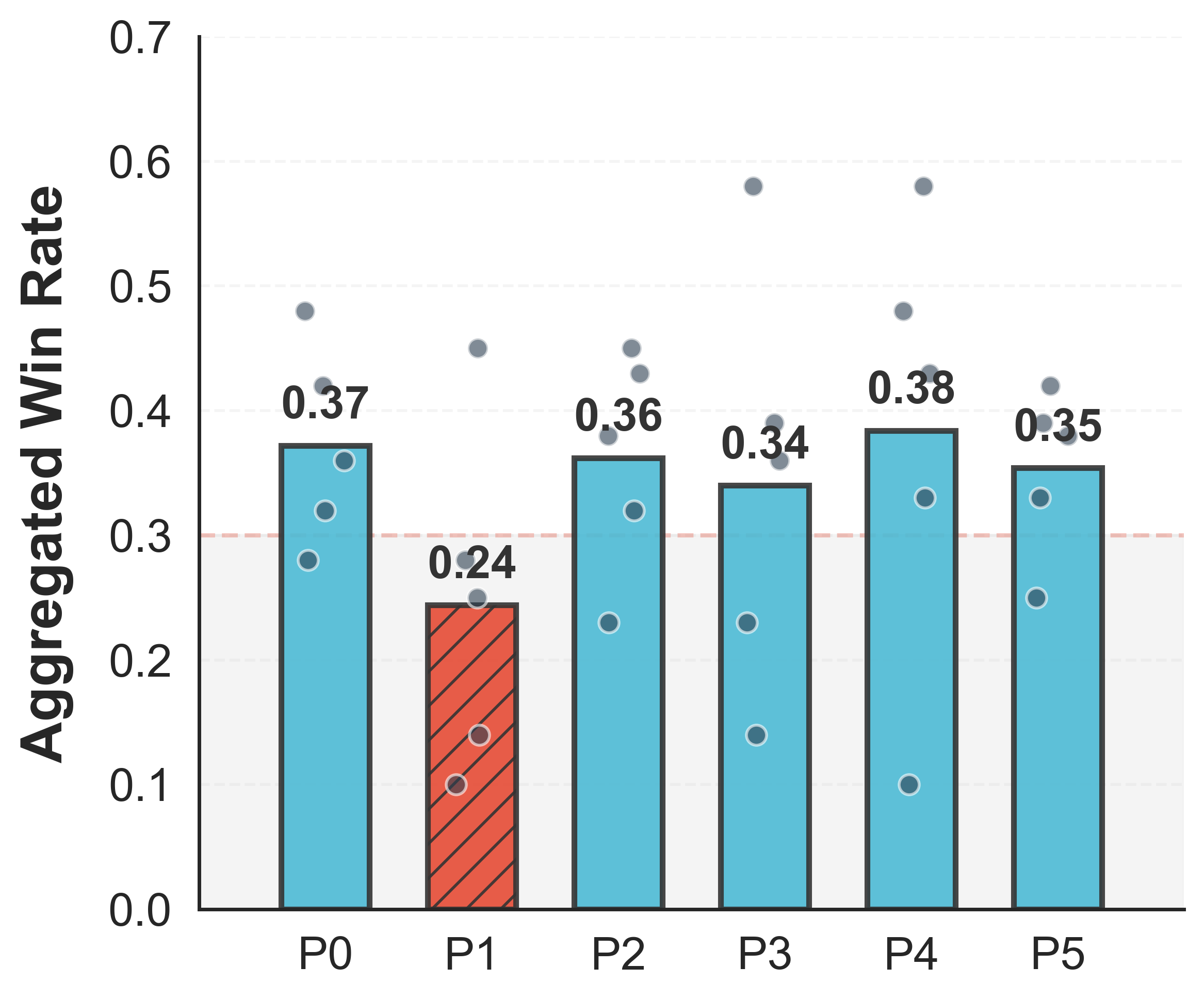}
       \caption{Pareto Win rates (PEQ v.s. BQ).}
       \label{pareto_win_rates}
   \end{subfigure}
    \caption{Empirical Analysis on Planning with Preferences.}
    % \vspace{-.2in}
\end{figure*}

% The comparison of preferences should be conducted under the premise that both environmental and logical constraints are satisfied. 
Preference comparisons are meaningful only when both environmental and logical constraints are satisfied. 
Given the limited FPR achieved by existing methods, 
we perform a separate analysis of preference optimization here. 
Specifically, we sampl 50 queries from the easy subset that NeSy-DeepSeek-Oracle successfully passed as seed samples. 
\blue{Based on these, two experimental settings were designed to explore the roles of LLMs and symbolic search.} 

\textbf{Single-Preference Optimization.} % Based on these, six subsets were created by introducing common preferences identified from user surveys. 
We first evaluated scenarios with a single preference objective using six subsets created from user surveys. 
Three comparative scenarios were designed to explore the roles of LLMs and symbolic search in optimizing preferences during NeSy Planning: 
(1) BQ: Baseline solutions without preference consideration. 
(2) PEQ: LLM-enhanced recommendations with natural language preferences. 
(3) PDS: Hybrid symbolic search optimizing preference objectives under 5-min constraints. 
From the results in Fig.~\ref{fig_preference} (where $\uparrow/\downarrow$ indicate maximization/minimization), 
we cound find that: (1) PEQ outperforms BQ in 5/6 preference scenarios, confirming LLMs' capacity to interpret natural language preferences during POI ranking. (2) PEQ underperforms on P2 (minimizing transport time to restaurants), likely from LLMs' misinterpretation of complex spatiotemporal constraints. 
These results support the DSL's scalability for preference optimization and highlight the need for more efficient algorithms for preference-aware planning. 

\blue{\textbf{Multi-Preference Trade-offs.} Real-world planning may involve balancing multiple, potentially conflicting preferences. To address this, we constructed 15 test subsets by pairing the six preferences (P0-P5) into all possible combinations (e.g., ``maximize daily attractions" + ``minimize hotel cost"). 
Excluding PDS due to the complexity of symbolic weighting, we compare PEQ against BQ using a \textbf{Pairwise Pareto Win-Rate}. PEQ is declared the winner (1.0) if it generates a feasible plan that Pareto-dominates BQ (strictly better on at least one metric without degrading the other) and the loser (0.0) if BQ conversely dominates PEQ. Cases where neither dominates or both fail on constraints are recorded as ties (0.5).} 
\blue{Fig.~\ref{pareto_win_rates} illustrates the win rates and their aggregation across 15 test settings. % PEQ achieves an overall win rate of 0.343 against BQ. 
The results reveal meaningful structures in conflict resolution: PEQ performs well when jointly optimizing synergistic attributes, such as ``maximize food cost ratio" + ``minimize hotel cost" (P3 \& P4, Win Rate 0.58), ``less inner transports time” + ``minimize average transport time to restaurants” (P1 \& P2, Win Rate 0.45). In contrast, PEQ underperforms in some cost-sensitive combinations such as ``minimize inner-city transport time" and ``minimize hotel cost" (P1 \& P4, Win Rate 0.10). These findings underscore the current limitations of LLMs in navigating rigid trade-offs between spatiotemporal efficiency and financial constraints, identifying a critical direction for future research.}

\section{Conclusion}
We present ChinaTravel, an open-ended benchmark for multi-day multi-POI travel planning focused on authentic Chinese needs. 
% We address the limitations of previous benchmarks by incorporating open-ended and diverse human queries, capturing real-world user needs. 
It addresses gaps in prior benchmarks by pairing open-ended human queries with a DSL-based framework for compositional constraint validation, enabling evaluation of feasibility, constraint satisfaction, and preference comparison. 
% Additionally, we propose a scalable evaluation framework based on DSL, enabling comprehensive assessments of feasibility, constraint satisfaction, and preference comparison. 
The empirical analysis reveals the potential of neuro-symbolic methods on constraint adherence. % and preference optimization. 
% We further reveal the open-world challenges: contextual grounding and compositional concept generalization. 
The open-world challenges identified, contextual grounding and compositional concept generalization, suggest actionable directions for future work. 
% These advancements provide a foundation for developing language agents capable of meeting diverse user requirements and delivering reliable travel solutions. 
% These advances provide a foundation for advancing language agents toward meeting open-ended user requirements and more reliable real-world travel planning. 
We hope ChinaTravel will facilitate progress in LLM-powered travel planning by providing a standardized evaluation framework and highlighting key challenges for improvement. 

\section{Acknowledgment}
This research was supported by Leading-edge Technology Program of Jiangsu Science Foundation (BK20232003), the Key Program of Jiangsu Science Foundation (BK20243012, BG2024036), Natural Science Foundation of China (62576162), and the Fundamental Research Funds for the Central Universities (022114380023).

\section{Ethics statement}

We adhere to the \href{https://iclr.cc/public/CodeOfEthics}{ICLR Code of Ethics} and conducted a proactive review of data collection, curation, evaluation, and release. 

\paragraph{Potential Positive impacts.} ChinaTravel is a research benchmark for complex, real-world trip planning, by stressing compositional constraints and verifiable outcomes, it aims to catalyze more reliable, constraint-aware assistants and to facilitate cross-disciplinary research. Its positive societal impacts include:   
(1) Improved Travel Planning Effectiveness: By rigorously testing agents' ability to handle multi-day itineraries and combinatorial constraints, this benchmark encourages the creation of more robust AI assistants, potentially reducing the time and effort users spend on organizing trips. (2) Validation for Real-World Applications: The benchmark establishes a critical foundation for deploying language agents in practical travel planning settings, where multi-objective planning and constraint satisfaction are essential. % (3) Promotion of Open Research: 
The release of this benchmark bridges cutting-edge LLMs with classical neuro-symbolic AI paradigms, fostering cross-disciplinary collaboration between academia and industry. It promotes the reliable, constraint-aware AI systems, while accelerating innovation in both foundational planning capabilities and real-world deployment scenarios. 

\paragraph{Potential negative impacts.} It largely depend on how future systems built upon this benchmark are deployed. For instance: (1) Bias and Fairness: If agents inherit biases from training data or misalign with diverse user preferences, they might disproportionately recommend certain POIs or services. Mitigation requires ongoing fairness audits and inclusive data practices. 
(2) Misuse Risks: Malicious actors could exploit highly capable planning agents to generate deceptive itineraries or manipulate travel services. Such risks underscore the need for ethical guidelines and safeguards in downstream applications. 
ChinaTravel is released for research purposes only. Any real-world deployment should include additional safety engineering, for example, explicitly warning users that agent-generated plans are suggestions, and implementing verification mechanisms (e.g., feasibility and constraint checks) before adoption.

\paragraph{Language and Regional Scope (Bias Considerations).}
Our benchmark focuses on Chinese cities and collects requirements from native Chinese speakers because, for POI-rich, locale-specific travel planning, interacting in the target user's language yields more faithful intent capture and more coherent system behavior. This mirrors common practice in domain-specific systems (e.g., TravelPlanner~\citep{TravelPlanner} uses English for U.S. scenarios). 
While our initial release centers on Chinese due to realistic usage and practical constraints (API costs and compute/latency budgets), the core components are language- and region-agnostic: the tool-grounded sandbox, the DSL-based verification framework, and the identified open-world challenge are independent of any particular language. Future iterations will extend ChinaTravel's language coverage to address global tourism demands. Our goal is not to privilege one language or culture, but to start from a high-fidelity setting where users naturally articulate open-ended, diverse travel requirements, enabling transparent, generalizable evaluation of reliable planning agents that better align with real-world deployment.

\section{Reproducibility Statement} 

An anonymous, downloadable codebase and the dataset splits are provided in the supplementary attachments (with a README that lists dependencies, exact run commands, and config files). The main paper and appendix together specify all components needed for reruns: benchmark design details and data details (App.~\ref{sec:design}), tutorials for the DSL and preferences (App.~\ref{app_concept_func} and~\ref{preference_example}), the search sketch, pseudo-code, and prompts for our NeSy Planning baseline (App.~\ref{app_nesy}), evaluation protocol and metrics (App.~\ref{app:metric}). We also document scientific artifacts (availability \& licensing) in App.~\ref{app:artifacts}. 

\bibliography{iclr2026_conference}

\begin{thebibliography}{41}
\providecommand{\natexlab}[1]{#1}
\providecommand{\url}[1]{\texttt{#1}}
\expandafter\ifx\csname urlstyle\endcsname\relax
  \providecommand{\doi}[1]{doi: #1}\else
  \providecommand{\doi}{doi: \begingroup \urlstyle{rm}\Url}\fi

\bibitem[Adamic \& Huberman(2002)Adamic and Huberman]{adamic2002zipf}
Lada~A Adamic and Bernardo~A Huberman.
\newblock Zipf's law and the internet.
\newblock \emph{Glottometrics}, 3\penalty0 (1):\penalty0 143--150, 2002.

\bibitem[Aky{\"u}rek et~al.(2020)Aky{\"u}rek, Aky{\"u}rek, and Andreas]{akyurek2020learning}
Ekin Aky{\"u}rek, Afra~Feyza Aky{\"u}rek, and Jacob Andreas.
\newblock Learning to recombine and resample data for compositional generalization.
\newblock \emph{arXiv preprint arXiv:2010.03706}, 2020.

\bibitem[Campbell et~al.(2002)Campbell, Hoane~Jr, and Hsu]{campbell2002deep}
Murray Campbell, A~Joseph Hoane~Jr, and Feng-hsiung Hsu.
\newblock Deep blue.
\newblock \emph{Artificial intelligence}, 134\penalty0 (1-2):\penalty0 57--83, 2002.

\bibitem[Chen et~al.(2024)Chen, Ge, Fu, Xiao, and Chen]{chen2024travelagent}
Aili Chen, Xuyang Ge, Ziquan Fu, Yanghua Xiao, and Jiangjie Chen.
\newblock Travel{A}gent: {A}n {AI} assistant for personalized travel planning.
\newblock \emph{arXiv preprint arXiv:2409.08069}, 2024.

\bibitem[Chen et~al.(2025)Chen, Ren, Chen, Yang, Sun, Yoon, and Ar{\i}k]{chen2025sets}
Jiefeng Chen, Jie Ren, Xinyun Chen, Chengrun Yang, Ruoxi Sun, Jinsung Yoon, and Sercan~{\"O} Ar{\i}k.
\newblock Sets: Leveraging self-verification and self-correction for improved test-time scaling.
\newblock \emph{arXiv preprint arXiv:2501.19306}, 2025.

\bibitem[Choi et~al.(2025)Choi, Yoon, Chen, Jha, and Pfister]{choi2025atlas}
Jihye Choi, Jinsung Yoon, Jiefeng Chen, Somesh Jha, and Tomas Pfister.
\newblock Atlas: Constraints-aware multi-agent collaboration for real-world travel planning.
\newblock \emph{arXiv preprint arXiv:2509.25586}, 2025.

\bibitem[Dai et~al.(2019)Dai, Xu, Yu, and Zhou]{DBLP:conf/nips/DaiX0Z19}
Wang{-}Zhou Dai, Qiu{-}Ling Xu, Yang Yu, and Zhi{-}Hua Zhou.
\newblock Bridging machine learning and logical reasoning by abductive learning.
\newblock In \emph{Advances in Neural Information Processing Systems}, pp.\  2811--2822, 2019.

\bibitem[Deng et~al.(2024)Deng, Dong, and Si]{deng2024enhancing}
Shujie Deng, Honghua Dong, and Xujie Si.
\newblock Enhancing and evaluating logical reasoning abilities of large language models.
\newblock In \emph{Proceedings of the ICLR 2024 Workshop on Secure and Trustworthy Large Language Models}, 2024.

\bibitem[Fodor(1975)]{fodor1975language}
Jerry Fodor.
\newblock \emph{The language of thought}.
\newblock Harvard University Press, 1975.

\bibitem[Fodor(2008)]{fodor2008lot}
Jerry~A Fodor.
\newblock \emph{LOT 2: The language of thought revisited}.
\newblock Oup Oxford, 2008.

\bibitem[Gleixner et~al.(2021)Gleixner, Hendel, Gamrath, Achterberg, Bastubbe, Berthold, Christophel, Jarck, Koch, Linderoth, et~al.]{gleixner2021miplib}
Ambros Gleixner, Gregor Hendel, Gerald Gamrath, Tobias Achterberg, Michael Bastubbe, Timo Berthold, Philipp Christophel, Kati Jarck, Thorsten Koch, Jeff Linderoth, et~al.
\newblock Miplib 2017: data-driven compilation of the 6th mixed-integer programming library.
\newblock \emph{Mathematical Programming Computation}, 13\penalty0 (3):\penalty0 443--490, 2021.

\bibitem[Gundawar et~al.(2024)Gundawar, Verma, Guan, Valmeekam, Bhambri, and Kambhampati]{gundawar2024robust}
Atharva Gundawar, Mudit Verma, Lin Guan, Karthik Valmeekam, Siddhant Bhambri, and Subbarao Kambhampati.
\newblock Robust planning with llm-modulo framework: Case study in travel planning.
\newblock \emph{arXiv preprint arXiv:2405.20625}, 2024.

\bibitem[Gupta \& Kembhavi(2023)Gupta and Kembhavi]{DBLP:conf/cvpr/GuptaK23}
Tanmay Gupta and Aniruddha Kembhavi.
\newblock Visual programming: Compositional visual reasoning without training.
\newblock In \emph{Proceedings of the {IEEE/CVF} Conference on Computer Vision and Pattern Recognition}, pp.\  14953--14962, 2023.

\bibitem[Halder et~al.(2024)Halder, Lim, Chan, and Zhang]{DBLP:journals/asc/HalderLCZ24}
Sajal Halder, Kwan~Hui Lim, Jeffrey Chan, and Xiuzhen Zhang.
\newblock A survey on personalized itinerary recommendation: From optimisation to deep learning.
\newblock \emph{Applied Soft Computing}, 152:\penalty0 111200, 2024.

\bibitem[Hao et~al.(2025)Hao, Chen, Zhang, and Fan]{MITSMT}
Yilun Hao, Yongchao Chen, Yang Zhang, and Chuchu Fan.
\newblock Large language models can solve real-world planning rigorously with formal verification tools.
\newblock In \emph{Proceedings of the 2025 Conference of the Nations of the Americas Chapter of the Association for Computational Linguistics}, Albuquerque, New Mexico, 2025.

\bibitem[Jimenez et~al.(2024)Jimenez, Yang, Wettig, Yao, Pei, Press, and Narasimhan]{DBLP:conf/iclr/JimenezYWYPPN24}
Carlos~E. Jimenez, John Yang, Alexander Wettig, Shunyu Yao, Kexin Pei, Ofir Press, and Karthik~R. Narasimhan.
\newblock Swe-bench: Can language models resolve real-world github issues?
\newblock In \emph{Proceedings of the 12th International Conference on Learning Representations}, 2024.

\bibitem[Ju et~al.(2024)Ju, Jiang, Cohen, Foss, Mitts, Zharmagambetov, Amos, Li, Kao, Fazel-Zarandi, et~al.]{ju2024globe}
Da~Ju, Song Jiang, Andrew Cohen, Aaron Foss, Sasha Mitts, Arman Zharmagambetov, Brandon Amos, Xian Li, Justine Kao, Maryam Fazel-Zarandi, et~al.
\newblock To the globe (ttg): Towards language-driven guaranteed travel planning.
\newblock In \emph{Proceedings of the 2024 Conference on Empirical Methods in Natural Language Processing: System Demonstrations}, pp.\  240--249, 2024.

\bibitem[Kambhampati et~al.(2024)Kambhampati, Valmeekam, Guan, Verma, Stechly, Bhambri, Saldyt, and Murthy]{DBLP:conf/icml/KambhampatiVGVS24}
Subbarao Kambhampati, Karthik Valmeekam, Lin Guan, Mudit Verma, Kaya Stechly, Siddhant Bhambri, Lucas Saldyt, and Anil Murthy.
\newblock Position: Llms can't plan, but can help planning in llm-modulo frameworks.
\newblock In \emph{Forty-first International Conference on Machine Learning}, Vienna, Austria, 2024.

\bibitem[Lake(2019)]{lake2019compositional}
Brenden~M Lake.
\newblock Compositional generalization through meta sequence-to-sequence learning.
\newblock \emph{Advances in neural information processing systems}, 32, 2019.

\bibitem[Liu et~al.(2020)Liu, An, Lou, Chen, Lin, Gao, Zhou, Zheng, and Zhang]{liu2020compositional}
Qian Liu, Shengnan An, Jian-Guang Lou, Bei Chen, Zeqi Lin, Yan Gao, Bin Zhou, Nanning Zheng, and Dongmei Zhang.
\newblock Compositional generalization by learning analytical expressions.
\newblock \emph{Advances in Neural Information Processing Systems}, 33:\penalty0 11416--11427, 2020.

\bibitem[Liu et~al.(2024)Liu, Chen, Hsu, Mao, and Wu]{DBLP:conf/iclr/LiuCHM024}
Weiyu Liu, Geng Chen, Joy Hsu, Jiayuan Mao, and Jiajun Wu.
\newblock Learning planning abstractions from language.
\newblock In \emph{Proceedings of the 12th International Conference on Learning Representations}, 2024.

\bibitem[Manhaeve et~al.(2018)Manhaeve, Dumancic, Kimmig, Demeester, and Raedt]{DBLP:conf/nips/ManhaeveDKDR18}
Robin Manhaeve, Sebastijan Dumancic, Angelika Kimmig, Thomas Demeester, and Luc~De Raedt.
\newblock Deepproblog: Neural probabilistic logic programming.
\newblock In \emph{Advances in Neural Information Processing Systems}, pp.\  3753--3763, 2018.

\bibitem[Mnih et~al.(2013)Mnih, Kavukcuoglu, Silver, Graves, Antonoglou, Wierstra, and Riedmiller]{DBLP:journals/corr/MnihKSGAWR13}
Volodymyr Mnih, Koray Kavukcuoglu, David Silver, Alex Graves, Ioannis Antonoglou, Daan Wierstra, and Martin~A. Riedmiller.
\newblock Playing {A}tari with deep reinforcement learning.
\newblock \emph{CoRR}, abs/1312.5602, 2013.

\bibitem[Pan et~al.(2023)Pan, Albalak, Wang, and Wang]{DBLP:conf/emnlp/PanAWW23}
Liangming Pan, Alon Albalak, Xinyi Wang, and William~Yang Wang.
\newblock Logic-{LM}: Empowering large language models with symbolic solvers for faithful logical reasoning.
\newblock In \emph{Findings of the Association for Computational Linguistics: {EMNLP}}, pp.\  3806--3824, 2023.

\bibitem[Piantadosi et~al.(2016)Piantadosi, Tenenbaum, and Goodman]{piantadosi2016logical}
Steven~T Piantadosi, Joshua~B Tenenbaum, and Noah~D Goodman.
\newblock The logical primitives of thought: Empirical foundations for compositional cognitive models.
\newblock \emph{Psychological review}, 123\penalty0 (4):\penalty0 392, 2016.

\bibitem[Sharma et~al.(2017)Sharma, Goyal, and Malik]{sharma2017intelligent}
Vibhor Sharma, Monika Goyal, and Drishti Malik.
\newblock An intelligent behaviour shown by chatbot system.
\newblock \emph{International Journal of New Technology and Research}, 3\penalty0 (4):\penalty0 263312, 2017.

\bibitem[Shinn et~al.(2024)Shinn, Cassano, Gopinath, Narasimhan, and Yao]{DBLP:conf/nips/ShinnCGNY23}
Noah Shinn, Federico Cassano, Ashwin Gopinath, Karthik Narasimhan, and Shunyu Yao.
\newblock Reflexion: language agents with verbal reinforcement learning.
\newblock In \emph{Advances in Neural Information Processing Systems}, 2024.

\bibitem[Silver et~al.(2017)Silver, Schrittwieser, Simonyan, Antonoglou, Huang, Guez, Hubert, Baker, Lai, Bolton, Chen, Lillicrap, Hui, Sifre, van~den Driessche, Graepel, and Hassabis]{DBLP:journals/nature/SilverSSAHGHBLB17}
David Silver, Julian Schrittwieser, Karen Simonyan, Ioannis Antonoglou, Aja Huang, Arthur Guez, Thomas Hubert, Lucas Baker, Matthew Lai, Adrian Bolton, Yutian Chen, Timothy~P. Lillicrap, Fan Hui, Laurent Sifre, George van~den Driessche, Thore Graepel, and Demis Hassabis.
\newblock Mastering the game of {G}o without human knowledge.
\newblock \emph{Nature}, 550\penalty0 (7676):\penalty0 354--359, 2017.

\bibitem[Tang et~al.(2024)Tang, Wang, Qu, Yan, Hou, Zhuang, Guo, Zhao, Zhao, and Ma]{DBLP:journals/corr/abs-2402-07204}
Yihong Tang, Zhaokai Wang, Ao~Qu, Yihao Yan, Kebing Hou, Dingyi Zhuang, Xiaotong Guo, Jinhua Zhao, Zhan Zhao, and Wei Ma.
\newblock Synergizing spatial optimization with large language models for open-domain urban itinerary planning.
\newblock \emph{CoRR}, abs/2402.07204, 2024.

\bibitem[Wang et~al.(2019)Wang, Donti, Wilder, and Kolter]{DBLP:conf/icml/WangDWK19}
Po{-}Wei Wang, Priya~L. Donti, Bryan Wilder, and J.~Zico Kolter.
\newblock {SATN}et: Bridging deep learning and logical reasoning using a differentiable satisfiability solver.
\newblock In \emph{Proceedings of the 36th International Conference on Machine Learning}, pp.\  6545--6554, 2019.

\bibitem[Wu et~al.(2025)Wu, Hwang, Kirichenko, and Russakovsky]{wu2025compact}
Xindi Wu, Hee~Seung Hwang, Polina Kirichenko, and Olga Russakovsky.
\newblock Compact: Compositional atomic-to-complex visual capability tuning.
\newblock \emph{arXiv preprint arXiv:2504.21850}, 2025.

\bibitem[Xi et~al.(2023)Xi, Chen, Guo, He, Ding, Hong, Zhang, Wang, Jin, Zhou, Zheng, Fan, Wang, Xiong, Zhou, Wang, Jiang, Zou, Liu, Yin, Dou, Weng, Cheng, Zhang, Qin, Zheng, Qiu, Huang, and Gui]{DBLP:journals/corr/abs-2309-07864}
Zhiheng Xi, Wenxiang Chen, Xin Guo, Wei He, Yiwen Ding, Boyang Hong, Ming Zhang, Junzhe Wang, Senjie Jin, Enyu Zhou, Rui Zheng, Xiaoran Fan, Xiao Wang, Limao Xiong, Yuhao Zhou, Weiran Wang, Changhao Jiang, Yicheng Zou, Xiangyang Liu, Zhangyue Yin, Shihan Dou, Rongxiang Weng, Wensen Cheng, Qi~Zhang, Wenjuan Qin, Yongyan Zheng, Xipeng Qiu, Xuanjing Huang, and Tao Gui.
\newblock The rise and potential of large language model based agents: {A} survey.
\newblock \emph{CoRR}, abs/2309.07864, 2023.

\bibitem[Xie et~al.(2024)Xie, Zhang, Chen, Zhu, Lou, Tian, Xiao, and Su]{TravelPlanner}
Jian Xie, Kai Zhang, Jiangjie Chen, Tinghui Zhu, Renze Lou, Yuandong Tian, Yanghua Xiao, and Yu~Su.
\newblock Travelplanner: {A} benchmark for real-world planning with language agents.
\newblock In \emph{Proceedings of the 41st International Conference on Machine Learning}, 2024.

\bibitem[Xiong et~al.(2025)Xiong, Payani, Yang, and Fekri]{DBLP:journals/corr/abs-2410-03136}
Siheng Xiong, Ali Payani, Yuan Yang, and Faramarz Fekri.
\newblock Deliberate reasoning in language models as structure-aware planning with an accurate world model.
\newblock In \emph{Proceedings of the 63rd Annual Meeting of the Association for Computational Linguistics (Volume 1: Long Papers)}, pp.\  31900--31931, Vienna, Austria, 2025.

\bibitem[Xiong et~al.(2026)Xiong, Payani, and Fekri]{xiong2026enhancing}
Siheng Xiong, Ali Payani, and Faramarz Fekri.
\newblock Enhancing language model reasoning with structured multi-level modeling.
\newblock In \emph{The 14th International Conference on Learning Representations}, 2026.

\bibitem[Yang et~al.(2024)Yang, Lu, Lam, and Cai]{yang2024exploring}
Haoran Yang, Hongyuan Lu, Wai Lam, and Deng Cai.
\newblock Exploring compositional generalization of large language models.
\newblock In \emph{Proceedings of the 2024 Conference of the North American Chapter of the Association for Computational Linguistics: Human Language Technologies (Volume 4: Student Research Workshop)}, pp.\  16--24, 2024.

\bibitem[Yao et~al.(2023{\natexlab{a}})Yao, Yu, Zhao, Shafran, Griffiths, Cao, and Narasimhan]{DBLP:conf/nips/YaoYZS00N23}
Shunyu Yao, Dian Yu, Jeffrey Zhao, Izhak Shafran, Tom Griffiths, Yuan Cao, and Karthik Narasimhan.
\newblock Tree of thoughts: Deliberate problem solving with large language models.
\newblock In Alice Oh, Tristan Naumann, Amir Globerson, Kate Saenko, Moritz Hardt, and Sergey Levine (eds.), \emph{Advances in Neural Information Processing Systems 36: Annual Conference on Neural Information Processing Systems 2023, NeurIPS 2023, New Orleans, LA, USA, December 10 - 16, 2023}, 2023{\natexlab{a}}.

\bibitem[Yao et~al.(2023{\natexlab{b}})Yao, Zhao, Yu, Du, Shafran, Narasimhan, and Cao]{DBLP:conf/iclr/YaoZYDSN023}
Shunyu Yao, Jeffrey Zhao, Dian Yu, Nan Du, Izhak Shafran, Karthik~R. Narasimhan, and Yuan Cao.
\newblock React: Synergizing reasoning and acting in language models.
\newblock In \emph{Proceedings of the 11th International Conference on Learning Representations}, 2023{\natexlab{b}}.

\bibitem[Ye et~al.(2025)Ye, Yin, He, Zhang, Yen, Gao, Durrett, and Chen]{ye2025longproc}
Xi~Ye, Fangcong Yin, Yinghui He, Joie Zhang, Howard Yen, Tianyu Gao, Greg Durrett, and Danqi Chen.
\newblock Longproc: Benchmarking long-context language models on long procedural generation.
\newblock \emph{arXiv preprint arXiv:2501.05414}, 2025.

\bibitem[Zhang et~al.(2023)Zhang, Chen, Jiang, Yu, Chen, Chen, Li, Wu, Zhang, Xiao, Wan, Wang, and Li]{DBLP:conf/emnlp/ZhangCJYCCLWZXW23}
Hongbo Zhang, Junying Chen, Feng Jiang, Fei Yu, Zhihong Chen, Guiming Chen, Jianquan Li, Xiangbo Wu, Zhiyi Zhang, Qingying Xiao, Xiang Wan, Benyou Wang, and Haizhou Li.
\newblock Huatuogpt, towards taming language model to be a doctor.
\newblock In \emph{Findings of the Association for Computational Linguistics: {EMNLP}}, pp.\  10859--10885, 2023.

\bibitem[Zheng et~al.(2024)Zheng, Mishra, Zhang, Chen, Chen, Nova, Hou, Cheng, Le, Chi, et~al.]{zheng2024natural}
Huaixiu~Steven Zheng, Swaroop Mishra, Hugh Zhang, Xinyun Chen, Minmin Chen, Azade Nova, Le~Hou, Heng-Tze Cheng, Quoc~V Le, Ed~H Chi, et~al.
\newblock Natural plan: Benchmarking llms on natural language planning.
\newblock \emph{arXiv preprint arXiv:2406.04520}, 2024.

\end{thebibliography}
\bibliographystyle{iclr2026_conference}

\appendix
\newpage
\renewcommand{\contentsname}{Table of Contents}
\tableofcontents

\section{The Use of Large Language Models}

\paragraph{Polish Writing:} We used off-the-shelf LLMs as general-purpose assist tools for sentence polishing during manuscript revision. All LLM-assisted edits were reviewed and revised by the authors. LLMs are not eligible for authorship; the authors take full responsibility for all content. %, including any text influenced by LLMs. 

\paragraph{Data synthesis:} We use LLM to generate two datasets \emph{easy} and \emph{medium} as the components of the ChinaTravel becnhamrk. The complete procedure for synthetic data generation, including prompts, sampling settings, filtering, and human verification criteria, is provided in the App.~\ref{app_querygen}. 

LLMs did not contribute to research ideation, experimental design, or core method development.

\section{Discussion of Limitations} 
Our research represents a significant step forward in evaluating the travel planning capabilities of language agents, but it is not without challenges. One limitation lies in its focus on Chinese travel planning. Due to the inherent differences in natural language, the translated versions of queries may fail to fully capture the challenges of understanding requirements in Chinese queries, potentially limiting its applicability in a global context. However, given the substantial demand within China's travel market, we believe a benchmark tailored to Chinese travel planning is both necessary and socially valuable. 
Although our benchmark is comprehensive, it may not encompass the full range of requirements encountered in real-world scenarios. The high cost of collecting authentic data has limited the number of human queries in our study. 
% Despite this, our work introduces, for the first time, a benchmark incorporating genuine natural language inquiries from human users in travel planning. This has posed new challenges to existing methods, including open language reasoning and the composition of previously unseen concepts, marking a significant step for the development of language agents. 
To address this, future work will focus on combining LLMs with real user queries to automate the generation of a wider variety of human-like queries. Continuous refinement and expansion of our benchmark are crucial for more accurately reflecting the realistic travel planning needs. 

ChinaTravel provides a verifiable, tool-equipped sandbox, but we currently focus on evaluation of prompt-based methods and do not train tool-using agents with RL post-training. We defer these due to resource constraints (compute for large-scale interaction), and open challenges in trajectory synthesis (coverage and off-policy bias). Thus, we plan to explore tool-use trajectory synthesis and corresponding agent training in future work.

\section{Discussion with Related Work}

\textbf{LLM-based Agents} % Large Language Model based Agents 
have demonstrated significant capability in understanding complex instructions and employing domain-specific tools to complete tasks, showcasing
their potential in fields such as visual reasoning~\citep{DBLP:conf/cvpr/GuptaK23}, healthcare~\citep{DBLP:conf/emnlp/ZhangCJYCCLWZXW23} and robotics~\citep{DBLP:conf/iclr/LiuCHM024}. 
This reduces the reliance of previous agents on domain-specific efforts, that is, either mainly following domain-specific rules to plan (rule-based agents, such as DeepBlue~\citep{campbell2002deep} and Eliza~\citep{sharma2017intelligent}) or mainly learning from domain-specific data to plan (reinforcement-learning-based agents, such as AlphaGo~\citep{DBLP:journals/nature/SilverSSAHGHBLB17} and Atari DQN~\citep{DBLP:journals/corr/MnihKSGAWR13}). 
While the language agents have shown promising results in some domains, most of their planning scenarios are limited to simple tasks with single objective function and fail in the travel planning benchmark with complex logical constraints. % on the % planning 
% results. 

% \textbf{OS Agents} 

\textbf{Neuro-Symbolic Learning} explores to combine traditional
symbolic reasoning with learning to enhance the reliability~\citep{DBLP:conf/nips/ManhaeveDKDR18, DBLP:conf/icml/WangDWK19, DBLP:conf/nips/DaiX0Z19}. 
In the era of large language models,~\citet{DBLP:conf/emnlp/PanAWW23} presents the LogicLM integrates LLMs with separate symbolic solvers for various logical reasoning tasks. They first utilize LLMs to translate a natural language problem into a symbolic formulation. Afterward, a deterministic symbolic solver performs inference on the formulated problem to ensure the correctness of the results. \citet{deng2024enhancing} supplement LogicLM with a Self-Refinement Module to enhance the reliability of LLM translation. 
In the travel planning domain,~\citet{MITSMT} presents a framework with a similar pipeline. It first extracts the logical constraints from natural language queries and then formalizes them into SMT code. Thanks to SMT solvers being sound and complete, this neuro-symbolic solution guarantees the generated plans are correct and has basically solved the TravelPlanner benchmark with a 97\% pass rate. 

\textbf{Travel Planning} is a time-consuming task even for humans, encompassing travel-related information gathering, POI selection, route mapping, and customization to meet diverse user needs~\citep{DBLP:journals/asc/HalderLCZ24}. Natural languages are one of the most common ways for users to express their travel requirements. However, the ambiguity and complexity of travel requirements make it still challenging for LLMs to generate accurate and reliable travel plans. 
\citet{TravelPlanner} presents the TravelPlanner benchmark for cross-city travel planning and reveals the inadequacies of pure-LLM-driven agents. 
TravelPlanner generates user queries through LLMs and provides a rigorous evaluation mechanism to verify whether the provided plans can meet the logical constraints in the queries. It has become a pivotal benchmark for language agents in real-world travel planning. 
\citet{DBLP:journals/corr/abs-2402-07204} study the open-domain urban itinerary planning where a single-day multi-POI plan is required. They integrates spatial optimization with large language models and present a system \textsc{IttNera}, to provide customized urban itineraries based on user needs. 
A concurrent work, TravelAgent~\citep{chen2024travelagent}, also considers a multi-day multi-POI travel planning problem for the specified city. It constructs an LLM-powered system to provide personalized plans. However, due to the high cost of collecting and annotating real travel needs, they evaluate the proposed TravelAgent in only 20 queries. This also demonstrates the necessity of introducing a new benchmark for travel planning. 

\section{Detailed Design of ChinaTravel}
\label{sec:design}
\subsection{Sandbox Information}
\label{app_sandbox}
We started collecting travel information with the motivation of planning a multi-day, multi-POI itinerary in four aspects: attractions, accommodation, activities, and transportation. 
Developers first determine the POI description information that needs to be obtained from the user's perspective, such as cuisine and hotel features. Based on this feature set, we collect public information to construct the database. For the design of APIs, we directly support queries based on the regular expressions from LLM agents. %, which we hope will promote the use of advanced tools during planning. 
At the same time, we expect the design of APIs to have similar features and characteristics to existing commercial APIs, enabling our dataset to be applicable to more realistic scenarios. The information our database contains is shown in Table \ref{sandbox_info} and the APIs we offer is in Table \ref{tab_apis}. In Table~\ref{tab:environment_constraints}, we provide the detailed information of environment constraints in ChinaTravel.
\begin{table*}[th]
    \centering
    \begin{tabular}{cl}
        \toprule
         Tool &   \multicolumn{1}{c}{Information}  \\
         \midrule
         \midrule
         Attractions & Name, Type, Latitude, Longitude, Opentime, Endtime, Price, \\ &Recommendmintime, Recommendmaxtime \\
         \midrule
         Accommodations & Name, Name\_en, Featurehoteltype, Latitude, Longitude, Price, Numbed \\
         \midrule
         Restaurants & Name, Latitude, Longitude, Price, Cuisinetype, Opentime, Endtime, \\ & Recommendedfood \\
         \midrule
         Transportation & Transportation in specific city including walk, metro and taxi \\
         \midrule
         IntercityTransport & Flight: FlightID, From, To, BeginTime, EndTime, Duration, Cost \\ & Train: TrainID, TrainType, From, To, BeginTime, EndTime, Duration, Cost \\
         \midrule
         Poi & Names of POIs(including intercity transportation hub) and their coordinates \\
         \bottomrule
         
    \end{tabular}
    \caption{Sandbox Information}
    \label{sandbox_info}
\end{table*}

\subsection{City-wise Distribution Statistics}
Our POI collection was conducted on a per-city basis, ensuring comparable distribution scales across urban datasets. Human queries exhibit a long-tailed distribution across cities, reflecting real-world travel patterns and highlighting practical deployment challenges for travel planning system. The detailed sandbox and dataset statistics are provided in Table~\ref{tab:city_stats} 

\begin{table}[thb]
\centering
\small
\caption{City-wise Statistics of Sandbox and Dataset}
\setlength{\tabcolsep}{3pt}
\label{tab:city_stats}
\begin{tabular}{lccccccc}
\toprule
City & Attractions & Hotels & Restaurants & Queries & Queries  & Queries & Queries \\
&  &  &  & (Total)& (Easy)&(Human-Val) & (Human-Test)\\
\midrule
Beijing   & 334 & 400 & 469 & 210 & 30 & 28 & 152 \\
Chengdu   & 332 & 378 & 466 & 229 & 36 & 15 & 178 \\
Chongqing & 346 & 372 & 436 & 191 & 36 & 16 & 139 \\
Guangzhou & 338 & 399 & 466 &  90 & 24 & 14 &  52 \\
Hangzhou  & 376 & 377 & 457 & 195 & 33 & 10 & 152 \\
Nanjing   & 322 & 372 & 467 & 123 & 30 & 18 &  75 \\
Shanghai  & 359 & 402 & 483 & 180 & 37 & 25 & 118 \\
Shenzhen  & 305 & 497 & 477 &  81 & 35 &  7 &  39 \\
Suzhou    & 358 & 292 & 468 &  69 &  9 & 12 &  48 \\
Wuhan     & 333 & 367 & 456 &  86 & 30 &  9 &  47 \\
\bottomrule
\end{tabular}
\end{table}

\begin{table*}[p]
    \centering
    \hspace*{-1cm}
    \begin{tabular}{p{2.5cm} p{4.8cm} p{5.5cm}}
        \toprule
        Tool & API & Docs \\
        \midrule
        \midrule
        Attractions & attractions\_keys(city) & Return a list of (key, type) pairs of the attractions data. \\
        & attractions\_select(city, key, func) & Return a DataFrame with data filtered by the specified key with the specified function. \\
        & attractions\_id\_is\_open(city, id, time) & Return whether the attraction with the specified ID is open at the specified time. \\
        & attractions\_nearby(city, point, topk, dist) & Return the top K attractions within the specified distance of the location. \\
        & attractions\_types & Return a list of unique attraction types.\\
        \midrule
        
        Accommodations & accommodations\_keys(city) & Return a list of (key, type) pairs of the  accommodations data. \\
        &  accommodations\_select(city, key, func) & Return a DataFrame with data filtered by the specified key with the specified function. \\
        & accommodations\_nearby(city, point, topk, dist) & Return the top K accommodations within the specified distance of the location. \\
        \midrule
        
        Restaurants  & restaurants\_keys(city) & Return a list of (key, type) pairs of the restaurants data. \\
        & restaurants\_select(city, key, func) & Return a DataFrame with data filtered by the specified key with the specified function. \\
        & restaurants\_id\_is\_open(city, id, time) & Return whether the restaurant with the specified ID is open at the specified time. \\
        & restaurants\_nearby(city, point, topk, dist) & Return the top K restaurants within the specified distance of the location. \\
        & restaurants\_with\_recommended\_food (city, food) & Return all restaurants with the specified food in their recommended dishes. \\
        & restaurants\_cuisine(city) & Return a list of unique restaurant cuisines. \\
        \midrule 
        
        Transportation & goto(city, start, end, start\_time, transport\_type) & Return a list of transportation options between two locations with the specified departure time and transportation mode. \\
        \midrule 
        IntercityTransport & intercity\_transport\_select(start\_city, end\_city, intercity\_type, earliest\_leave\_time) & Return the intercity transportation information between two cities. \\
        \midrule
        Others & notedown(description, content) & Write the specified content to the notebook \\
        & plan(query) & Generates a plan based on the notebook content and query and report the plan is done. \\
        & next\_page() & Get the next page of the latest Result history if it exists. Because of the length limited, all returned DataFrame information is split into 10 rows per page. \\
        \bottomrule
        
        \bottomrule
    \end{tabular}
    \caption{APIs}
    \label{tab_apis}
\end{table*}

\begin{table}[p]
\centering
% \hspace*{-1.5cm}
% \scriptsize
\begin{tabular}{@{}p{2.cm} p{5.3cm} p{5.5cm}@{}}
\toprule
\textbf{Category} & \textbf{Environment Constraints} & \textbf{Semantics} \\
\midrule

\multirow{4}{*}{\makecell{Cross-city\\Transportation}}
& Intercity transportation events must occur. & The first event and last event must be cross-city transports. \\
& Available Trains or Airplanes across cities. & The provided TrainID/FlightID, origin and destination should be valid in the travel sandbox. \\
& Correct information of price, duration. & The price and duration information should match the travel sandbox. \\
& Detailed cost on inter-city transportation & Provide number of tickets and cost of each inter-city activity. $cost = price \times tickets$ \\
\midrule

\multirow{3}{*}{\makecell{Inner-city\\ Transportation}}
& Available Metro, Taxi or Walking between different positions. & The provided routes should be valid in the travel sandbox. \\
& Correct information of price, distance, and duration. & These details should match the travel sandbox. \\
& Detailed cost on inner-city transportation & Provide number of tickets/cars and cost. Taxi: 4 people per car. $cost = price \times tickets$, $cost = price \times cars$ \\
\midrule

\multirow{5}{*}{Attractions}
& Available attractions in the target city & The provided attractions should be valid in the travel sandbox. \\
& Visiting during opening hours. & Activities must respect the attraction's opening time. \\
& Correct price information. & Must match the sandbox. \\
& Detailed cost of attraction activity. & Provide ticket number and total cost. $cost = price \times tickets$ \\
& No repeated attractions. & Attractions should not repeat across the trip. \\
\midrule

\multirow{6}{*}{Restaurants}
& Available restaurants in the target city & Must be valid in the travel sandbox. \\
& Visiting during opening hours. & Same as above. \\
& Correct price information. & Must match the sandbox. \\
& Detailed cost of restaurant activity. & $cost = price \times tickets$ \\
& No repeated restaurants. & Same restaurant should not be visited more than once. \\
& Meals served in proper time slots. & Breakfast: 06:00–09:00; Lunch: 11:00–14:00; Dinner: 17:00–20:00. \\
\midrule

\multirow{4}{*}{Accommodation}
& Available accommodations in target city. & Must be valid in the travel sandbox. \\
& Correct price and room type. & Must match the sandbox. \\
& Detailed accommodation cost. & $cost = price \times rooms$ \\
& Required for trips over one day. & A hotel is necessary for multi-day trips. \\
\midrule

\multirow{2}{*}{Time}
& Activity duration details. & Must include start and end time; end time must be after start. \\
& Activities in chronological order. & Events listed in order, respecting preceding transport arrivals. \\
\midrule

Space
& Transport info for changing positions. & If positions differ, the transport route must be included. \\
\bottomrule
\end{tabular}
\vspace{0.5em}
\caption{Environment Constraints and Semantics in ChinaTravel Environment}
\label{tab:environment_constraints}
\end{table}

\begin{table*}[th]
   \centering
   \begin{tabular}{ll}
      \toprule
      \rowcolor[HTML]{EFEFEF} 
      \multicolumn{2}{c}{\textbf{\textit{Logical Constraint}}} \\
      % \hline
      \midrule
      Transportation & The required type of transportation. \\
      Attraction & \textcolor{brown}{The required type or specified attractions.} \\
      Restaurant & The required type or \textcolor{brown}{specified restruants}. \\
      Accommodation & The number of rooms and the room type must meet the requirements.\\
      & \textcolor{brown}{The required features or specified hotels.} \\
      Budget & The total cost is within required budget. \\
      % \hline
      \midrule
      % \rowcolor{brown!40} 
      \rowcolor[HTML]{EFEFEF} 
      \multicolumn{2}{c}{ \textcolor{brown}{\textbf{\textit{Unseen Logical Constraints in Human data}}} } \\
      \midrule
      POIs & \textcolor{brown}{The negation}/conjunction/\textcolor{brown}{disjunction of given POIs} \\
      Time & \textcolor{brown}{The duration of specific activities is within the limitation} \\
      Budget & \textcolor{brown}{The cost of specific activities is within the required budget} \\
      \bottomrule
   \end{tabular}
   \caption{Descriptions of \textbf{Logical Constraints} for two benchmarks. Constraints in black are common in both TravelPlanner and ChinaTravel. Metrics in \textcolor{brown}{brown} are the metrics only in our benchmark. 
   }
   \label{logical_constraints}
\end{table*}

\begin{table*}[th]
   \centering
   \begin{tabular}{ll}
      \toprule
      \rowcolor[HTML]{EFEFEF} 
      \multicolumn{2}{c}{ \textcolor{brown}{\textbf{\textit{Preference Requirements}}} } \\
      \midrule
      Daily attractions $\uparrow$ & \textcolor{brown}{Visit as many attractions as possible}  \\
      Transport time $\downarrow$ & \textcolor{brown}{Minimize the travel time between POIs} \\
      Transport time to the restaurants $\downarrow$ & \textcolor{brown}{Minimize the travel time to restaurants} \\
      Food cost ratio $\uparrow$ & \textcolor{brown}{Maximize the proportion of dining expenses} \\
      Hotel cost $\downarrow$ & \textcolor{brown}{Minimize accommodation costs}\\
      Distance to POI $\downarrow$ & \textcolor{brown}{Visit places as close to \{POI\} as possible} \\
      \bottomrule
   \end{tabular}
   \caption{Descriptions of \textbf{Preference Requirements} in ChinaTravel benchmark. 
   }
   \label{preference_req_des}
\end{table*}

\subsection{Tutorial of DSL Expression}

\label{app_concept_func} 

Here is a tutorial, that provides a step-by-step guide to utilizing ChinaTravel’s Domain-Specific Language (DSL) with predefined concept functions for expressing logical constraints and preferences. 

\paragraph{DSL Overview} In the main body of this paper, we have detailed the basics of our DSL in the Table~\ref{tab_dsl}. The DSL is a Python-like language designed to formalize travel planning requirements into executable code. It enables automated validation of itineraries against user constraints and preferences. Key components include: 1) \textit{Concept Functions}: Predefined functions (e.g., activity\_cost, poi\_distance) that extract attributes from travel plans. 
2) \textit{Operators}: Logical (and, or, not), arithmetic (+, -, *, /), and comparison operators ($<$, $>$, ==).
3) \textit{Control Structures}: Loops (for), conditionals (if), and set operations (union, intersection). More examples are provided in Figure~\ref{dsl_example}.

\begin{figure*}[htbp]
    \centering
    \vspace{-.1in}
    \includegraphics[width=\linewidth]{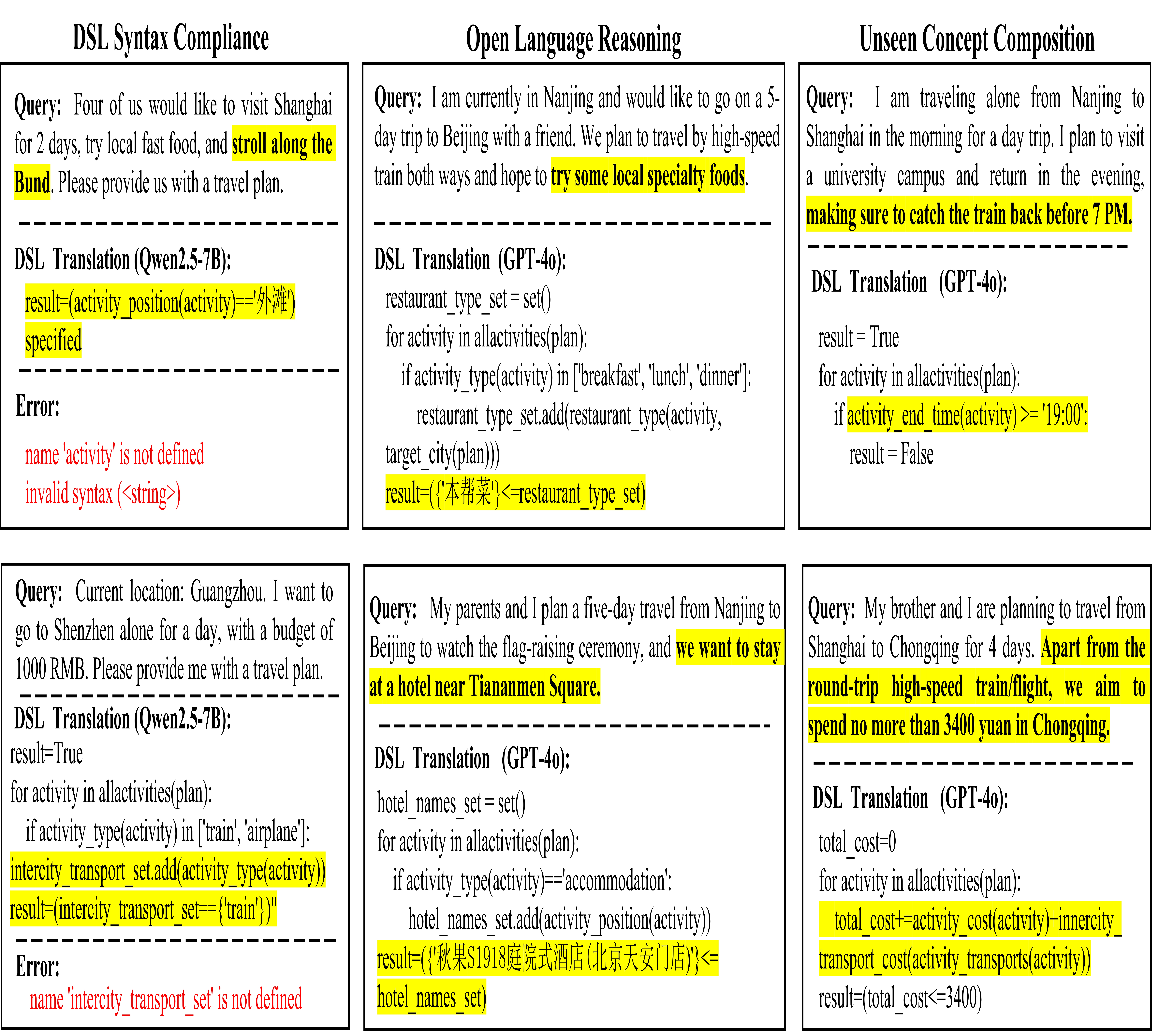}
    \caption{Challenges in the Neuro-Symbolic Planning.}
    \label{challenge_NL2DSL_2}
    \vspace{-.2in}
\end{figure*}

\paragraph{Core Concept Functions} 
We have defined 35 concept functions. Their definition and implementation is in Table \ref{tab_concept_func_0}, \ref{tab_concept_func_1}, \ref{tab_concept_func_2} and \ref{tab_concept_func_3}. 
Below are common use cases:
\newpage
Example: Budget Constraint
User Query: "Total expenses must not exceed 5000 CNY."
\begin{lstlisting}[style=pystyle]
total_cost = 0  
for act in all_activities(plan):  
    total_cost += activity_cost(act)  
    total_cost += innercity_transport_cost(activity_transports(act))
return total_cost <= 5000  
\end{lstlisting}
The function all\_activities(plan) iterates through all activities in the itinerary.
The function activity\_cost retrieves the cost of each activity. 
The function innercity\_transport\_cost sums transportation expenses. 
Based on Python syntax, combining these concept functions can calculate the cost of the entire plan, thereby determining whether the budget constraints are met. 

\paragraph{Debugging Tips}
(1) Syntax Validation: Use the python compiler to check for syntax errors (e.g., missing colons, undefined variables).
(2) Unit Testing: Test individual concept functions (e.g., poi\_distance) with mock itineraries.
(3) Iterative Refinement: For ambiguous requirements (e.g., "local cuisine"), map natural language to precise DSL concepts from sandbox information (e.g., restaurant\_type(act, city) == "Beijing Cuisine"). 

\paragraph{Integration with Neuro-Symbolic Agents.} 
(1) NL2DSL Translation: Convert user queries into DSL using LLMs (e.g., "Try local food" $\rightarrow$ restaurant\_type(POI, city) == "Beijing Cuisine" when the destination city is Beijing).
(2) Symbolic Validation: Execute DSL code to verify plans against logical constraints. 
(3) Search Optimization: Use DSL-defined preferences (e.g., minimize(transport\_time)) to rank candidate itineraries. 

\captionsetup[lstlisting]{labelformat=empty}
\begin{table*}[p]
    \centering
    \vspace*{-1cm}
    \small
    % \hspace*{-1.5cm}
    \begin{tabular}{p{2cm}p{2cm}l}
        \toprule
        Function Name & Meaning & Implementation \\
        \midrule
        \midrule
        day\_count & total days in the plan & \begin{lstlisting}[style=pystyle, frame=none]
def day_count(plan):
    return len(plan["itinerary"])
        \end{lstlisting}  \\
        \midrule
        people\_count & number of people in the trip & \begin{lstlisting}[style=pystyle, frame=none]
def people_count(plan):
    return plan["people_number"]
        \end{lstlisting}  \\
        \midrule
        start\_city & start city of the plan & \begin{lstlisting}[style=pystyle, frame=none] 
def start_city(plan):
    return plan["start_city"]
        \end{lstlisting} \\
        \midrule
        target\_city & target city of the plan & \begin{lstlisting}[style=pystyle, frame=none] 
def target_city(plan):
    return plan["target_city"]
        \end{lstlisting} \\
        \midrule
        allactivities & all the activities in the plan & \begin{lstlisting}[style=pystyle, frame=none]
def allactivities(plan):
    activity_list = []
    for day_activity in plan["itinerary"]:
        for act in day_activity["activities"]:
            activity_list.append(act)
    return activity_list            
        \end{lstlisting} \\
        \midrule
        allactivities\_- count & the number of activities in the plan & \begin{lstlisting}[style=pystyle, frame=none]
def allactivities_count(plan):
    count = 0
    for day_activity in plan["itinerary"]:
        count += \ 
            len(day_activity["activities"])
    return count          
        \end{lstlisting} \\
        \midrule
         dayactivities & all the activities in the specific day [1, 2, 3, ...] & \begin{lstlisting}[style=pystyle, frame=none]
def dayactivities(plan, day):
    activity_list = []
    for act in plan["itinerary"]\
        [day - 1]["activities"]:
        activity_list.append(act)
    return activity_list         
        \end{lstlisting} \\
        \midrule
        activity\_cost & the cost of specific activity without transport cost & \begin{lstlisting}[style=pystyle, frame=none]
def activity_cost(activity):
    return activity.get("cost", 0)
        \end{lstlisting}  \\
        \midrule
        activity\_posi\-tion & the position name of specific activity & \begin{lstlisting}[style=pystyle, frame=none]
def activity_position(activity):
    return activity.get("position", "")
        \end{lstlisting}  \\
        \midrule
        activity\_price & the price of specific activity & \begin{lstlisting}[style=pystyle, frame=none]
def activity_price(activity):
    return activity.get("price", 0)
        \end{lstlisting}  \\
        \midrule
        activity\_type & the type of specific activity & \begin{lstlisting}[style=pystyle, frame=none]
def activity_type(activity):
    return activity.get("type", "")
        \end{lstlisting}  \\
        \midrule
        activity\_tickets & the number of tickets needed for specific activity & \begin{lstlisting}[style=pystyle, frame=none]
def activity_tickets(activity):
    return activity.get("tickets", 0)
        \end{lstlisting}  \\
        \midrule
        activity\_trans\-ports & the transport information of specific activity & \begin{lstlisting}[style=pystyle, frame=none]
def activity_transports(activity):
    return activity.get("transports", [])
        \end{lstlisting}  \\
        \midrule
        activity\_- start\_time & the start time of specific activity & \begin{lstlisting}[style=pystyle, frame=none]
def activity_start_time(activity):
    return activity.get("start_time")
        \end{lstlisting}  \\
        \midrule
        activity\_-end\_time & the end time of specific activity & \begin{lstlisting}[style=pystyle, frame=none]
def activity_end_time(activity):
    return activity.get("end_time")
        \end{lstlisting}  \\
         \bottomrule
    \end{tabular}
    \caption{Concept Function}
    \label{tab_concept_func_0}
\end{table*}

\begin{table*}[p]
    \centering
    \vspace*{-1cm}
    \small
    % \hspace*{-1.5cm}
    \begin{tabular}{p{2cm}p{2cm}l}
        \toprule
        Function Name & Meaning & Implementation \\
        \midrule
        \midrule
        activity\_time & the duration of specific activity & \begin{lstlisting}[style=pystyle, frame=none]
def activity_time(activity):
    start_time = activity.get("start_time")
    end_time = activity.get("end_time")
    if start_time and end_time:
        st_h, st_m = \
            map(int, start_time.split(":"))
        ed_h, ed_m = \
            map(int, end_time.split(":"))
        return \
            (ed_m - st_m) + (ed_h - st_h) * 60
    return -1
        \end{lstlisting}  \\
        \midrule
        poi\_recom\-mend\_time & the recommend time of specific poi(attraction) in the city & \begin{lstlisting}[style=pystyle, frame=none]
def poi_recommend_time(city, poi):
    select = Attractions().select
    attrction_info = \
        select(city, key="name", 
               func=lambda x: x == poi).iloc[0]
    recommend_time = \
        (attrction_info["recommendmintime"]) \
        * 60
    return recommend_time
        \end{lstlisting}  \\
        \midrule
        poi\_distance & the distance between two POIs in the city & \begin{lstlisting}[style=pystyle, frame=none]
def poi_distance(city, poi1, poi2):
    start_time="00:00"
    transport_type="walk"
    goto = Transportation().goto
    return goto(city, poi1, poi2, start_time, 
                transport_type)[0]["distance"]
        \end{lstlisting}  \\
        \midrule
         innercity\_-transport\_cost & the total cost of specific innercity transport  & \begin{lstlisting}[style=pystyle, frame=none]
def innercity_transport_cost(transports, mode):
    cost = 0
    for transport in transports:
        if node is None or \
            transport.get("type") == node:
            cost += transport.get("cost", 0)
    return cost
        \end{lstlisting}  \\
        \midrule
          innercity\_-transport\_price & the price of innercity transport & \begin{lstlisting}[style=pystyle, frame=none]
def innercity_transport_price(transports):
    price = 0
    for transport in transports:
        price += transport["price"]
    return price
        \end{lstlisting}  \\
        \midrule
          innercity\_-transport\_-distance & the distance of innercity transport & \begin{lstlisting}[style=pystyle, frame=none]
def innercity_transport_distance\
    (transports, mode=None):
    distance = 0
    for transport in transports:
        if mode is None or \
            transport.get("type") == mode:
            distance += \
                transport.get("distance", 0)
    return distance
        \end{lstlisting}  \\
        \midrule
          innercity\_-transport\_-time & the duration of innercity transport & \begin{lstlisting}[style=pystyle, frame=none]
def innercity_transport_time(transports):
    def calc_time_delta(end_time, start_time):
        hour1, minu1 = \
            int(end_time.split(":")[0]), \
                int(end_time.split(":")[1])
        hour2, minu2 = \
            int(start_time.split(":")[0]), \
                int(start_time.split(":")[1])
        return (hour1 - hour2) * 60\
              + (minu1 - minu2)
        \end{lstlisting}  \\
         \bottomrule
    \end{tabular}
    \caption{Concept Function}
    \label{tab_concept_func_1}
\end{table*}

\begin{table*}[p]
    \centering
    \vspace*{-1cm}
    \small
    % \hspace*{-1.5cm}
    \begin{tabular}{p{2cm}p{2cm}l}
        \toprule
        Function Name & Meaning & Implementation \\
        \midrule
        \midrule
        metro\_tickets & the number of metro tickets if the type of transport is metro & \begin{lstlisting}[style=pystyle, frame=none]
def metro_tickets(transports):
    return transports[1]["tickets"]
        \end{lstlisting}  \\
        \midrule
        taxi\_cars & the number of taxi cars if the type of transport is taxi & \begin{lstlisting}[style=pystyle, frame=none]
def taxi_cars(transports):
    return transports[0]["cars"]
        \end{lstlisting}  \\
        \midrule
        room\_count & the number of rooms of accommodation & \begin{lstlisting}[style=pystyle, frame=none]
def room_count(activity):
    return activity.get("rooms", 0)
        \end{lstlisting}  \\
        \midrule
        room\_count & the number of rooms of accommodation & \begin{lstlisting}[style=pystyle, frame=none]
def room_count(activity):
    return activity.get("rooms", 0)
        \end{lstlisting}  \\
        \midrule
        room\_type & the type of room of accommodation & \begin{lstlisting}[style=pystyle, frame=none]
def room_type(activity):
    return activity.get("room_type", 0)
        \end{lstlisting}  \\
        \midrule
        restaurant\_-type & the type of restaurant's cuisine in the target city & \begin{lstlisting}[style=pystyle, frame=none]
def restaurant_type(activity, target_city):
    restaurants = Restaurants()
    select_food_type = \
        restaurants.select(
        target_city, key="name", 
        func=lambda x: x == activity["position"]
    )["cuisine"]
    if not select_food_type.empty:
        return select_food_type.iloc[0]
    return ""
        \end{lstlisting}  \\
        \midrule
        attraction\_-type & the type of attraction in the target city & \begin{lstlisting}[style=pystyle, frame=none]
def attraction_type(activity, target_city):
    attractions = Attractions()
    select_attr_type = \
        attractions.select(
        target_city, key="name", 
        func=lambda x: x == activity["position"]
    )["type"]
    if not select_attr_type.empty:
        return select_attr_type.iloc[0]
    return ""
        \end{lstlisting}  \\
        \midrule
        accommo\-dation\_type & the feature of accommodation in the target city & \begin{lstlisting}[style=pystyle, frame=none]
def accommodation_type(activity, target_city):
    accommodations = Accommodations()
    select_hotel_type = \
        accommodations.select(
        target_city, key="name", 
        func=lambda x: x == activity["position"]
    )["featurehoteltype"]
    if not select_hotel_type.empty:
        return select_hotel_type.iloc[0]
    return ""
        \end{lstlisting}  \\
         \bottomrule
    \end{tabular}
    \caption{Concept Function}
    \label{tab_concept_func_2}
\end{table*}

\begin{table*}[th]
    \centering
    % \vspace*{-1cm}
    \small
    % \hspace*{-1.5cm}
    \begin{tabular}{p{2cm}p{2cm}l}
        \toprule
        Function Name & Meaning & Implementation \\
        \midrule
        \midrule
        innercity\_-transport\_-type & the type of innercity transport & \begin{lstlisting}[style=pystyle, frame=none]
def innercity_transport_type(transports):
    if len(transports) == 3:
        return transports[1]["mode"]
    elif len(transports) == 1:
        return transports[0]["mode"]
    return ""
        \end{lstlisting}  \\
        \midrule
        intercity\_-transport\_-type & the type of intercity transport & \begin{lstlisting}[style=pystyle, frame=none]
def intercity_transport_type(activity):
    return activity.get("type", "")
        \end{lstlisting}  \\
        \midrule
        innercity\_-transport\_-start\_time & the start time of innercity transport & \begin{lstlisting}[style=pystyle, frame=none]
def innercity_transport_start_time(transports):
    return transports[0]["start_time"]
        \end{lstlisting}  \\
        \midrule
        innercity\_-transport\_-end\_time & the end time of innercity transport & \begin{lstlisting}[style=pystyle, frame=none]
def intercity_transport_end_time(transports):
    return transports[-1]["end_time"]
        \end{lstlisting}  \\
        \midrule
        intercity\_\-transport\_\-origin & the origin city of intercity transport & \begin{lstlisting}[style=pystyle, frame=none]
def intercity_transport_origin(activity):
    if "start" in activity:
        for city in city_list:
            if city in activity["start"]:
                return city
    return ""
        \end{lstlisting}  \\
        \midrule
        intercity\_\-transport\_\-destination & tthe destination city of intercity transport & \begin{lstlisting}[style=pystyle, frame=none]
def intercity_transport_destination(activity):
    if "end" in activity:
        for city in city_list:
            if city in activity["end"]:
                return city
    return ""
        \end{lstlisting}  \\

         \bottomrule
    \end{tabular}
    \caption{Concept Function}
    \label{tab_concept_func_3}
\end{table*}

\begin{figure*}
    \centering
    \vspace{-.3in}
    \includegraphics[width=\linewidth]{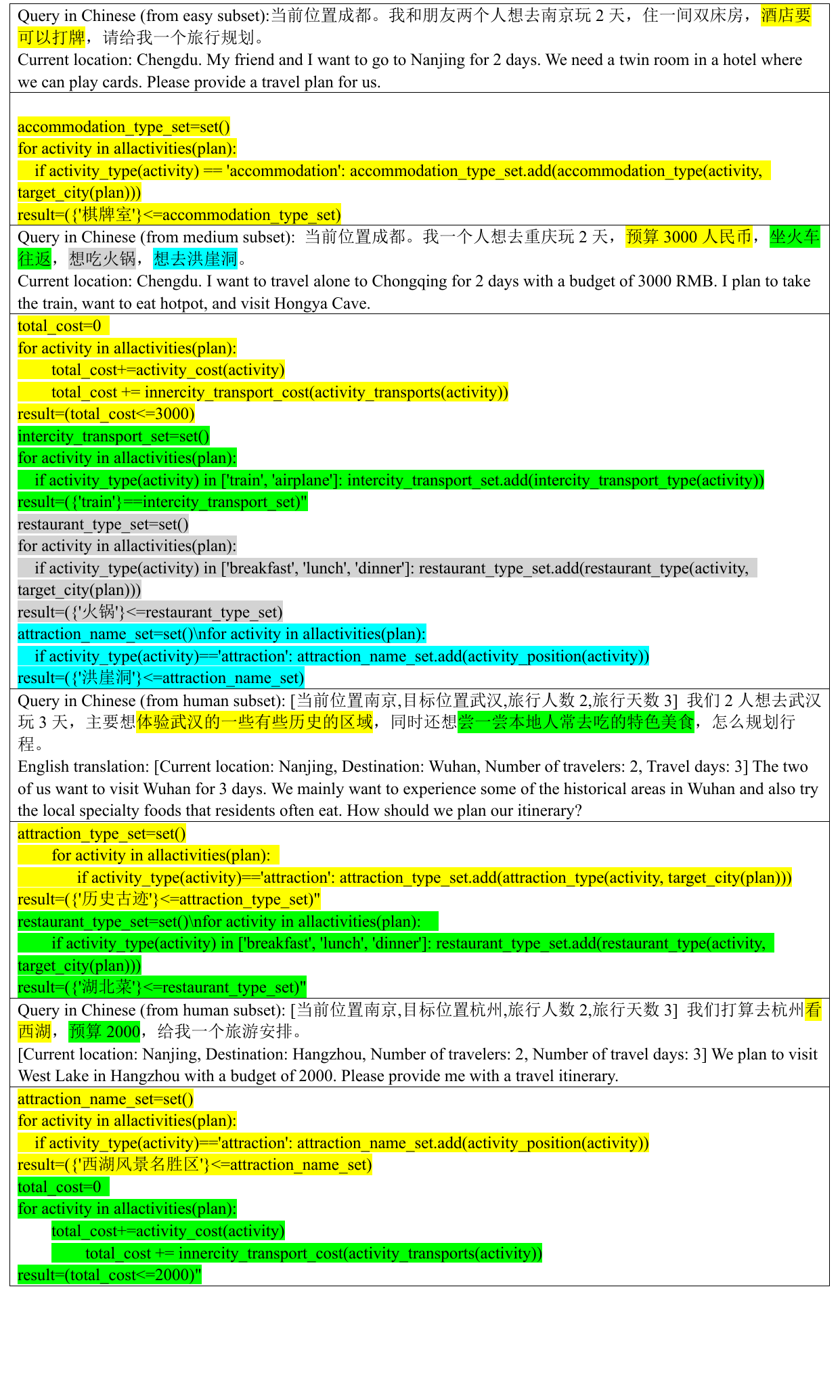}
    \vspace{-.8in}
    \caption{Examples of travel requirements and their DSL expressions.}
    \label{dsl_example}
\end{figure*}

\subsection{Query Synthesis}
\label{app_querygen}
We designed common travel information (origin, destination, days, number of people) and logical constraints based on the nature of travel tasks. To facilitate scalable queries for ChinaTravel, we randomly constructed query skeletons from the aforementioned information and used advanced LLMs to generate natural language queries from these skeletons. In practice, we provide the LLMs with more intuitive hard logic constraints to ensure the LLMs do not make mistakes and use a Python script to convert it after generating the query. 
The automatically generated data is categorized into two difficulty levels: In the \textit{Easy} level, user inputs encompass a single logical requirement, sourced from categories such as transportation, restaurants, attractions, and accommodations. In the \textit{Medium} level, user inputs involve 2 to 5 logical requirements, introducing more complex constraints. During the generation, we encourage the LLMs to provide varied and human-like expressions, necessitating a deeper understanding and processing to accurately interpret and fulfill the user's needs. 
For instance, the logical requirement ``taste Beijing cuisine" could correspond to the natural language query: ``Try local food in Beijing." We utilize prompt engineering to guide LLMs in refining natural language expressions to facilitate automated generation.
One of the prompts is shown in Figure \ref{fig_data_generation_prompt}. Several examples of generated data is in Figure \ref{fig_data_example}. As a result, we obtain the synthetic queries across diverse travel requirements, including 28 restaurant types, 23 attraction types, 29 hotel features, and more than 130 specific POIs.

\subsection{Diversity of Synthetic Data and Bias Mitigation}

\label{app:bias}
This subsection provides a detailed analysis of ChinaTravel’s hybrid query design, addressing concerns about synthetic data limitations and real-world representativeness. 
% ChinaTravel integrates both synthetic and human-authored queries to balance scalability and realism. 
When synthesizing data, we randomly constructed constraints based on the types and specific visit requirements of POIs such as restaurants, accommodations, transports, and attractions, thereby ensuring the diversity of the dataset. 
% The human query subset comprises 154 samples collected through structured questionnaires, which introduce complex real-world constraints such as time-bound returns (e.g., explicit requirements like ``return before 7 PM") and activity-specific budget allocations. These queries also incorporate colloquial expressions that reflect native Chinese travel preferences, exemplified by phrases like local specialty foods frequented by residents. 
The synthetic queries are generated through LLM-based paraphrasing techniques and systematically categorized into two tiers: Easy-tier queries contain single logical constraints (e.g., specific cuisine requirements), while Medium-tier queries combine 3--5 interdependent constraints (e.g., compound conditions like ``budget $\leq$ 3000 CNY + train transport + hotpot dining").

To mitigate synthetic data bias and enhance diversity, two strategies were implemented. First, constraint combinations were deliberately diversified across temporal, spatial, and cost dimensions, as detailed in Table~\ref{logical_constraints}. Second, a human validation layer filters out unrealistic LLM-generated queries, such as physically implausible itineraries like "visiting 10 attractions within one day." 

% The current human query subset remains limited by annotation costs, as discussed in the limitation section. In future work, we will advance data collection by integrating LLMs with real user queries to automate and diversify the generation of human-like queries. Additionally, all human queries and automated synthesis tools will be publicly released to support community-driven benchmark extensions. 

\subsection{DSL Expression for Preference}
\label{preference_example}

We introduce six common preferences from user surveys to construct the preference sub-datasets. Table~\ref{preference_req_des} provides a summary of these preferences. 

The corresponding DSL could be formulated as follows. 
\begin{lstlisting}[style=pystyle]
# The number of attractions visited
count = 0
for act_i in all_activities(plan):
  if activity_type(act_i)=="attraction": count = count + 1
return count
\end{lstlisting}

\begin{figure*}[th]
\begin{tcolorbox}[
                standard jigsaw,
                % colback=white, % 背景颜色
                % colframe=white, % 边框颜色
                coltitle = black,
                colbacktitle = white,
                title=An Example of Prompts for Data Generation,     % 标题
                opacityback=0,
                fonttitle=\bfseries,   % 标题字体
                % halign=left,         
                % valign=center,         
                % colupper=black,        % 内容文字颜色
                % width=\linewidth,      % 设置宽度
                 % height=5cm,            % 设置高度
]

% \begin{CJK}{UTF8}{gbsn}
% 你是一个用户，你想请ai制定一个旅行规划，请根据以下的例子构建一些自然语言的询问，并提供对应的逻辑约束表达。注意tickets和people\_number一样。\\
% 例子：\\
% JSON：\\ 
% \{ \\
%     \hspace*{2em}"start\_city": "北京", \\
%     \hspace*{2em}"target\_city": "南京",\\
%     \hspace*{2em}"hard\_logic": [\\
%     \hspace*{4em}    "days==2",\\
%     \hspace*{4em}    "people\_number==1",\\
%     \hspace*{4em}    "tickets==1",\\
%     \hspace*{4em}    "\{'南京大排档'\} <= restaurant\_names",\\
%     \hspace*{2em}],\\
%     \hspace*{2em}"nature\_language": "当前位置北京。我一个人想去南京玩2天，想吃南京大排档，请给我一个旅行规划。"\\
% \} \\
% 使用如下的餐饮。\\
% 店名：\{\} \\
% 即要求restaurant\_names包含这个店。\\
% 注意，餐饮不一定完全按照提供的特征的名字来，可以使用近义词，比如如果提供的是泳池，可以使用想在酒店游泳这样的自然语言询问 \\
% 注意，你现在的出发地点为\{\},目标地点为\{\}。人数\{\},天数\{\} \\
% 现在请给一个json询问, \\
% JSON：\\
% \end{CJK}

\begin{lstlisting}[frame=none]
# You are a user who wants to ask an AI agent to help you plan a trip. Please construct some natural language inquiries based on the following example and provide the corresponding logical constraint expressions. Note that "tickets" and "people_number" are the same.
# Example:
# JSON:
# {}
# Use the following restaurants.
# Restaurant name: {}
# This means that "restaurant_names" should include this restaurant.
# The dining options may not always be exactly as described by the provided features; synonyms can be used. For example, if the hotel's feature is a pool, you could ask naturally in language like "I want to swim in the hotel pool."
# Now, your departure location is {}, and your destination is {}. The number of people is {}, and the number of days is {}.
# Now please provide a JSON inquiry.
# JSON:
\end{lstlisting}
\end{tcolorbox}
\caption{An example of prompts for data generation. This example is about restaurant\_name. By replacing this with other constraints or combining multiple constraints, we can generate data with different levels of difficulty based on different constraints.}
\label{fig_data_generation_prompt}
\end{figure*}
\newpage

\begin{lstlisting}[style=pystyle]
# The average travel time between POIs
time_cost = 0
transport_count = 0
for activity in allactivities(plan):       transports =activity_transports(activity)
    transport_count += 1        time_cost += innercity_transport_time(transports)
average_time_cost = time_cost / transport_count if transport_count > 0 else -1
return average_time_cost
\end{lstlisting}
\begin{lstlisting}[style=pystyle]
# The average travel time to restaurants
restaurant_count = 0
time_cost = 0
for activity in allactivities(plan):
    if activity_type(activity) in ['breakfast', 'lunch', 'dinner']:
        restaurant_count += 1
        time_cost += innercity_transport_time(activity_transports(activity))
if restaurant_count == 0:
    average_time_cost = -1
else:
    average_time_cost = time_cost / restaurant_count
return average_time_cost
\end{lstlisting}
\begin{lstlisting}[style=pystyle]
# The ratio of food cost
food_cost = 0
total_cost = 0
for activity in allactivities(plan):
    total_cost += activity_cost(activity)
    total_cost +=innercity_transport_cost(activity_transports(activity))
    if activity_type(activity) in ['breakfast', 'lunch', 'dinner']:
        food_cost += activity_cost(activity)
food_cost_ratio = food_cost / total_cost if total_cost > 0 else -1
return food_cost_ratio
\end{lstlisting}
\begin{lstlisting}[style=pystyle]
# The cost of accommodations
accommodation_cost = 0
for activity in allactivities(plan):
    if activity_type(activity) == 'accommodation':
        accommodation_cost += activity_cost(activity)"
return accommodation_cost
\end{lstlisting}
\begin{lstlisting}[style=pystyle]
# The average distance to poi (e.g. xxx)
target_poi = 'xxx'
poi_list = list()
total_distance = 0
poi_count=0
city = target_city(plan)
for activity in allactivities(plan):
    if activity_type(activity) in ['breakfast', 'lunch', 'dinner', 'accommodation', 'attraction']:
        poi_list.append(activity_position(activity))
for poi in poi_list:
    total_distance += poi_distance(city, target_poi, poi)
    poi_count += 1
average_dist_cost = total_distance / poi_count if poi_count > 0 else -1
return average_dist_cost
\end{lstlisting}

\begin{figure*}[thp]
\begin{tcolorbox}[
                standard jigsaw,
                % colback=white, % 背景颜色
                % colframe=white, % 边框颜色
                coltitle = black,
                colbacktitle = white,
                title=Examples of Generated Data,     % 标题
                opacityback=0,
                fonttitle=\bfseries,   % 标题字体
                % halign=left,         
                % valign=center,         
                % colupper=black,        % 内容文字颜色
                % width=\linewidth,      % 设置宽度
                 % height=5cm,            % 设置高度
]
\begin{CJK}{UTF8}{gbsn}
\textbf{Example 1} \\
\{ \\
    \hspace*{2em}"start\_city": "杭州", \\
    \hspace*{2em}"target\_city": "上海", \\
    \hspace*{2em}"hard\_logic": [ \\
        \hspace*{4em}"days==2", \\
        \hspace*{4em}"people\_number==1", \\
        \hspace*{4em}"tickets==1", \\
        \hspace*{4em}"\{'本帮菜'\} $\leq$ food\_type" \\
    \hspace*{2em}], \\
    \hspace*{2em}"nature\_language": "当前位置杭州。我一个人想去上海玩2天，想尝试当地的特色菜，请给我一个旅行规划。" \\
\} \\
\newline
\textbf{Example 2} \\
\{\\
    \hspace*{2em}"start\_city": "深圳",\\
    \hspace*{2em}"target\_city": "北京",\\
    \hspace*{2em}"hard\_logic": [\\
        \hspace*{4em}"days==2",\\
        \hspace*{4em}"people\_number==3",\\
        \hspace*{4em}"intercity\_transport==\{'airplane'\}",\\
        \hspace*{4em}"tickets==3",\\
        \hspace*{4em}"rooms==3",\\
        \hspace*{4em}"room\_type==1"\\
    \hspace*{2em}],\\
    \hspace*{2em}"nature\_language": "当前位置深圳。我们三个人计划去北京玩两天，选择飞机出行，开三间大床房。请给我一个旅行规划。"\\
\}\\
\newline
\textbf{Example 3} \\
\{ \\
    \hspace*{2em}"start\_city": "重庆", \\
    \hspace*{2em}"target\_city": "苏州", \\
    \hspace*{2em}"hard\_logic": [ \\
        \hspace*{4em}"days==3", \\
        \hspace*{4em}"people\_number==3", \\
        \hspace*{4em}"cost$\leq$7300", \\
        \hspace*{4em}"\{'日本料理'\} $\leq$ food\_type", \\
        \hspace*{4em}"intercity\_transport==\{'train'\}", \\
       \hspace*{4em} "tickets==3", \\
        \hspace*{4em}"rooms==2", \\
        \hspace*{4em}"room\_type==2" \\
    \hspace*{2em}], \\
    \hspace*{2em}"nature\_language": "当前位置重庆。我们三个人计划去苏州玩三天，选择火车出行，想吃日本料理，预算7300元，开两间双床房。请给我一个旅行规划。" \\
\}
\end{CJK}

\end{tcolorbox}
\caption{Examples of Generated Data}
\label{fig_data_example}
\end{figure*}

\subsection{Benchmark Difficulty and Applicability} 
While the Human subset presents significant challenges, the baseline NeSy solution has achieved 60.6\% and 46.7\% FPR on Easy and Medium subsets, respectively, providing developers with actionable validation points for initial testing and refinement. Additionally, the Human subset’s extreme difficulty arises from open language reasoning and unseen concept composition, key challenges absent in prior benchmarks but unavoidable in practice. By explicitly formalizing these challenges, ChinaTravel has provided a roadmap for advancing agents toward real-world robustness. Despite current LLMs' limitations in handling unseen combinations, their success in code generation suggests that post-training with DSL may enhance their understanding of diverse travel needs, moving toward real-world applications.

\subsection{\blue{DSL Extension}}
\blue{The design of DSL is a \textbf{modular, domain-agnostic framework} whose \textbf{core operators are reusable} beyond the current instantiation. Concretely, it separates generic compositional operators, logical, arithmetic, set, and temporal constructs, from a pluggable library of domain-specific predicates and attribute-access functions. Extending the constraint library to include new concepts, such as `a scenic rating of a route' or `avoid areas with high COVID-19 cases', is a straightforward, two-step, incremental process, not a framework overhaul.} 

\blue{(1). \textbf{Sandbox extensio}. Integrate the new attribute into the sandbox by adding a corresponding field to the relevant entities. For example, to support scenic beauty of a route, we could add a numeric \texttt{scenic\_rating} attribute to attraction entries, to model avoid areas with high COVID-19 cases, we can add a boolean \texttt{covid\_risk} field to POIs.} 

\blue{(2) \textbf{DSL function definition.}
Expose this attribute through a small helper function or predicate in the DSL library
(e.g., \texttt{get\_scenic\_rating(attraction)} or \texttt{get\_covid\_risk(POI)}).
User requests such as ``prefer scenic routes'' and ``avoid covid risk'' can then be
rendered as constraints such as:}
\begin{lstlisting}[style=pystyle]
# maximize scenic_score_sum as a soft preference
scenic_score_sum=0
for act_i in all_activities(plan):
  if activity_type(act_i)=="attraction": scenic_score_sum += activity_position(act_i)
return scenic_score_sum
\end{lstlisting}
\begin{lstlisting}[style=pystyle]
# avoid covid risk as a hard constraint
risk_flag =0
for act_i in all_activities(plan):
  risk_flag += get_covid_risk(activity_position(act_i))
return (risk_flag==0)
\end{lstlisting}
\blue{These two steps correspond exactly to \textbf{make the information available} and textbf{provide a way for the agent to query it}. They are both textbf{necessary and close to minimal}, no changes are required to the core DSL grammar, compositional operators, planner, or verification engine.} 

\blue{Moreover, new concepts can naturally be combined with existing temporal and structural concepts to express richer user requirements, like `visit the most scenic attraction in the itinerary on day 1 or avoid COVID-risk restaurants on day 1 and COVID-risk attractions on day 2'. 
ChinaTravel is explicitly designed to make such user-friendly, open-ended compositional constraints representable and automatically checkable, and we hope this will draw the community's attention to these more realistic forms of constraint-aware LLM agents.}

\section{Discussion with TravelPlanner} 

TravelPlanner’s logical constraints contain the total cost, 15 cuisines, 5 house rules, 3 room types, and 3 intercity transports. ChinaTravel's logical constraints contain the total cost, 42 cuisines, 15 attraction types, 78 hotel features, 2 room types, 2 intercity-transports types, 3 inner-city-transports types, and specific POI names (attractions, restaurants, hotels). Crucially, our benchmark introduces compositional constraints derived from human queries (e.g., ``return before 7 PM", ``cost of intercity transports"), reflecting real-world complexity. The key advancement lies in addressing open-language reasoning and unseen concept composition, which represent significant challenges beyond TravelPlanner’s scope. Our Domain-Specific Language (DSL) enables automated validation of these combinatorial requirements, bridging the gap between synthetic and real-world needs.

\begin{figure*}[ht]
    % \centering
    \includegraphics[width=\textwidth]{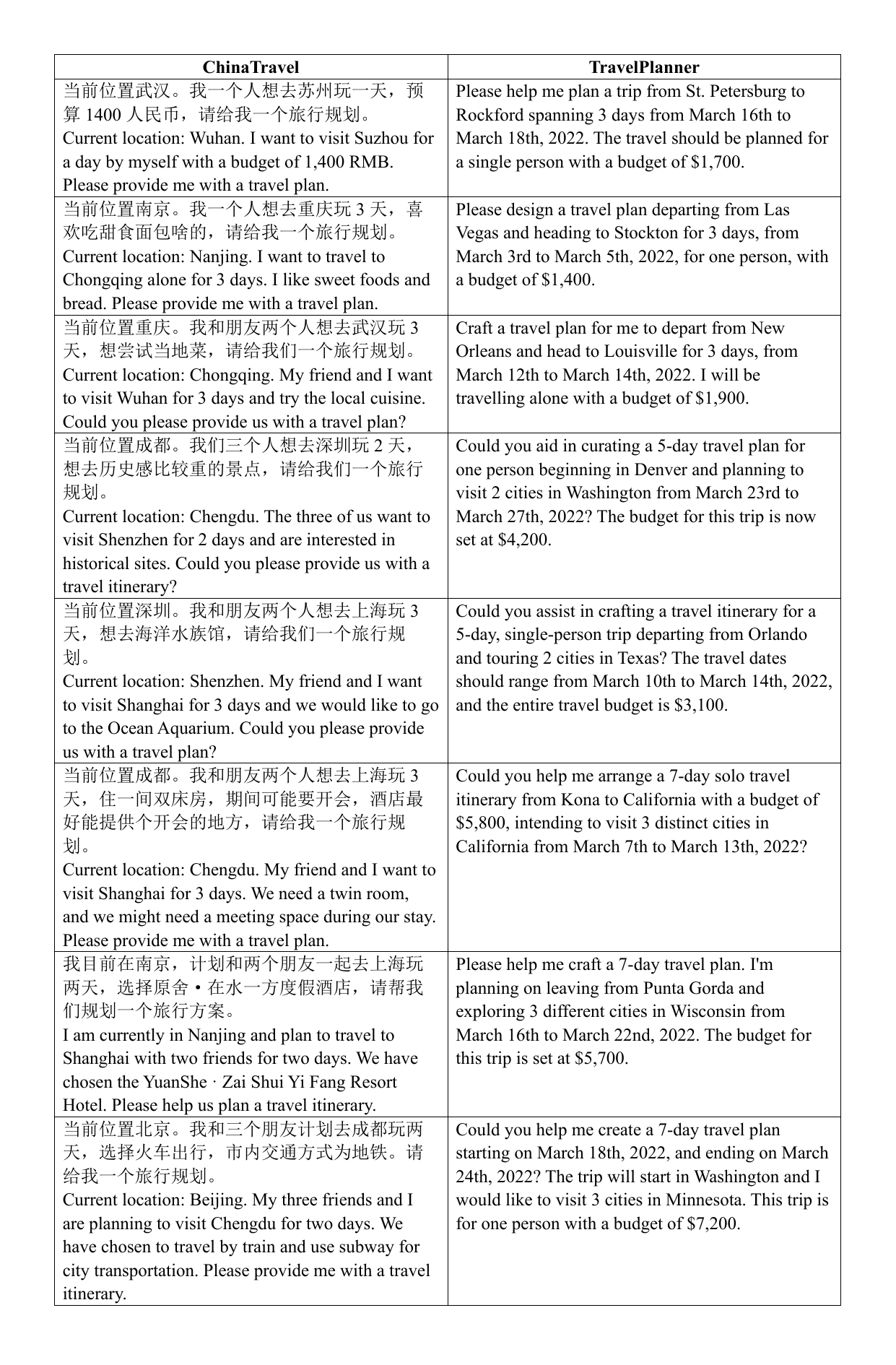}
    \caption{Examples of easy-level queries from ChinaTravel and TravelPlanner.}
    \label{fig_easy_query}
\end{figure*}

\begin{figure*}[ht]
    % \centering
    \includegraphics[width=\textwidth]{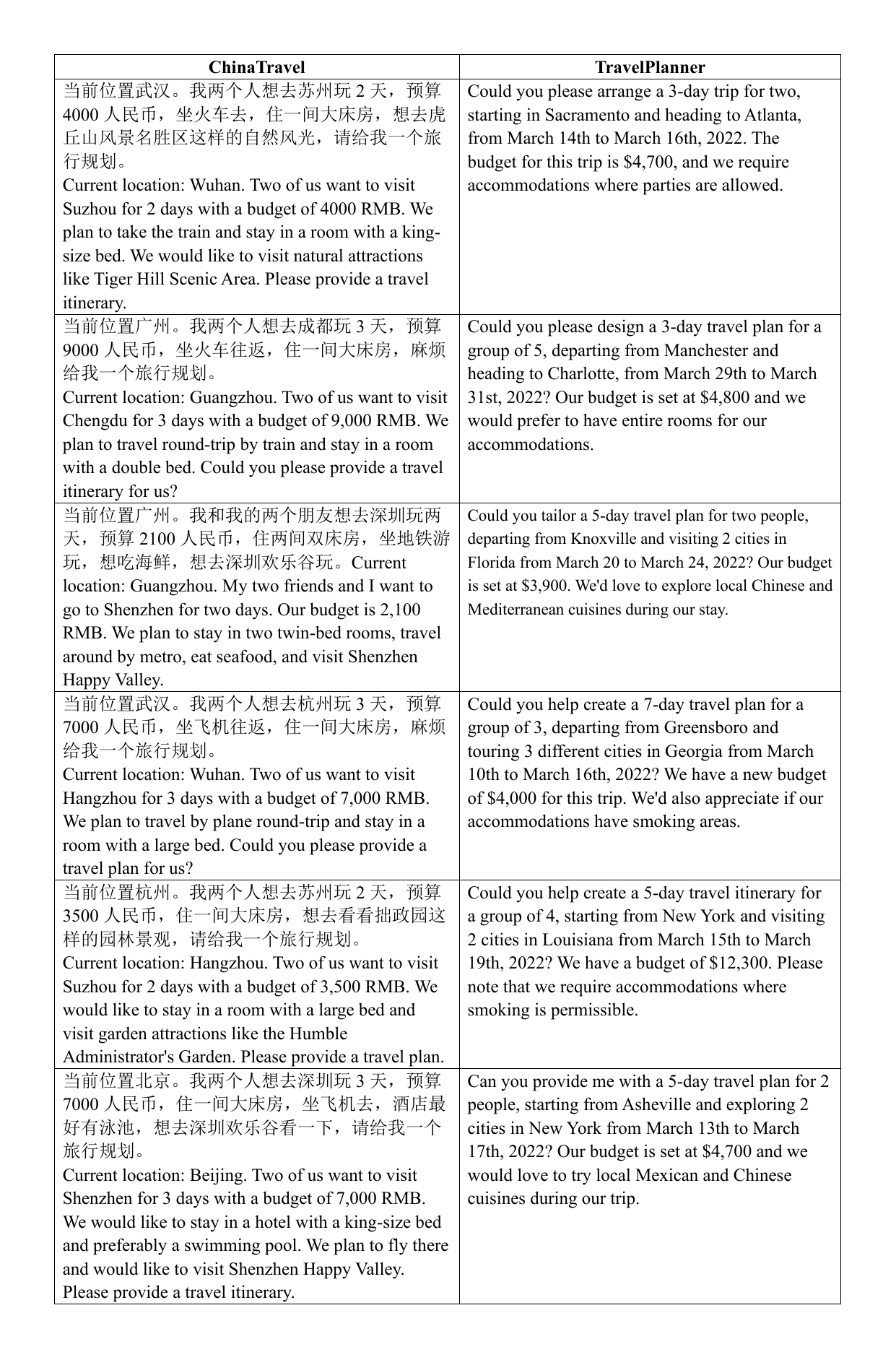}
    \caption{Examples of medium-level queries from ChinaTravel and TravelPlanner.}
    \label{fig_medium_query}
\end{figure*}

\begin{figure*}[ht]
    % \centering
    
    \vspace{-.7in}
    \includegraphics[width=\textwidth]{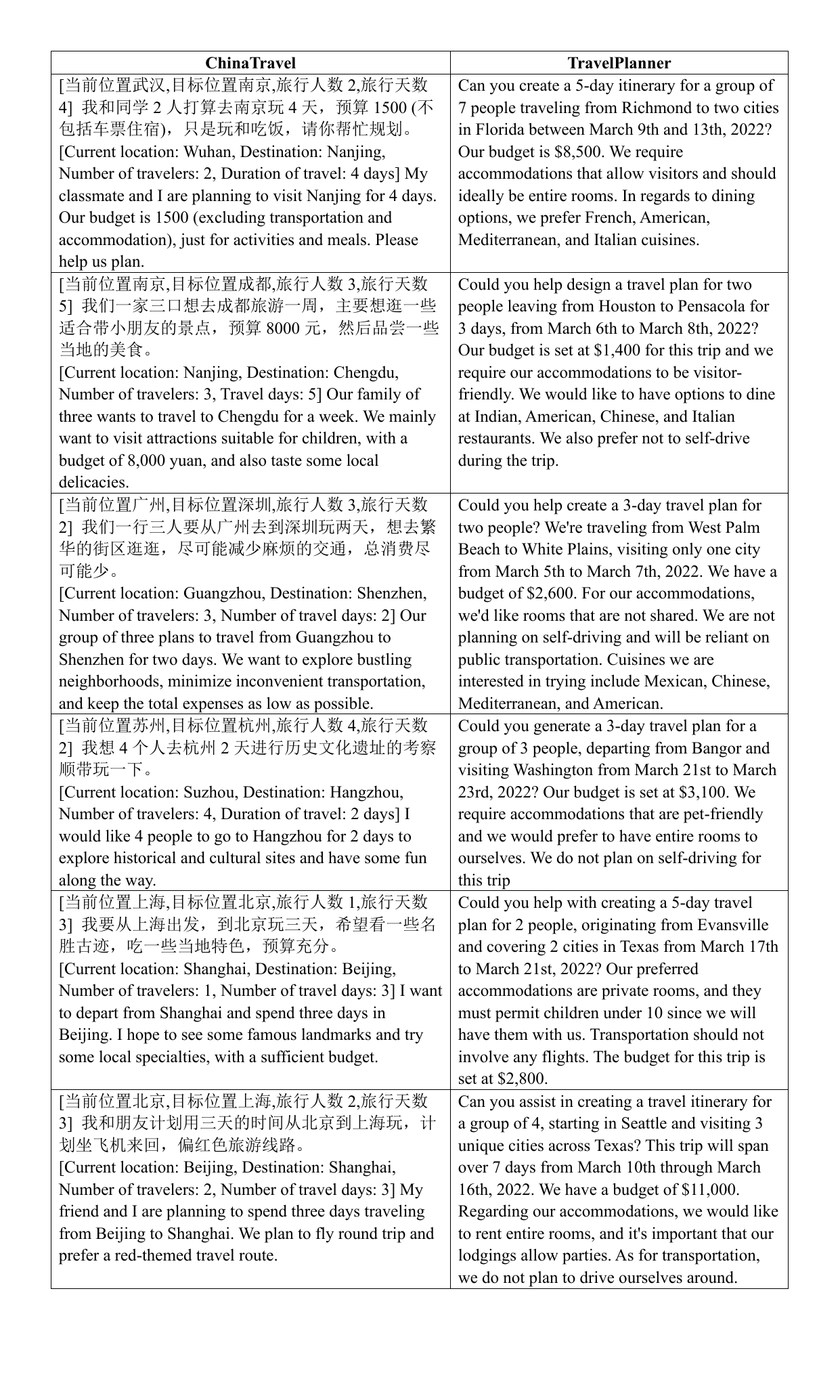}
    \vspace{-1in}
    \caption{Examples of human/hard level queries from ChinaTravel and TravelPlanner.}
    \label{fig_hard_query}
\end{figure*}

We also provide some example queries and corresponding examples from the TravelPlanner at each level in Figure~\ref{fig_easy_query},~\ref{fig_medium_query}, and \ref{fig_hard_query}. 

As shown in Figure~\ref{fig_easy_query}, in the ``easy level", TravelPlanner only includes constraints on cost. In contrast, ChinaTravel demonstrates significant advantages over TravelPlanner, particularly in terms of personalized support for specific Points of Interest (POIs) and more realistic transportation and time management. These advantages are crucial for developing and evaluating language agents that can handle real-world travel planning scenarios effectively. ChinaTravel allows users to directly specify POI names, such as "Nanjing DaPaXiang" or "HuQiu Mountain Scenic Area," requiring the agent to precisely match the entity information from the travel sandbox. 

As shown in Figure~\ref{fig_medium_query}, in the medium set, TravelPlanner includes queries with 2 types of constraints: cost and cuisine, or cost and accommodation. In contrast, ChinaTravel includes queries with 2-5 constraints, reflecting more complex and diverse multi-constraint requirements. This difference highlights the ability of ChinaTravel to handle more realistic and varied travel planning scenarios.

As shown in Figure~\ref{fig_hard_query}, TravelPlanner includes queries with multiple constraints, such as cost, accommodation type, and cuisine preferences. However, ChinaTravel goes a step further by including queries with unseen logical constraints and more colloquial expressions. These unseen logical constraints and colloquial expressions are essential for travel planning agents to handle real-world users effectively. They reflect the complexity and diversity of real-world travel planning scenarios, where users may have diverse requirements that need to be understood and addressed. By incorporating these elements, ChinaTravel bridges the gap between current academic research benchmarks and real-world application demands, making it a more comprehensive and realistic benchmark for evaluating the capabilities of travel planning agents. 

\blue{Here, we further provide a performance comparison across TripPlanning~\citet{zheng2024natural}, TravelPlanner~\citet{TravelPlanner} (Val-180) and our ChinaTravel (Human-154).}
\begin{table}[tbh]
\centering
\footnotesize
\begin{tabular}{l l| c c c}
\toprule
\textbf{Method} & \textbf{Model} & \textbf{TripPlanning} & \textbf{TravelPlanner} & \textbf{ChinaTravel} \\
\midrule
Pure-LLM &  & 37.1 & 4.44 & 2.59 \\
         & GPT & 31.1 (GPT-4) & 4.4 (GPT-4-Turbo) & 0 (GPT-4o) \\
\midrule
NeSy & TTG(DS-V3) & - & 91.7 & 1.29 \\
     & LLM-Modulo(DS-V3) & 98.5 & 25.55 & 2.59 \\
\bottomrule
\end{tabular}
\caption{\blue{Performance comparison across benchmarks.}}
\label{tab_comparison_benchmark}
\end{table}

\blue{As shown in the Tab.~\ref{tab_comparison_benchmark}, we could find that: (1) \textbf{Catastrophic Failure of Pure LLMs}: While Pure LLMs show decent performance on TripPlanning (DeepSeek 37.1\%, GPT-4 31.1\%), a pure reasoning task, their success rate dramatically drops to around 4.4\% when tool-calling is introduced in TravelPlanner. Moreover, when facing the compositional complexity and open-ended nature of ChinaTravel, LLM performance collapses to near-zero (e.g., GPT-4o achieves 0\%). This highlights that both TravelPlanner and ChinaTravel poses an agentic challenges, which existing LLMs cannot handle, as we claimed in the paper. (2) \textbf{Failure of SOTA Neuro-Symbolic Methods}: TTG excels on TravelPlanner (91.7\%) because its symbolic logic is a good fit for TravelPlanner's fixed and predefined constraints. However, TTG's success rate plummets to 1.29\% on ChinaTravel. As we analyze in Sec. 4.2 and Fig. 6a, this confirms that TTG's constrained symbolic system cannot generalize to long-horizon planning required by ChinaTravel.
LLM-Modulo demonstrates improvement over pure LLM on TravelPlanner (4.4\%$\rightarrow$25.55\%) via symbolic constraint feedback, but still fails on ChinaTravel (2.59\%). This again validates our argument: ChinaTravel is not merely a harder version of existing benchmarks, it requires a new level of planning difficulty that current SOTA methods lack. 
These results unequivocally confirm that ChinaTravel introduces a new open-ended dimension that exposes the limits of both pure LLM and current neuro-symbolic agent designs, thus strongly validating its contribution as a novel, challenging benchmark.
}

\section{NeSy Planning}
\label{app_nesy}

Since the Z3 solver from~\citep{MITSMT} would restructure the tool API to return travel information expressed in specific Z3 variables, which may not be feasible given that APIs in the real world are typically black boxes that agents can only call. 
Following their two-stage solution, we first extract logical constraints from natural language. Based on these constraints, we implement a step-by-step plan generation process using depth-first search, mimicking how humans plan to travel by arranging activities one by one. 
As shown in Fig.~\ref{fig_dfs}, we first 
translate the natural languages to logical constraints through prompting. 
generate the next activity type based on the current plan, and then recursively generate the next activity until the goal is reached. The generated plan is then used to solve the problem.  
 In the second step, we define the rule-based activity selection and score function. For example, if the current time is in the [10:30, 12:30] and there is no scheduled lunch in the current plan, then the agent should find a restaurant to have lunch at this time. If the current time is after 22:00 and there are no open-time attractions nearby, the agent should choose to return to the hotel. For the score function, we select the restaurants that satisfy the required cuisine and sort the candidates by the price if there a budget constraints in the constraints $C$. These ranking functions will help us to find a feasible solution as soon as possible. In ChinaTravel, the duration arrangement of activities is continuous and difficult to enumerate and search. We pre-define a meal or a visit to an attraction as 90 minutes, and when there are less than 90 minutes until closing time, the event continues until the closing time. Given these designs, we adapt the neural-symbolic solution into a multi-POI planning problem and evaluate it in the ChinaTravel benchmark.

 \begin{algorithm*}[t]
   \caption{Depth-First Greedy Search }
   \label{alg2}
   \begin{algorithmic}
   \REQUIRE Constraints $C$, current plan $p$, 
   % \IF{not ConstraintValidation(p, C)}
   % \STATE \textbf{return} False, p  \COMMENT{The current plan has volidated the constraints.}
   % \ENDIF
   \IF{the least activity is an intercity-transport from destination to origin}
   \STATE \textbf{return} ConstraintValidation(p, C), p \COMMENT{The plan $p$ is finished, return the validation result.}
   % \ELSE
   % \STATE \textbf{return} False, p
   % \ENDIF 
   \ENDIF 
   \STATE type = GetNextActivityType(p) \COMMENT{Select the next type of activities, e.g. lunch, attraction.}
   \STATE candidates = ToolUse(type) \COMMENT{Collect the corresponding information for the activity type}
   \STATE scores = LLMScore(candidates, p, C) \COMMENT{Score candidates through constraints C.}
   \FOR{activity in candidates} 
   \STATE p.push(activity)  \COMMENT{Perform a greedy search with priority ranking.}
   \STATE flag, p = Depth-FirstGreedySearch(C, p)
   \IF{flag}
   \STATE \textbf{return} True, p \COMMENT{Return the solution $p$ if the validation is passed.} 
   \ENDIF
   \STATE p.pop(activity)
   \ENDFOR
   \STATE \textbf{return} False, p \COMMENT{Fail to find a solution with the given conditions.} 
   \end{algorithmic}
\end{algorithm*}

Given that some queries are particularly challenging due to the limited number of feasible plans, we set the maximum runtime for the symbolic sketch from interactive search to 5 minutes per query, excluding the LLM inference time, to ensure a fair comparison across different models. If a plan satisfying the generated DSL validation is found within the time limit, it is returned directly. Otherwise, the program halts when the time limit is reached, and the plan that satisfies environmental constraints while achieving the highest number of validation code successes among all intermediate results is returned. In cases where no environment-compliant plan is identified, the partially completed plan generated up to that point is returned. 

In the Figure~\ref{POIR_type},~\ref{POIR_rest} and~\ref{POIR_attr}, we provide the prompts of the LLM POI-ranking phases. 

\section{Evaluation Metric in Competition}
\label{app:metric}
The Delivery Rate (DR), Environmental Pass Rate (EPR), Logical Pass Rate (LPR), and Final Pass Rate (FPR) have been detailed in TravelPlanner~\citep{TravelPlanner}. 
To make the paper more self-contained, we also provide the corresponding definition here. 

\paragraph{Delivery Rate:} This metric assesses whether agents can successfully deliver a formatted plan. For the Nesy planning, if a solution that satisfies the logical constraints has not been found by the time out, the search is terminated, and the current solution that satisfies the environmental constraints is returned. If no solution that satisfies the environmental constraints is obtained, an empty plan is returned. 
Therefore, unlike the pure LLM method, which primarily assesses the Delivery Rate based on whether the output meets the formatting requirements, the nesy planning method, which uses depth-first-search to arrange POIs one by one, shows differences in the Delivery Rate. These differences mainly reflect the proportion of effective solutions obtained within a limited time based on the LLM's POI recommendation. This proportion demonstrates the degree to which the LLM prioritizes POI arrangements from a natural language perspective and meets formalized logical requirements. The more accurately the LLM can arrange POIs that are beneficial for long-horizon planning, the more likely it is to obtain effective solutions and improve the Delivery Rate. 

\paragraph{Environmental Pass Rate} Comprising the environmental dimensions (as detailed in Tab.~\ref{tab:environment_constraints}), this metric evaluates whether a language agent can accurately incorporate sandbox information into their generated plans. 

\begin{equation*}
    EPR-micro = \frac{\sum_{p\in P} \sum_{c\in Env}\1_{\textit{passed}(c, p)} }{|P| * |Env|}
\end{equation*}

\begin{equation*}
    EPR-macro = \frac{\sum_{p\in P} \prod_{c\in Env}\1_{\textit{passed}(c, p)} }{|P|}
\end{equation*}

\paragraph{Logical Pass Rate} Comprising the logical dimensions (as detailed in Tab.~\ref{logical_constraints}), this metric evaluates whether a language agent can accurately meet the personalized requirements of the queries. 
\begin{equation*}
    LPR-micro = \frac{\sum_{p\in P}  \sum_{c\in C_p} \1_{\textit{passed}(C_p, p)}}{\sum_{p\in P}  |C_p|}
\end{equation*}

\begin{equation*}
    LPR-macro = \frac{\sum_{p\in P}  \prod_{c\in C_p} \1_{\textit{passed}(C_p, p)}}{ |P|}
\end{equation*}

\paragraph{Final Pass Rate} This metric represents the proportion of feasible plans that meet all aforementioned constraints among all tested plans. It serves as an indicator of agents’ proficiency in producing plans that meet a practical standard.
\begin{equation*}
    FPR = \frac{\sum_{p\in P} \1_{\textit{passed}(Env, p)}\!\cdot\! \1_{\textit{passed}(C_p, p)}}{|P|}
\end{equation*}

\begin{table*}[th]
   \centering
    % \hspace*{-1.5cm}
    \small
   \caption{Multi-Preference Comparison of BQ and PEQ.}
   \label{preference_req}
   \setlength{\tabcolsep}{3pt} 
   \begin{tabular}{c|cccc|cccc|cc}
      \toprule
      Preference Combination & \multicolumn{2}{c}{Vaule-1} & \multicolumn{2}{c}{Vaule-2} & \multicolumn{2}{c}{Rank-1} & \multicolumn{2}{c}{Rank-2} & \multicolumn{2}{c}{Agg. Rank.}\\
      & BQ & PEQ & BQ & PEQ & BQ & PEQ & BQ & PEQ & BQ & PEQ \\
      \midrule
      P0 $\uparrow$, P1 $\downarrow$ & 0.79 & \textbf{0.83} & \textbf{28.0} & 29.7 & \textbf{1.44} & 1.55 & \textbf{1.44} & 1.55 & \textbf{1.44} & 1.55 \\
      \midrule
      P0 $\uparrow$, P2 $\downarrow$ & 0.82 & \textbf{1.26} & \textbf{29.0} & 31.9 & 1.56& \textbf{1.43} & \textbf{1.43} & 1.56 & 1.5 & 1.5 \\ 
      \midrule
      P0 $\uparrow$, P3 $\uparrow$ & 0.81 & \textbf{0.94} & 0.18 & \textbf{0.20} & \textbf{1.42} & 1.57 & 1.59 & \textbf{1.40} & 1.51 & \textbf{1.48} \\
      \midrule
      P0 $\uparrow$, P4 $\downarrow$ & 0.79 & \textbf{0.97} & 1221 & \textbf{441} & 1.46 & \textbf{1.53} & 1.73 & \textbf{1.26} & 1.59 & \textbf{1.40} \\
      \midrule
      P0 $\uparrow$, P5 $\downarrow$ & 0.78 & \textbf{0.91} & \textbf{33.6} & 34.0 & \textbf{1.37} & 1.62 & 1.70 & \textbf{1.29} & 1.54 & \textbf{1.45} \\
      \midrule
      P1 $\downarrow$, P2 $\downarrow$ & 28.2 & \textbf{27.8} & \textbf{26.6} & 30.1 & 1.62 & \textbf{1.37} & \textbf{1.48} & 1.51 & 1.55 & \textbf{1.44} \\
      \midrule
      P1 $\downarrow$, P3 $\uparrow$ & \textbf{28.2} & 36.2 & 0.20 & \textbf{0.27} & \textbf{1.31} & 1.68 & 1.6 & \textbf{1.4} & \textbf{1.45} & 1.54 \\
      \midrule
      P1 $\downarrow$, P4 $\downarrow$ & \textbf{30.3} & 44.8 & 1440 & \textbf{585} & \textbf{1.14} & 1.85 & 1.77 & \textbf{1.22} & \textbf{1.45} & 1.54 \\
      \midrule
      P1 $\downarrow$, P5 $\downarrow$ & \textbf{30.1} & 38.3 & 30.7 & \textbf{30.2} & \textbf{1.27} & 1.72 & 1.69 & \textbf{1.30} & \textbf{1.48} & 1.51 \\
      \midrule
      P2 $\downarrow$, P3 $\uparrow$ & 24.7 & \textbf{23.3} & 0.27 & 0.27 & \textbf{1.43} & 1.56 & 1.60 & \textbf{1.39} & 1.52 & \textbf{1.47} \\
      \midrule
      P2 $\downarrow$, P4 $\downarrow$ & 24.1 & \textbf{21.1} & 1687 & \textbf{719} & 1.51 & \textbf{1.48} & 1.89 & \textbf{1.10} & 1.70 & \textbf{1.29} \\
      \midrule
      P2 $\downarrow$, P5 $\downarrow$ & \textbf{28.0} & 30.8 & 29.4 & \textbf{26.0} & 1.51 & \textbf{1.48} & 1.89 & \textbf{1.10} & 1.70 & \textbf{1.29} \\
      \midrule
      P3 $\uparrow$, P4 $\downarrow$ & 0.18 & \textbf{0.26} & 1229 & \textbf{531} & 1.64 & \textbf{1.35} & 1.69 & \textbf{1.30} & 1.66 & \textbf{1.33} \\
      \midrule
      P3 $\uparrow$, P5 $\downarrow$ & 0.22 & 0.22 & 33.3 & \textbf{29.0} & 1.51 & \textbf{1.48} & 1.84 & \textbf{1.15} & 1.68 & \textbf{1.31} \\
      \midrule
      P4 $\downarrow$, P5 $\downarrow$ & 1366 & \textbf{767} & 33.1 & \textbf{31.6} & 1.67 & \textbf{1.32} & \textbf{1.45} & 1.54 & 1.56 & \textbf{1.43} \\
      \midrule
      % Win/Tie/Loss & & & & & & & & & 4/1/10 & \textbf{10/1/4} \\
      Aggregated Ranking & & & & & & & & & 1.56 & \textbf{1.43} \\
      \bottomrule
   \end{tabular}
\end{table*}
\section{Additional Experimental Results}

\subsection{Results with Large Reasoning Model}
\label{results_lrm}
The current experimental results have covered Qwen3-8B, the largest CoT-enabled reasoning models we could feasibly run within our local computational resources. We have further conducted the additional experiments with DeepSeek-R1 and DeepSeek-R1-Distill-Qwen-7B. 

Note that R1-Act is inherently a reason-then-act paradigm. The results show that pure-neural methods still struggle on ChinaTravel. Interestingly, DeepSeek-R1 does not consistently outperform R1-Distill-Qwen-7B. From the observation over experiments, one plausible reason is that R1 tends to over-think, which weakens final instruction-following in long contexts and yields unsatisfactory performance. Encouragingly, even 7-8B LLMs already exhibit some DSL-translating ability, so the community can conduct cost-effective post-training research on ChinaTravel with modest resources. 
\begin{table*}[tbh]
\centering
\small
\caption{Results on the Easy and Human-154 subsets.}
\label{tab:easy_human154_combined}
\setlength{\tabcolsep}{1pt} 
\begin{tabular}{@{}llrrrrrrr|rrrrrrr@{}}
\toprule
& & \multicolumn{7}{c}{\textbf{Easy}} & \multicolumn{7}{c}{\textbf{Human-154}}\\
\cmidrule(lr){3-9}\cmidrule(l){10-16}
Method & Model &
DR & \multicolumn{2}{c}{EPR} & \multicolumn{2}{c}{LPR} & C-LPR & FPR &
DR & \multicolumn{2}{c}{EPR} & \multicolumn{2}{c}{LPR} & C-LPR & FPR \\
& & & Mic. & Mac. & Mic. & Mac. & & & Mic. & Mac. & Mic. & Mac. & \\
\midrule
Act          & R1                 & 43.3 & 31.6 &  2.9 & 39.2 & 24.9 &  2.7 &  2.9 & 38.0 & 21.1 &  0.0 & 33.3 & 15.5 &  0.0 &  0.0 \\
NeSy         & R1-D.-Qwen-7B & 53.3 & 53.3 & 53.3 & 49.7 & 38.7 & 49.7 & 49.5 & 59.1 & 58.8 & 52.0 & 51.3 & 29.9 & 45.2 & 28.6 \\
NeSy         & R1                 & 58.3 & 58.3 & 58.3 & 53.7 & 32.6 & 53.7 & 32.6 & 46.8 & 46.6 & 45.5 & 39.2 & 23.4 & 38.0 & 23.4 \\
\midrule
NeSy \tiny{oracle} & R1-D.-Qwen-7B & 63.7 & 63.7 & 63.7 & 61.3 & 53.7 & 61.3 & 53.7 & 40.9 & 40.8 & 39.0 & 36.0 & 31.8 & 34.3 & 30.5 \\
NeSy \tiny{oracle} & R1                 & 50.0 & 50.0 & 50.0 & 49.8 & 46.3 & 49.8 & 46.3 & 43.5 & 43.5 & 42.9 & 39.7 & 33.8 & 39.2 & 33.8 \\
\bottomrule
\end{tabular}
\end{table*}

\subsection{Results on Medium Set}
For organizational coherence in the manuscript, we elected not to include medium-complexity experimental results in the main text. The medium set features user inputs containing 3-5 logical requirements, representing the mid-range complexity tier that bridges simple queries and the highly complex open-ended scenarios. 

\begin{table*}[tbh]
    %\small
    \footnotesize
    \centering
   % \vspace{-.5in}
   % \resizebox{.9\textwidth}{!}{%
   \setlength{\tabcolsep}{1.pt}
   \caption{Results of different LLMs and planning strategies on the ChinaTravel \textit{medium} subset.}
   \label{results_medium}
   \begin{tabular}{ll|ccccccc|ll|ccccccc}
      \toprule      
       %&\multirow{2}{*}{\makecell{LLMs}} 
       & &\multirow{2}{*}{DR} & \multicolumn{2}{c}{EPR} & \multicolumn{2}{c}{\makecell{LPR}} & \multirow{2}{*}{\makecell{C-LPR}} & \multirow{2}{*}{\makecell{FPR}} & & & \multirow{2}{*}{DR} & \multicolumn{2}{c}{EPR} & \multicolumn{2}{c}{\makecell{LPR}} & \multirow{2}{*}{\makecell{C-LPR}} & \multirow{2}{*}{\makecell{FPR}} \\
     \cmidrule(lr){4-7}
       &  & & Mic. & Mac. & Mic. & Mac.  & & & & & & Mic. & Mac. & Mic. & Mac. \\
      \midrule
      \multirow{2}{*}{\makecell{Act}} 
      & \makecell{\includegraphics[width=.023\linewidth]{imgs/deepseek.jpg}} & 72.7 & 52.3 & 0 & 63.5 & 15.3 & 0 & 0 & \multirow{3}{*}{\makecell{NSP}} & \makecell{\includegraphics[width=.023\linewidth]{imgs/deepseek.jpg}} & 71.3 & 71.9 & 69.3 & 69.4 & 50.0 & 69.3 & 46.7 \\
      & \makecell{\includegraphics[width=.023\linewidth]{imgs/gpt.jpg}} & 97.4 & 70.5 & 0 & 89.3 & 55.3 & 0 & 0& &\makecell{\includegraphics[width=.023\linewidth]{imgs/gpt.jpg}} & 68.0 & 68.0& 68.0 & 64.1	46.6 & 64.1 & 46.7\\
      \cline{3-9}
      \multirow{2}{*}{\makecell{ReAct \\ \scriptsize{(zero-shot)}}} & \makecell{\includegraphics[width=.023\linewidth]{imgs/deepseek.jpg}} & 41.3 & 35.2 & 0 & 37.6 & 4.0 & 0 & 0 & & \makecell{\includegraphics[width=.023\linewidth]{imgs/qwen_logo.jpeg}}& 53.3&45.9&16.0&49.2&33.3&14.8&8.50\\
      \cline{10-18}
      & \makecell{\includegraphics[width=.023\linewidth]{imgs/gpt.jpg}} & 92.0 & 54.8 & 0 & 78.6 & 22.7 & 0 & 0 & \multirow{3}{*}{\makecell{NSP \\ \scriptsize{oracle}}} & \makecell{\includegraphics[width=.023\linewidth]{imgs/deepseek.jpg}} & 68.6&65.4&54.0&66.2&61.3&52.5&54.0\\
      \cline{3-9}
      \multirow{2}{*}{\makecell{ReAct \\ \scriptsize{(one-shot)}}} & \makecell{\includegraphics[width=.023\linewidth]{imgs/deepseek.jpg}} & 82.7 & 77.1 & 3.33 & 82.6 & 48.7 & 2.95 & 1.33 & & \makecell{\includegraphics[width=.023\linewidth]{imgs/gpt.jpg}} & 60.8&59.4&54.9&60.3&58.2&60.3&56.9\\
      & \makecell{\includegraphics[width=.023\linewidth]{imgs/gpt.jpg}} & 94.7 & 69.2 & 0.67 & 91.8 & 64.0 & 0.53 & 0& & \makecell{\includegraphics[width=.023\linewidth]{imgs/qwen_logo.jpeg}}& 53.3&51.3&36.6&51.9&43.3&34.8&34.6
\\
      \bottomrule
    \end{tabular}
\end{table*}

\subsection{Multi-Preference Comparison}
For multi-preference scenarios (e.g., balancing "attraction visits $\uparrow$" and "transport time $\downarrow$"), we adopt an averaged aggregation approach, where rankings reflect the combined performance across all preferences. 
This framework ensures scalability and objectivity.  

To rigorously evaluate the ability of language agents to balance multiple soft constraints, we constructed 15 test subsets by pairing six user preferences (P0–P5) into all possible combinations (e.g., "P0 + P1"). Each subset contains queries with two preference requirements. We compared two methods, Baseline Query (BQ) and Preference-Enhanced Query (PEQ), by quantifying their performance through our DSL-based Preference Ranking metric. For each subset, we extracted numerical scores for both preferences (Value-1 and Value-2), computed individual rankings (Rank-1, Rank-2), and derived an aggregated ranking (Agg. Rank.) to reflect overall performance. 
The results are provided in the Table~\ref{preference_req}. 

From these results, we could find that: (1) \textbf{PEQ Outperforms BQ in Most Scenarios}: In 10/15 combinations, PEQ achieves superior aggregated rankings (Aggregated Ranking = 1.43 vs. BQ's 1.56). 
Notably, PEQ demonstrates stable improvements on preferences P3 (e.g., maximizing dining quality↑) and P4 (e.g., minimizing accommodation costs↓). For instance: In "P0$\uparrow$ + P4$\downarrow$", PEQ reduces accommodation costs by 64\% (Value-2: 441 vs. BQ's 1221) while maintaining high attraction counts (Value-1: 0.97 vs. 0.79). 
For "P3$\uparrow$ + P4$\downarrow$", PEQ simultaneously improves dining quality (Value-1: 0.26 vs. BQ's 0.18) and lowers costs (Value-2: 531 vs. 1229). This stability likely stems from the direct impact of POI selection on these preferences. LLMs in PEQ effectively prioritize low-cost hotels or high-quality restaurants through natural language hints (e.g., "reduce the cost on accommodations"), enabling explicit alignment with P3 and P4 requirements. 
(2) \textbf{Challenges in Balancing Multiple Preferences}: The results also reveal inherent difficulties in harmonizing conflicting preferences, particularly when optimizing one requirement necessitates sacrificing another. 
Notably, in the P0$\uparrow$ + P1$\downarrow$ scenario, PEQ underperforms BQ on both preferences, highlighting the inherent difficulty in resolving conflicting objectives. 
While PEQ marginally improves attraction counts (Value-1: 0.83 vs. BQ's 0.79), it incurs a 5.7\% increase in transport time (Value-2: 29.7 vs. BQ's 28.0). This trade-off results in a worse aggregated ranking for PEQ (1.55 vs. BQ's 1.44), indicating that the combined effect of conflicting preferences negates the benefits of natural language guidance. 
In 9/15 combinations, PEQ improves one preference at the expense of the other. For example: P1$\downarrow$ + P4$\downarrow$: PEQ reduces accommodation costs by 59\% (Value-2: 585 vs. BQ's 1440) but increases transport time by 48\% (Value-1: 44.8 vs. 30.3). 
The inability to concurrently satisfy both preferences underscores the limitations of current LLM-driven prioritization in handling trade-offs. 

Our experiments demonstrate that the neuro-symbolic agent (PEQ), enhanced by LLM-driven POI recommendation, outperforms baseline methods in multi-preference travel planning. 
By integrating natural language hints to guide POI selection, PEQ effectively translates user requirements into actionable itineraries, demonstrating its capability to handle synergistic preferences. However, balancing inherently conflicting objectives remains challenging. This highlights the need for future advancements, such as domain-specific fine-tuned LLMs to better resolve preference conflicts or multi-objective optimization techniques to systematically navigate trade-offs.

\subsection{Analysis of Pure-LLM Methods}
Pure LLM-based methods have demonstrated significant shortcomings in constraint satisfaction, as evidenced by their near-zero success rates in benchmarks like TravelPlanner. We also attempt the multi-round refinement methods like Reflexion. While theoretically promising, it is still impractical in our context. In preliminary evaluations, Reflexion not only failed to achieve improvements in constraint satisfaction (consistent 0\% FPR) but also incurred prohibitive computational costs due to its reliance on iterative token-heavy interactions. This rendered large-scale evaluation infeasible given our resource constraints. In light of their current limitations in constraint satisfaction, NeSy frameworks remain the effective pathway for real-world travel planning. Therefore, in the main body of this work, we mainly analyze the Nesy method. 

In this section, we further summarize the key failure modes of pure-LLM-based methods observed in our experiments:

(1) \textbf{Incorrect API Calls:} LLMs frequently generate invalid or hallucinated API calls, leading to cascading errors in downstream planning. For instance, models may query non-existent APIs (e.g., city\_transport\_select instead of inter\_city\_transport\_select) or misuse parameters (e.g., filtering attractions by an unsupported feature like "bus"). Such errors exhaust API call limits and prevent agents from retrieving essential information. 

(2) \textbf{Repetitive Output Loops} In iterative planning frameworks like ReAct, LLMs often enter infinite loops when resolving constraints. For example, an agent might repeatedly query transportation details for all candidate attractions, even after selecting one, due to a failure to update its internal state. This behavior mimics the ``hallucination loops" reported in TravelPlanner paper. 

(3) \textbf{Reasoning-Action Inconsistency.}
In ReAct framework, the model first reasons and then takes an action. However, the reasoning and the action are not always consistent. For example, the model may reason that the user wants to book a flight, but then take an action to check the information of trains. Another example is that the model may detect that the expenses exceed the budget but does not respond to this and ultimately generates a plan that exceeds the budget. 

(4) \textbf{Critical Information Missing.} 
Even when intermediate steps (e.g., API responses) are logged in a "notebook," LLMs frequently omit essential details when synthesizing final plans. A recurring failure is neglecting return transportation (e.g., omitting the train from Shanghai back to Beijing), which violates feasibility constraints. 

\begin{figure*}[t]
    \centering
    \includegraphics[width=\linewidth]{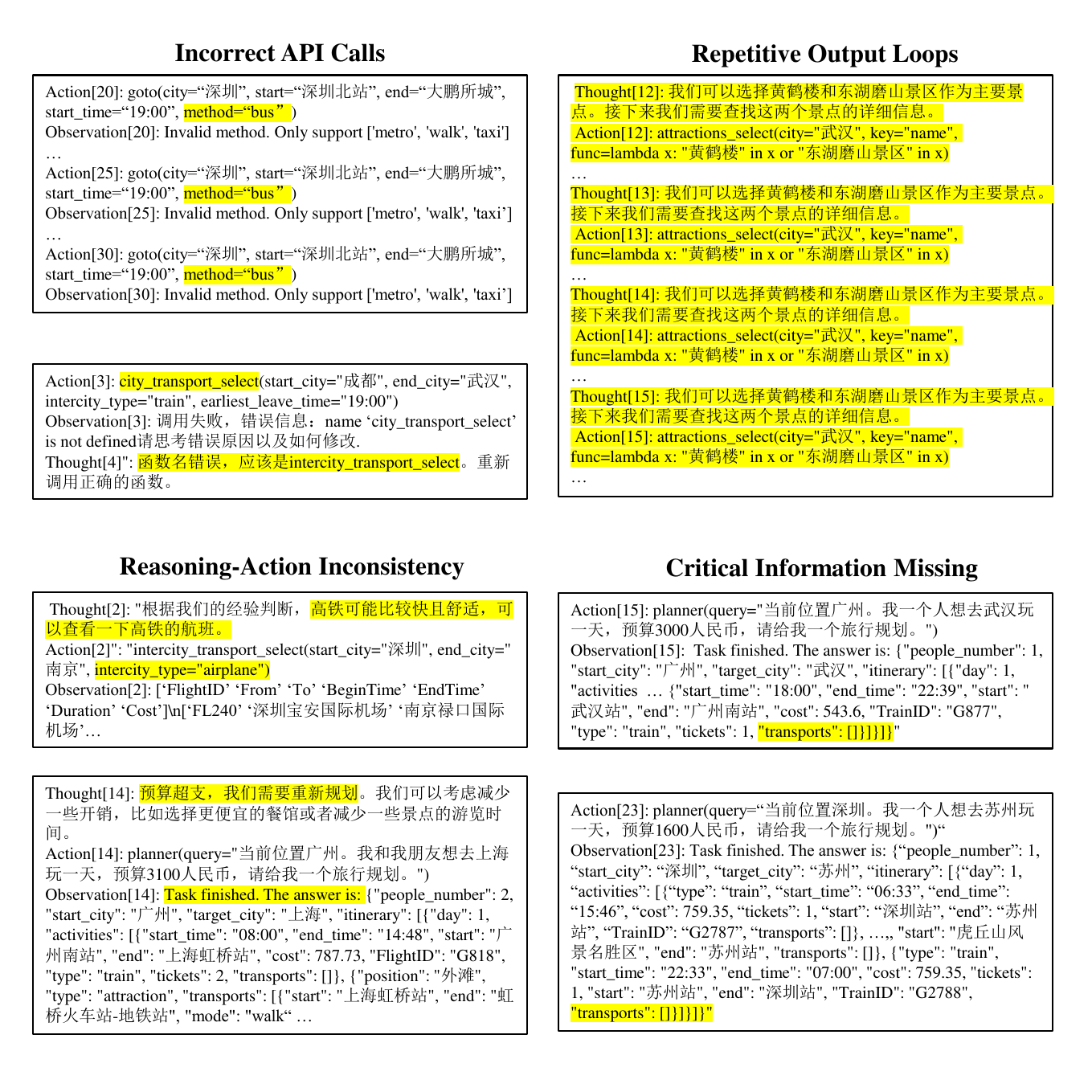}
    \caption{Fail case studies of React-one-shot DeepSeek Method.}
    \label{react_fail}
\end{figure*}

Figure~\ref{react_fail} provides the fail examples of ReAct (one-shot) with DeepSeek, which outperforms other pure-LLM-based methods in the main experiments. 

These limitations underscore the inadequacy of pure-LLM-based approaches for deployment in long-horizon and constraint-rich domains like travel planning.

\subsection{\blue{The Ablation Study of The Proposed NeSy Planning}}

\paragraph{\blue{The Impact of Iterative NL2DSL Translation.}} \blue{Tab.~\ref{tab:main_tab} explicitly compares NeSy Planning with or without oracle translation. This quantifies the translation module's impact, i.e., it is critical but currently a bottleneck due to unseen concept composition (as shown in the Fig.~\ref{generalization_error}).} 

\paragraph{\blue{The Impact of Symbolic Search Sketch.}}  \blue{In NeSy Planning, the symbolic search sketch uses DSL constraints to guide sequential construction with backtracking, whereas the LLM-modulo baseline only applies the same constraints for post-hoc error correction without search. LLM-modulo serves as an ablation. As a result, this search-based decomposition turns constraint feedback into much more effective plan refinement.} 

\paragraph{\blue{Impact of LLM-driven POI Ranking.}}
\blue{We ran NeSy and NeSy(Oracle) with random POI ranking while keeping all other components unchanged. As shown in the Tab.~\ref{tab_random_poi_ranking}, this leads to large and consistent drops in FPR: on the easy split from $74.0 \rightarrow 30.3$ and $52.6 \rightarrow 25.6$, and on the human split from $45.4 \rightarrow 38.3$ and $37.0 \rightarrow 31.8$. These results indicate that the LLM-driven ranking makes a substantial contribution by steering symbolic search toward semantically appropriate POIs. At the same time, even the random-ranking NeSy variants still significantly outperform pure-LLM agents in Sec. 4.2, suggesting that POI ranking is an important but not sole factor and that NL2DSL translation plus symbolic search are also crucial to the gains of NeSy Planning.}

\begin{table}[ht]
\centering
\begin{tabular}{l l c | l l c}
\toprule
\textbf{Easy-300} & \textbf{POI-Ranking} & \textbf{FPR} $\uparrow$ & \textbf{Human-154} & \textbf{POI-Ranking} & \textbf{FPR} $\uparrow$ \\
\midrule
NeSy & LLM & \textbf{52.6} & NeSy & LLM & \textbf{37.0} \\
NeSy & Random & 25.6 & NeSy & Random & 31.8 \\
\midrule
NeSy(Oracle) & LLM & \textbf{74.0} & NeSy(Oracle) & LLM & \textbf{45.4} \\
NeSy(Oracle) & Random & 30.3 & NeSy(Oracle) & Random & 38.3 \\
\bottomrule
\end{tabular}
\caption{\blue{Comparison of FPR across Easy-300 and Human-154 under different POI-Ranking methods.}}
\label{tab_random_poi_ranking}
\end{table}

\subsection{\blue{Results on English Setting}}
\blue{We have extended ChinaTravel to English and it is now a multilingual benchmark resource, making it convenient to global researchers and facilitating comparability. 
In the Tab.~\ref{results_easy_human}, we provide the preliminary validation on \textit{easy-300} and \textit{human-154}. The results confirm that the fundamental challenge raised by ChinaTravel is language-independent.} 
\begin{table*}[tbh]
    \footnotesize
    \centering
    \setlength{\tabcolsep}{1.pt}
    \begin{tabular}{l l|ccccccc|ccccccc}
      \toprule      
      \multirow{2}{*}{Method} 
      & \multirow{2}{*}{LLMs} 
      & \multirow{2}{*}{DR} 
      & \multicolumn{2}{c}{EPR} 
      & \multicolumn{2}{c}{LPR} 
      & \multirow{2}{*}{C-LPR} 
      & \multirow{2}{*}{FPR} 
      & \multirow{2}{*}{DR} 
      & \multicolumn{2}{c}{EPR} 
      & \multicolumn{2}{c}{LPR} 
      & \multirow{2}{*}{C-LPR} 
      & \multirow{2}{*}{FPR} \\
      \cmidrule(lr){4-7}\cmidrule(lr){11-14}
      &  &  & Mic. & Mac. & Mic. & Mac. &  &  &  & Mic. & Mac. & Mic. & Mac. \\
      \midrule
      ReAct 
      & \makecell{\includegraphics[width=.023\linewidth]{imgs/deepseek.jpg}} & 92.3 & 53.6 & 2.33 & 77.2 & 39.3 & 2.15 & 2.0 
      & 78.6 & 52.5 & 0.65 & 78.1 & 42.9 & 0.88 & 0 \\
      \midrule
      NeSy Planning 
      & \makecell{\includegraphics[width=.023\linewidth]{imgs/deepseek.jpg}}   & 82.3 & 81.9 & 81.3 & 77.2 & 57.7 & 76.6 & 57.7 
      & 59.7 & 59.1 & 57.8 & 51.5 & 41.6 & 49.6 & 40.9 \\
      & \makecell{\includegraphics[width=.023\linewidth]{imgs/gpt.jpg}}  & 75.7 & 75.1 & 75.0 & 70.0 & 49.7 & 69.6 & 49.7 
        & 49.3 & 49.3 & 47.4 & 41.4 & 33.8 & 40.2 & 33.8 \\
      & \makecell{\includegraphics[width=.023\linewidth]{imgs/qwen_logo.jpeg}} & 77.0 & 74.4 & 40.3 & 70.2 & 48.3 & 37.6 & 26.0 
      & 41.6 & 41.0 & 39.6 & 36.7 & 26.6 & 34.9 & 26.0 \\
      \midrule
      NeSy Planning
      & \makecell{\includegraphics[width=.023\linewidth]{imgs/deepseek.jpg}}   & 76.7 & 76.7 & 76.7 & 73.5 & 63.7 & 73.5 & 66.3 
      & 68.2 & 68.1 & 66.2 & 59.8 & 51.9 & 57.7 & 51.9 \\
      Oracle Translation 
      & \makecell{\includegraphics[width=.023\linewidth]{imgs/gpt.jpg}}  & 79.7 & 79.7 & 79.7 & 76.7 & 67.3 & 76.7 & 67.3 
      & 53.9 & 53.8 & 52.5 & 44.6 & 40.9 & 43.8 & 40.3 \\
      & \makecell{\includegraphics[width=.023\linewidth]{imgs/qwen_logo.jpeg}} & 77.0 & 77.0 & 77.0 & 73.9 & 62.3 & 73.9 & 62.3 
      & 61.0 & 61.0 & 59.7 & 52.2 & 43.5 & 51.1 & 43.5 \\
      \bottomrule
    \end{tabular}
    \caption{\blue{Results on \textit{Easy-300} and \textit{Human-154} from ChinaTravel-EN.}}
    \label{results_easy_human}
\end{table*}

\blue{From the results, we could find that, the performance of pure LLM methods on long-horizon agentic planning remains near 0\% in the English setting. This validates our core finding that LLMs fundamentally struggle, regardless of the sandbox language. Moreover, the results of NeSy methods, we could find the DSL translation bottleneck is still essential for grounding complex constraints.}

\subsection{\blue{Analysis of the Challenge on Composition Complexity}}
\blue{We further conducted experiments on Human-1000 to investigate how model performance scales with the complexity of composition. Following the compositional generalization community, we define composition complexity (C) as the number of basic concepts involved in a DSL requirement. Specifically, we evaluate the matching rate (\%) of POI Reasoning (correctly mapping user intent to specific POI requirements) and Syntax Generation (correctly translating query to the POI-masked DSL syntax) as the number of constraints (C) increases from 1 to 5. The results are provided in the Tab.~\ref{tab_composition_depth}.} 

\begin{table}[tbh]
\centering
\footnotesize
\begin{tabular}{l|ccccc|ccccc}
\toprule
 & \multicolumn{5}{c|}{POI Reasoning} & \multicolumn{5}{c}{Syntax Generation} \\
\cmidrule(lr){2-6}\cmidrule(lr){7-11}
\textbf{method} & C=1 & C=2 & C=3 & C=4 & C=5 & C=1 & C=2 & C=3 & C=4 & C=5 \\
\midrule
DeepSeek-V3 & 100 & 100 & 83.9 & 80.7 & 50.0 & 63.9 & 0 & 2.2 & 0 & 0 \\
GPT-4o      & 100 & 100 & 63.9 & 59.0 & 24.9 & 46.5 & 0 & 9.1 & 0 & 0 \\
Qwen3-8B    & -   & -   & -   & -   & -   & 39.2 & 0 & 0 & 0 & 0 \\
\bottomrule
\end{tabular}
\caption{\blue{Challenge Analysis with different constraint numbers $C$.}}
\label{tab_composition_depth}
\end{table}

\blue{\textbf{Performance Degradation with Composition Depth}: The experimental results clearly show that agent performance degrades significantly as the number of composed concepts (C) increases. This finding is consistent with general observations in the compositional generalization community. 
These results also further provide C-dependent evidence for our core claim: the compositional challenges introduced by ChinaTravel, both in syntax structure and semantic understanding, represent a fundamental bottleneck for existing LLMs.}

\section{Statements about Scientific Artifacts} 
\label{app:artifacts}
The ChinaTravel benchmark is designed to facilitate research in natural language processing and artificial intelligence, specifically for travel planning tasks. ChinaTravel includes a travel sandbox, user queries, and an evaluation framework intended for non-commercial, academic research purposes. 

\paragraph{Availability.} 
We will publicly release the ChinaTravel benchmark upon publication to facilitate community research. We look forward to broader adoption and extension of this benchmark.

\paragraph{Licenses.} The ChinaTravel benchmark and its associated datasets are licensed under the \href{https://creativecommons.org/licenses/by-nc/4.0/}{Creative Commons Attribution-NonCommercial 4.0 International (CC-BY-NC 4.0)} license. This license allows for the free use, distribution, and reproduction of the benchmark in any medium, provided that appropriate credit is given to the original authors and the source of the data is acknowledged, and that the use is for non-commercial purposes only. 

\paragraph{Data anonymization and offensive content.}
We anonymized the human queries during collection and instructed participants to avoid including sensitive information. We removed queries containing offensive content during the data cleaning process. 

\section{Statements about Human Participants}
\label{app:participants}
For the collection of Human-154, we recruited over 250 volunteers through a structured questionnaire to collect authentic Chinese travel requirements. Participants were informed about the public use of their data and instructed to avoid including sensitive personal information. During data cleaning, offensive content and identifiable details were removed. While no explicit ethics board approval is mentioned, we ensured compliance with anonymization practices and obtained participant consent for data inclusion. The final dataset contains 154 human-derived queries reflecting diverse real-world travel needs. 

\subsection{Instructions Given To Participants} 

To gather the authentic travel requirements, we collected data through a carefully designed questionnaire. We provided the following instruction information to the participants:

\begin{enumerate}
    \item The specific constraints the agent can handle and the corresponding details, including the types and specific names of attractions, restaurants, and hotels; requirements for intercity transportation (airplane or train) and urban transportation (walk, taxi or subway); as well as budget limitations for overall expenses or specific activities (such as accommodation and intercity transportation).
    \item The necessary information should be provided in the query, including the departure and destination cities of the trip, the number of travel days and constraint information.
    \item A detailed example with the query and travel planning response. 
    % \item Difficulty levels and corresponding examples, including easy, medium, and hard levels. 
\end{enumerate}

\begin{figure*}[ht]
    % \centering
    \includegraphics[width=\textwidth]{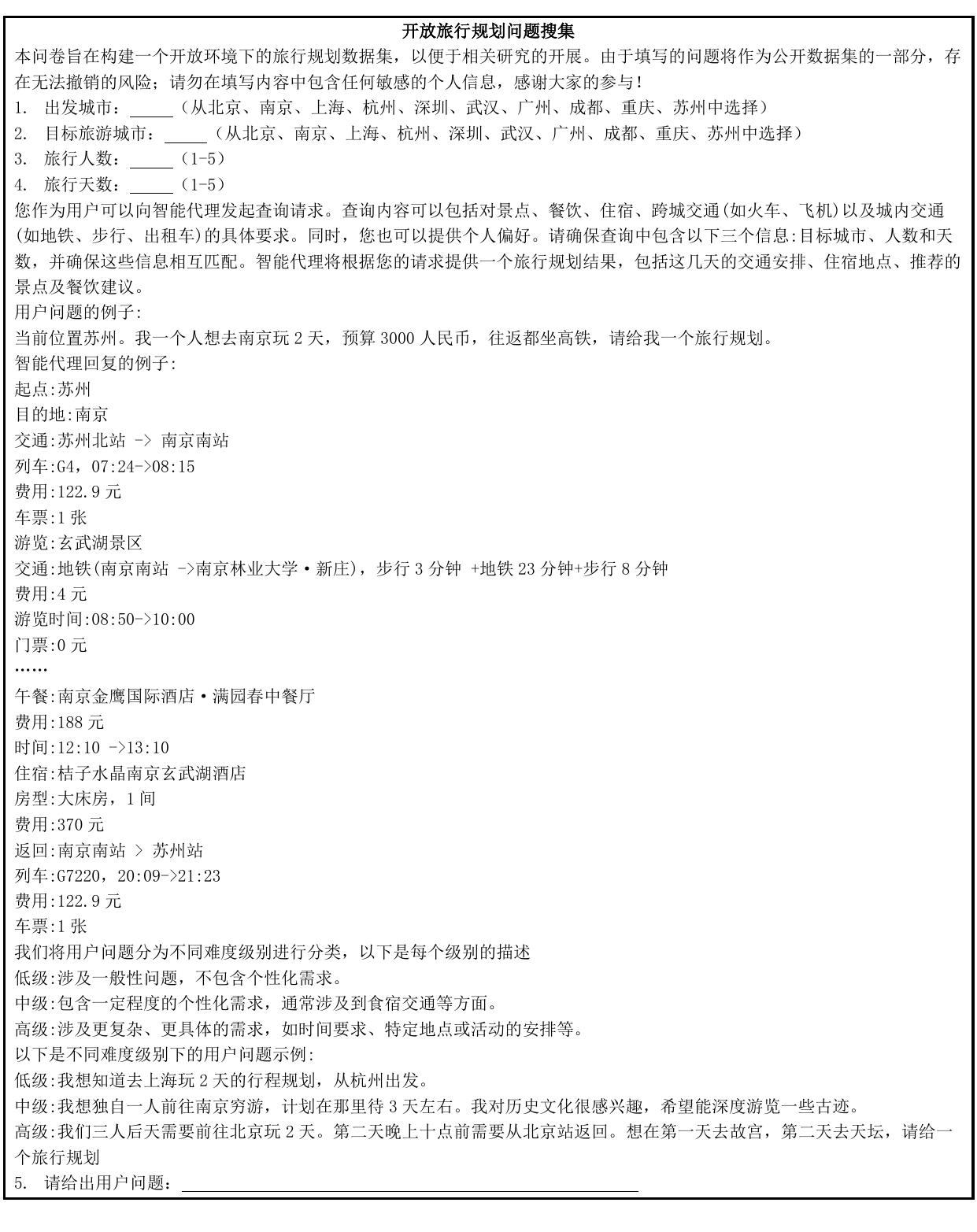}
    \caption{Questionnaire}
    \label{fig:questionnaire_cn}
\end{figure*}

\begin{figure*}[ht]
    % \centering
    \includegraphics[width=\textwidth]{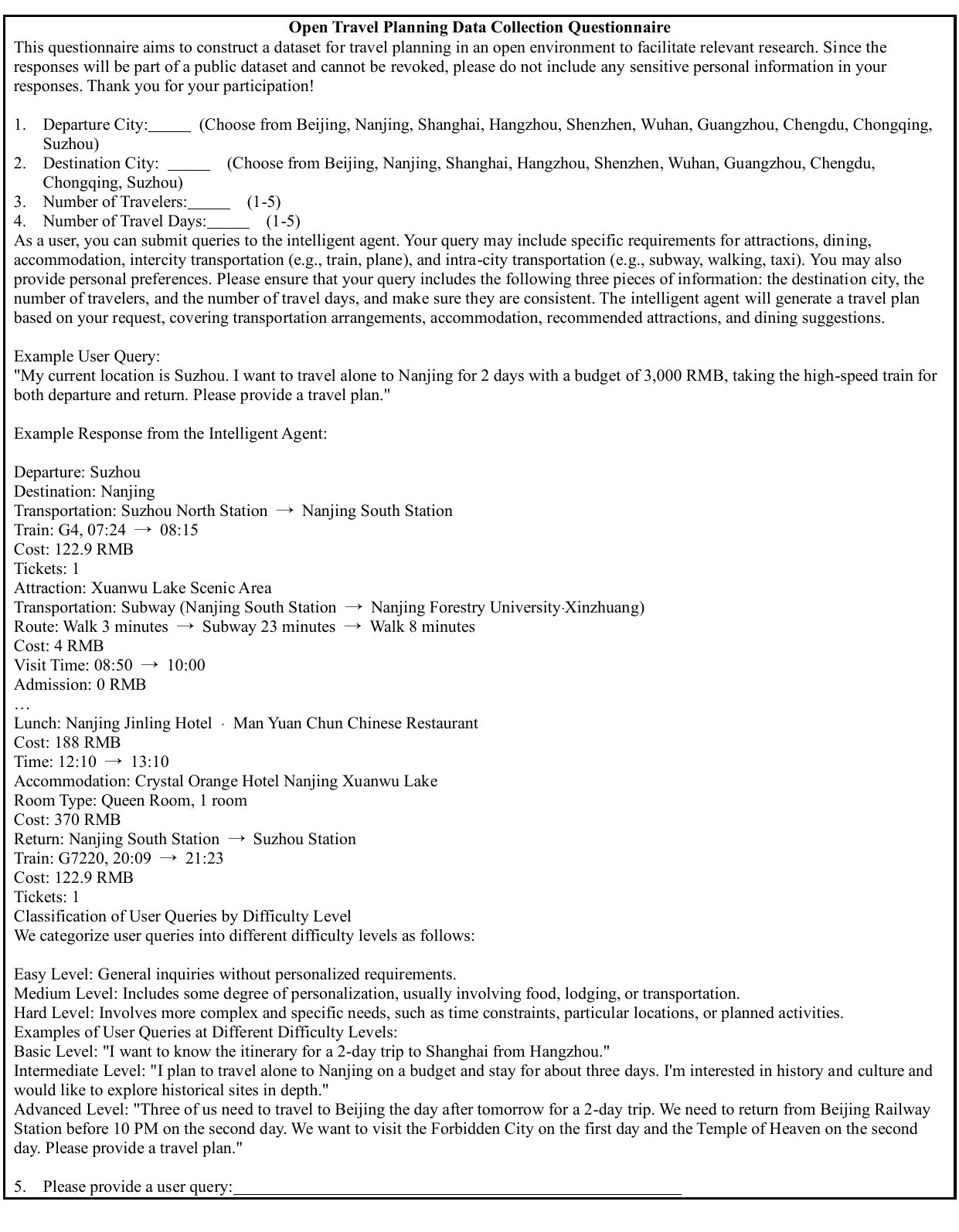}
    \caption{The translated version of the questionnaire}
    \label{fig:questionnaire_en}
\end{figure*}

Fig.~\ref{fig:questionnaire_cn} and Fig.~\ref{fig:questionnaire_en} respectively show the questionnaire and its translated version. 

\subsection{Recruitment And Payment}
For the collection of Human-154, we recruited a total of 250 student volunteers to provide authentic Chinese travel requirements. The participants included 121 undergraduate students, 86 master's students, and 43 doctoral students. The task of understanding the query background and providing travel requirements was estimated to take 1-2 minutes per participant. Given the simplicity of the task and the fact that it did not require extensive professional background or expertise, we compensated each participant with 1 CNY. This compensation was deemed adequate considering the nature of the task and the time required to complete it. The payment was determined based on the estimated time and the straightforward nature of the natural language requirements, ensuring a fair and reasonable reward for the participants. 

For Human-1000, we partnered with WJX (a professional survey platform) to scale data collection. Each valid query was incentivized with 6 CNY. After WJX's initial screening, our team rigorously annotated responses, filtering invalid entries (e.g., nonsensical inputs). It finally yielded 1,000 high-quality queries meeting DSL annotation standards, ensuring both diversity and alignment with real-world planning scenarios. 

\subsection{Data Consent}

When collecting the data, we clearly informed the participants about the usage of the data and the potential irreversible risks of it becoming part of a public dataset. We did not track the ID information of the questionnaire respondents. Additionally, we reminded participants not to include any sensitive personal information in the questionnaire responses. During the data cleaning process, we directly removed queries containing offensive content and filtered out sensitive identity information. % in the remaining queries.

\subsection{Institute Ethics Approval and Risk Mitigation}

Our questionnaire posed no more than minimal risk: it collected only non-sensitive travel preferences, caused no physical or psychological harm, and preserved participant anonymity. The foreseeable risks were limited to minor time cost. All participants were clearly informed about data usage and gave voluntary consent. In our institute, minimal-risk studies like ours are exempt from convening a dedicated ethics committee. Moreover, our institute explicitly confirms that our questionnaire minimized any potential risk to participants and formally authorized the creation and release of the benchmark. 

The risk mitigation strategies we employed are as follows.

\textbf{Risk Assessment and Disclosure:} We conducted a thorough assessment concluding the study posed minimal risk. All identified potential risks were fully disclosed to participants. 
\textbf{Informed Consent:} Written informed consent was obtained from every participant prior to involvement. Consent documents clearly explained the study purpose, procedures, potential risks, data handling (anonymization and usage), voluntary nature, and right to withdraw. 
\textbf{Privacy Protection:} Strict data anonymization protocols were applied. No personally identifiable information (PII) is present in the collected or released dataset. Data security measures were enforced. 
\textbf{Voluntary Participation and Fairness:} Participation was voluntary, and fair compensation was provided. Thank you for your suggestions. We will add them in the final revision. 

\subsection{Risk Statement for Participants}

Here's the English translation of our risk statement for participants:

This questionnaire aims to create an open-environment travel planning dataset to support academic research. Important Notes:  

Data Irrevocability: As a public dataset, submitted data may not be revoked once published.

Indefinite Retention: As a public dataset, submitted data may be retained indefinitely.

Anonymization: All submitted data will be anonymized.

Sensitive Information: Please DO NOT include any sensitive personal information in your responses. (Note: We will collect limited personal information solely to analyze data source diversity. This information will be strictly protected and used only for this specific purpose).

Dataset License: The dataset will be released under the CC BY-NC-SA 4.0 license.
Summary: This license allows free use, modification, and sharing for non-commercial purposes only, provided users:

Give appropriate credit (attribution),
Share any adaptations under the same license (share alike),
And do not use the material commercially.

Full License: We strongly recommend reviewing the complete CC BY-NC-SA 4.0 license terms:
Consent Declaration:
By submitting this questionnaire, you explicitly consent to our use of the data you provide for non-commercial purposes, including but not limited to:

Algorithm/model development and optimization.
Publication of academic research.
Any other uses permitted under the CC BY-NC-SA 4.0 license.

\subsection{Characteristics of Annotators} 
Our data collection process solely involved travel requirements and did not include any protected information, such as sexual orientation or political views as defined under the General Data Protection Regulation (GDPR). All data were collected from native Chinese speakers to ensure that the travel requirements fully align with the context and nuances of the Chinese language. This approach was taken to accurately capture the needs and preferences of the target population, which is primarily composed of Chinese-speaking individuals. The annotators were recruited from a diverse range of academic backgrounds, including undergraduate, master's, and doctoral students, to provide a broad and representative set of travel requirements. 

\subsection{DSL Annotation for Human Data}
The annotation process for the human data involved four stages to ensure the accuracy and validity of the Domain-Specific Language (DSL) annotations:
(1) Initial DSL Version Generation: GPT-4o was utilized to provide the initial version of the DSL annotations for the human data. This step aimed to leverage the language model's capabilities to generate a baseline for further refinement. 
(2) Data Annotation Team Revision: A team of five data annotators was responsible for reviewing and revising the DSL annotations. The team members divided the workload and made necessary corrections to the DSL annotations to ensure their accuracy and relevance to the travel requirements.
(3) Primary Developer Verification and Correction: Three of the main developers of the benchmark conducted a thorough review of all the DSL annotations. They verified the correctness of the annotations and made revisions as needed. This stage also involved the exclusion of any invalid queries that could not be verified within the sandbox environment. 
(4) Final Verification by Primary Developers: The same three main developers performed a final check on all the DSL annotations. This step ensured that the annotations were accurate, consistent, and met the required standards for the benchmark.

Throughout the annotation process, the focus was on ensuring that the DSL annotations accurately captured the travel requirements and were valid within the context of the ChinaTravel benchmark's sandbox environment. The annotation process for human data required a deep understanding of the ChinaTravel DSL and involved joint debugging and verification with the sandbox information. This significantly limited the size of the annotation team, as only a limited number of annotators had the necessary expertise and familiarity with both the DSL and the sandbox environment. Additionally, the process was time-consuming and required meticulous attention to detail, further constraining the rate at which the human dataset could grow. Despite these challenges, the rigorous annotation process ensured the quality and reliability of the human data, which is crucial for the evaluation and development of language agents in real-world travel planning. % scenarios. 

\subsection{\blue{Temporal Coverage of Human Data}}
\blue{The Human-154 Val Set was collected from August 2024 to October 2024. The Human-1000 Test Set was collected longitudinally over a significant period, spanning November 2024 to April 2025. The six-month collection window for the Human-1000 test set ensures natural temporal diversity. This longitudinal approach captured a broad spectrum of real-world implicit intents related to major festivals (e.g., Spring Festival), seasonal travel patterns (winter breaks, spring outings), and varying weather/peak periods. This confirms our test set is not overfitted to a single season, thus providing a robust evaluation of agent generalization capabilities.}

\begin{figure*}[th]
\begin{tcolorbox}[
                standard jigsaw,
                % colback=white, % 背景颜色
                % colframe=white, % 边框颜色
                coltitle = black,
                colbacktitle = white,
                title=Prompts for POI recommendation,     % 标题
                opacityback=0,
                fonttitle=\bfseries,   % 标题字体
                % halign=left,         
                % valign=center,         
                % colupper=black,        % 内容文字颜色
                % width=\linewidth,      % 设置宽度
                 % height=5cm,            % 设置高度
]

\begin{lstlisting}[frame=none]

NEXT_POI_TYPE_INSTRUCTION = """ 
   You are a travel planning assistant. 
   The user's requirements are: {}. 
   Current travel plans are: {}. 
   Today is {}, current time is {}, current location is {}, and POI_type_list is {}. 
   Select the next POI type based on the user's needs and the current itinerary. 
   Please answer in the following format.
   Thought: [Your reason]
   Type: [type in POI_type_list]
    """
\end{lstlisting}
\end{tcolorbox}
\caption{Prompts for next-POI-type recommendation}
\label{POIR_type}
\end{figure*}

\begin{figure*}[th]
\begin{tcolorbox}[
                standard jigsaw,
                % colback=white, % 背景颜色
                % colframe=white, % 边框颜色
                coltitle = black,
                colbacktitle = white,
                title=Prompts for restaurants recommendation,     % 标题
                opacityback=0,
                fonttitle=\bfseries,   % 标题字体
                % halign=left,         
                % valign=center,         
                % colupper=black,        % 内容文字颜色
                % width=\linewidth,      % 设置宽度
                 % height=5cm,            % 设置高度
]

\begin{lstlisting}[frame=none]


RESTAURANT_RANKING_INSTRUCTION = """
    You are a travel planning assistant. 
    The user's requirements are: {user_requirements}. 
    The restaurant info is:
    {restaurant_info}
    The past cost for intercity transportation and hotel accommodations is: {past_cost}.
    
    Your task is to select and rank restaurants based on the user's needs and the provided restaurant information. Consider the following factors:
    1. Restaurant name
    2. Cuisine type
    3. Price range
    4. Recommended food
    
    Additionally, keep in mind that the user's budget is allocated across multiple expenses, including intercity transportation and hotel accommodations. Ensure that the restaurant recommendations fit within the remaining budget constraints after accounting for the past cost. 
    Note that the price range provided for each restaurant is the average cost per person per meal, the remaining budget must cover the cost of three meals per day for {days} days.
    
    For each day, recommend at least 6 restaurants, combining restaurants for all days together.  
    
    Your response should follow this format:
    
    Thought: [Your reasoning for ranking the restaurants]
    RestaurantNameList: [List of restaurant names ranked by preference, formatted as a Python list]
    """

\end{lstlisting}
\end{tcolorbox}
\caption{Prompts for restaurant recommendation}
\label{POIR_rest}
\end{figure*}

\begin{figure*}[th]
\begin{tcolorbox}[
                standard jigsaw,
                % colback=white, % 背景颜色
                % colframe=white, % 边框颜色
                coltitle = black,
                colbacktitle = white,
                title=Prompts for attractions recommendation,     % 标题
                opacityback=0,
                fonttitle=\bfseries,   % 标题字体
                % halign=left,         
                % valign=center,         
                % colupper=black,        % 内容文字颜色
                % width=\linewidth,      % 设置宽度
                 % height=5cm,            % 设置高度
]

\begin{lstlisting}[frame=none]


ATTRACTION_RANKING_INSTRUCTION = """
    You are a travel planning assistant. 
    The user's requirements are: {user_requirements}. 
    The attraction info is:
    {attraction_info}
    The past cost for intercity transportation and hotel accommodations is: {past_cost}.
    
    Your task is to select and rank attractions based on the user's needs and the provided attraction information. Consider the following factors:
    1. Attraction name
    2. Attraction type
    3. Location
    4. Recommended duration
    
    Additionally, keep in mind that the user's budget is allocated across multiple expenses, including intercity transportation and hotel accommodations. Ensure that the attraction recommendations fit within the remaining budget constraints after accounting for the past cost.
    
    For each day, recommend at least 8 attractions, combining attractions for all days together. To ensure a comprehensive list, consider a larger pool of candidates and prioritize diversity in attraction type and location.
    
    Your response should follow this format:
    
    Thought: [Your reasoning for ranking the attractions]
    AttractionNameList: [List of attraction names ranked by preference, formatted as a Python list]

    Example:
    Thought: Based on the user's preference for historical sites and natural attractions, the attractions are ranked as follows:
    AttractionNameList: ["Attraction1", "Attraction2", ...]
    """

\end{lstlisting}
\end{tcolorbox}
\caption{Prompts for attraction recommendation}
\label{POIR_attr}
\end{figure*}

\section{The Implementation of TTG Baseline}
\label{app_ttg}
\subsection{Constraints Formulation}
TTG~\citep{ju2024globe} models the travel planning problem as a MILP (Mixed-Integer Linear Programming) problem. We adapt their formulation into ChinaTravel for solver-based optimization and the specific parameters, variable and constraint settings can be found in Tab.~\ref{tab:ttg_parameters}\ref{tab:ttg_variables}\ref{tab:ttg_constraints}.

\subsection{Experimental Setting} 
Although TTG performs very well on Travelplanner, the solver takes slightly more than 1 second on average to complete the computation. On the ChinaTravel benchmark, the rapid growth of constraints in TTG becomes computationally prohibitive. If we use the full sandbox, the average number of constraints will exceed \textbf{10B} (For detailed calculations of variable sizes and the number of constraints, please refer to Tab.~\ref{tab:ttg_variable_sizes}\ref{tab:ttg_constraint_sizes}). Therefore, we only include 22 POIs (2 hotels, 10 attractions, 5 restaurants, 5 stations, 100 intercity transports each for arrivals and departures) and use one hour as a time step. We use LLMs to select them from sandbox to ensure sufficient flexibility in handling different queries. Nonetheless, its constraint scale still reaches $320k \times \text{days}$ and the number of variables also reaches $36k \times \text{days}$. In comparison, the commonly used benchmark for evaluating MILP solvers, MIPLIB 2017~\cite{gleixner2021miplib}, contains only 10 instances with more than 320k constraints and about 60 instances with over 36k variables (out of a total of 1065 instances).

In our main experiments, using the SCIP solver from the PuLP package, TTG was allocated a relaxed 15-minute search limitation. However, this configuration yielded only 18\% valid solutions on easy-subset instances, with the final pass rate (FPR) further reduced to 8\% due to the solver's pruning heuristics. Fig.~\ref{challenge_nesy}(a) illustrates the solution time of TTG on 1- to 3-day itineraries. Within the time limit, solutions were found for merely 23\% of two-day and 6\% of three-day itineraries.

\begin{table}[ht]
\centering
\renewcommand{\arraystretch}{1.4}
\begin{tabular}{ll}
\hline
\textbf{Parameter} & \textbf{Meaning} \\
\hline
$\textit{hotelNum}$ & Number of hotels \\
$\textit{attrNum}$ & Number of attractions \\
$\textit{restNum}$ & Number of restaurants \\
$\textit{transNum}$ & Number of transport modes \\
$\textit{stationNum}$ & Number of stations \\
$\textit{goNum}$ & Number of arriving trains/buses \\
$\textit{backNum}$ & Number of departing trains/buses \\
$\textit{timeStep}$ & Number of time steps \\
\hline
$\textit{locNum} = \textit{hotelNum} + \textit{attrNum} + \textit{restNum}$ & Total number of POI locations except stations\\
$\textit{totalNum} = \textit{locNum} + \textit{stationNum}$ & Total number of all locations including stations \\
\hline
\end{tabular}
\vspace{0.5cm}
\caption{Definition of parameters used in TTG}
\label{tab:ttg_parameters}
\end{table}

\begin{table}[ht]
\centering
\begin{tabular}{ll}
\hline
\textbf{Variable} & \textbf{Meaning} \\
\hline
$u[\text{idx}][t]$ & The traveler is at location \texttt{idx} at time $t$ \\
$\text{event}[t]$ & The traveler's location changes at time $t$ \\
$\text{hotel}[\text{idx}][d]$ & Number of times the traveler visits hotel \texttt{idx} on day $(d+1)$ \\
$\text{attr}[\text{idx}]$ & Number of times the traveler visits attraction \texttt{idx} \\
$\text{rest}[\text{idx}][\text{meal}]$ & Number of times the traveler visits restaurant \texttt{idx} at meal \texttt{meal} \\
$z_{\text{hotel}}, z_{\text{attr}}, z_{\text{rest}}, \delta$ & Auxiliary variables \\
$\text{needEat}[m]$ & Whether the traveler needs to eat meal $m$ (during intercity travel) \\
$\text{check[idx][t]}$ & Whether the attraction idx is open at time t \\
$\text{y[(i, j, tr,t)]}$ & The solution, a matrix of shape $\textit{totalNum} \times \textit{totalNum} \times \textit{transNum} \times \textit{timeStep}$ \\
\hline
\end{tabular}
\vspace{0.5cm}
\caption{Variables used in TTG}
\label{tab:ttg_variables}
\end{table}

\begin{table}[ht]
\centering
\renewcommand{\arraystretch}{1.5}
\begin{tabular}{p{4cm}p{10cm}}
\hline
\textbf{Constraint Type} & \textbf{Mathematical Formulation} \\
\hline
\textbf{Spatio-temporal} &
$\delta[\text{idx}][t] \geq u[\text{idx}][t+1] - u[\text{idx}][t]$ \\
\textbf{Constraints} & $\delta[\text{idx}][t] \geq u[\text{idx}][t] - u[\text{idx}][t+1]$ \\
& $\text{event}[t] = 0 \Rightarrow u[\text{idx}][t] = u[\text{idx}][t+1]$ \\
& $\text{event}[t] = 1 \Rightarrow \sum_{\text{idx}} \delta[\text{idx}][t] = 2$ \\
& $\sum_i u[i][t] = 1$ \\
\hline
\textbf{Hotel Constraints} &
$z_\text{hotel}[\text{idx}][t] = u[\text{idx}][t] \wedge \text{event}[t]$ \\
& $\text{hotel}[\text{idx}][d] = \sum_{t = d \cdot \text{stepPerDay}}^{(d+1)\cdot \text{stepPerDay}} z_\text{hotel}[\text{idx}][t]$ \\
& $\sum_{\text{idx}} \text{hotel}[\text{idx}][d] = 1$ \\
\hline
\textbf{Attraction Constraints} &
$z_\text{attr}[\text{idx}][t] = u[\text{idx}][t] \wedge \text{event}[t]$ \\
& $\text{attr}[\text{idx}] = \sum_{t} z_\text{attr}[\text{idx}][t]$ \\
& $\sum_{\text{idx}} \text{attr}[\text{idx}] \geq \text{min\_attr}$ \\
& $\text{check}[\text{idx}][t] = \text{False} \Rightarrow u[\text{idx}][t] = 0$ \\
\hline
\textbf{Meal Necessity} &
$\text{needEat}[m] = 1 \Rightarrow a[m] < T_{\text{dep}}$ \\
& $\text{needEat}[m] = 1 \Rightarrow b[m] > T_{\text{arr}}$ \\
\hline
\textbf{Innercity Transport} &
$y[(i,j,\text{tran},t)] \leq u[i][t]$ \\
\textbf{Constraints} & $y[(i,j,\text{tran},t)] \leq \text{event}[t]$ \\
& $y[(i,j,\text{tran},t)] \leq u[\text{tran}][t+1]$ \\
& $y[(i,j,\text{tran},t)] \leq u[\text{tran}][t+\delta]$ \\
& $y[(i,j,\text{tran},t)] \leq \text{event}[t+\delta]$ \\
& $y[(i,j,\text{tran},t)] \leq u[j][t+\delta+1]$ \\
\hline
\textbf{Restaurant Constraints} &
$z_\text{rest}[\text{idx}][t] = u[\text{idx}][t] \wedge \text{event}[t]$ \\
& $\text{rest}[\text{idx}][m] = \sum_{t = a[m]}^{b[m]} z_\text{rest}[\text{idx}][t]$ \\
& $\sum_{\text{idx}} \text{rest}[\text{idx}][m] \leq \text{needEat}[m]$ \\
& $\text{check}[\text{idx}][t] = \text{False} \Rightarrow u[\text{idx}][t] = 0$ \\
\hline
\textbf{Intercity Travel} &
$\sum_i \text{interGo}[i] = 1$ \\
\textbf{Constraints} & $\sum_i \text{interBack}[i] = 1$ \\
& $\text{interGo}[i] = 1 \Rightarrow u[\text{goStation}[i]][t] = 1$ \\
& $\text{interBack}[i] = 1 \Rightarrow u[\text{backStation}[i]][t] = 1$ \\
\hline
\end{tabular}
\caption{Constraints used in TTG}
\label{tab:ttg_constraints}
\end{table}

\begin{table}[ht]
\centering
\renewcommand{\arraystretch}{1.4}
\begin{tabular}{lc}
\hline
\textbf{Variable} & \textbf{Dimension} \\
\hline
$u[\text{idx}][t]$ & $(\textit{totalNum} + \textit{transNum}) \times \textit{timeStep}$ \\
$\delta[\text{idx}][t]$ & $(\textit{totalNum} + \textit{transNum}) \times (\textit{timeStep} - 1)$ \\
$\text{event}[t]$ & $\textit{timeStep}$ \\
$\text{hotel}[\text{idx}][d]$ & $\textit{hotelNum} \times \textit{days}$ \\
$z_\text{hotel}[\text{idx}][t]$ & $\textit{hotelNum} \times \textit{timeStep}$ \\
$\text{attr}[\text{idx}]$ & $\textit{attrNum}$ \\
$z_\text{attr}[\text{idx}][t]$ & $\textit{attrNum} \times \textit{timeStep}$ \\
$\text{rest}[\text{idx}][\text{meal}]$ & $\textit{restNum} \times 3 \times \textit{days}$ \\
$z_\text{rest}[\text{idx}][t]$ & $\textit{restNum} \times \textit{timeStep}$ \\
$y[(i, j, \text{tr}, t)]$ & $\textit{totalNum} \times \textit{totalNum} \times \textit{transNum} \times \textit{timeStep}$ \\
$\text{total}$ & $\textit{days} \times\textit{stepPerHour} \times 36k$ \\
\hline
\end{tabular}
\caption{Variable sizes in TTG}
\label{tab:ttg_variable_sizes}
\end{table}

\begin{table}[ht]
\centering
\renewcommand{\arraystretch}{1.4}
\begin{tabular}{lc}
\hline
\textbf{Category} & \textbf{Estimated Size} \\
\hline
Spatio-temporal constraints & $(\textit{totalNum} + \textit{transNum}) \times (4 \times \textit{timeStep} + 3)$ \\
Hotel constraints & $\textit{hotelNum} \times (3 \times \textit{timeStep} + \textit{days})$ \\
Attraction constraints & $4 \times \textit{attrNum} \times \textit{timeStep}$ \\
Restaurant constraints & $\textit{restNum} \times (4 \times \textit{timeStep} + \textit{days})$ \\
Urban transport constraints & $7 \times \textit{totalNum}^2 \times \textit{transNum} \times \textit{timeStep} + 4 \times \textit{totalNum} \times \textit{timeStep}$ \\
Intercity transport constraints & $(\textit{goNum} + \textit{backNum}) \times \textit{timeStep}$ \\
\hline
\end{tabular}
\caption{Number of constraints sizes in TTG}
\label{tab:ttg_constraint_sizes}
\end{table}

\end{document}